\documentclass{ieeeaccess}
  
\usepackage{cite}
\usepackage{algorithmic}
\usepackage{textcomp}

\usepackage{commath}

\usepackage{subfig}
\usepackage{graphicx}
\usepackage{hhline}
\usepackage{diagbox}
\usepackage{xfrac}

\usepackage{mathtools}
\usepackage{amssymb}
\usepackage{amsbsy}
\usepackage{amsmath}
\newcommand\numberthis{\addtocounter{equation}{1}\tag{\theequation}}

\usepackage{adjustbox}

\usepackage[]{footmisc}
\usepackage{url}
\usepackage{bm}
\usepackage{multirow}

\usepackage{tikz}
\NewSpotColorSpace{PANTONE}
\AddSpotColor{PANTONE} {PANTONE3015C} {PANTONE\SpotSpace 3015\SpotSpace C} {1 0.3 0 0.2}
\SetPageColorSpace{PANTONE}%

\begin{document}

\history{ }
\doi{ }

\title{Truly Generalizable Radiograph Segmentation with Conditional Domain Adaptation}
\author{Hugo~Oliveira\authorrefmark{1},
        Edemir~Ferreira\authorrefmark{1},
        and~Jefersson~dos~Santos\authorrefmark{1}}
\address[1]{Computer Science Department, Universidade Federal de Minas Gerais, Belo Horizonte, Brazil}
\tfootnote{Authors would like to thank NVIDIA for the donation of the GPUs that allowed the execution of all experiments in this paper. We also thank CAPES, CNPq (424700/2018-2), and FAPEMIG (APQ-00449-17) for the financial support provided for this research.}

\markboth
{Oliveira \headeretal: Truly Generalizable Radiograph Segmentation with Conditional Domain Adaptation}
{Oliveira \headeretal: Truly Generalizable Radiograph Segmentation with Conditional Domain Adaptation}

\corresp{Corresponding author: Hugo N. Oliveira (e-mail: oliveirahugo@dcc.ufmg.br).}

\begin{abstract}
Digitization techniques for biomedical images yield different visual patterns in radiological exams. These differences may hamper the use of data-driven approaches for inference over these images, such as Deep Neural Networks. Another noticeable difficulty in this field is the lack of labeled data, even though in many cases there is an abundance of unlabeled data available. Therefore an important step in improving the generalization capabilities of these methods is to perform Unsupervised and Semi-Supervised Domain Adaptation between different datasets of biomedical images. In order to tackle this problem, in this work we propose an Unsupervised and Semi-Supervised Domain Adaptation method for segmentation of biomedical images using Generative Adversarial Networks for Unsupervised Image Translation. We merge these unsupervised networks with supervised deep semantic segmentation architectures in order to create a semi-supervised method capable of learning from both unlabeled and labeled data, whenever labeling is available. We compare our method using several domains, datasets, segmentation tasks and traditional baselines, such as unsupervised distance-based methods and reusing pretrained models both with and without Fine-tuning. We perform both quantitative and qualitative analysis of the proposed method and baselines in the distinct scenarios considered in our experimental evaluation. The proposed method shows consistently better results than the baselines in scarce labeled data scenarios, achieving Jaccard values greater than 0.9 and good segmentation quality in most tasks. Unsupervised Domain Adaptation results were observed to be close to the Fully Supervised Domain Adaptation used in the traditional procedure of Fine-tuning pretrained networks.
\end{abstract}


\titlepgskip=-15pt

\maketitle

\newcommand{\currprop}{1.0}

\section{Introduction}
\label{sec:introduction}

\PARstart{R}{adiology} has been a useful tool for assessing health conditions since the last decades of the $19^{th}$ century, when X-Rays were first used for medical purposes. Since then, it has become an essential tool for detecting, diagnosing and treating medical issues. More recently, algorithms have been coupled with radiology imaging techniques and other medical information in order to provide second opinions to physicians via Computer-Aided Detection/Diagnosis (CAD) systems. In recent decades, Machine Learning algorithms were incorporated into more modern CAD systems, providing automatic methodologies for finding patterns in big data scenarios, improving the capabilities of human physicians.

During the last half decade traditional Machine Learning pipelines have been losing ground to integrated Deep Neural Networks (DNNs) that can be trained from end-to-end \cite{Litjens:2017}. DNNs can integrate both the steps of feature extraction and statistical inference over unstructured data, such as images, temporal signals or text. Deep Learning models for images usually are built upon some form of trainable convolutional operation \cite{Krizhevsky:2012}. Convolutional Neural Networks (CNNs) are the most popular architectures for supervised image classification in Computer Vision. Variations of CNNs can be found in both detection \cite{Girshick:2015,Ren:2017,He:2017} and segmentation \cite{Long:2015,Ronneberger:2015,Badrinarayanan:2017} models.

\begin{figure*}[h!]
    \centering
    \includegraphics[page=1, trim=0.25cm 0.25cm 0.25cm 0.25cm, clip, width=0.75\textwidth]{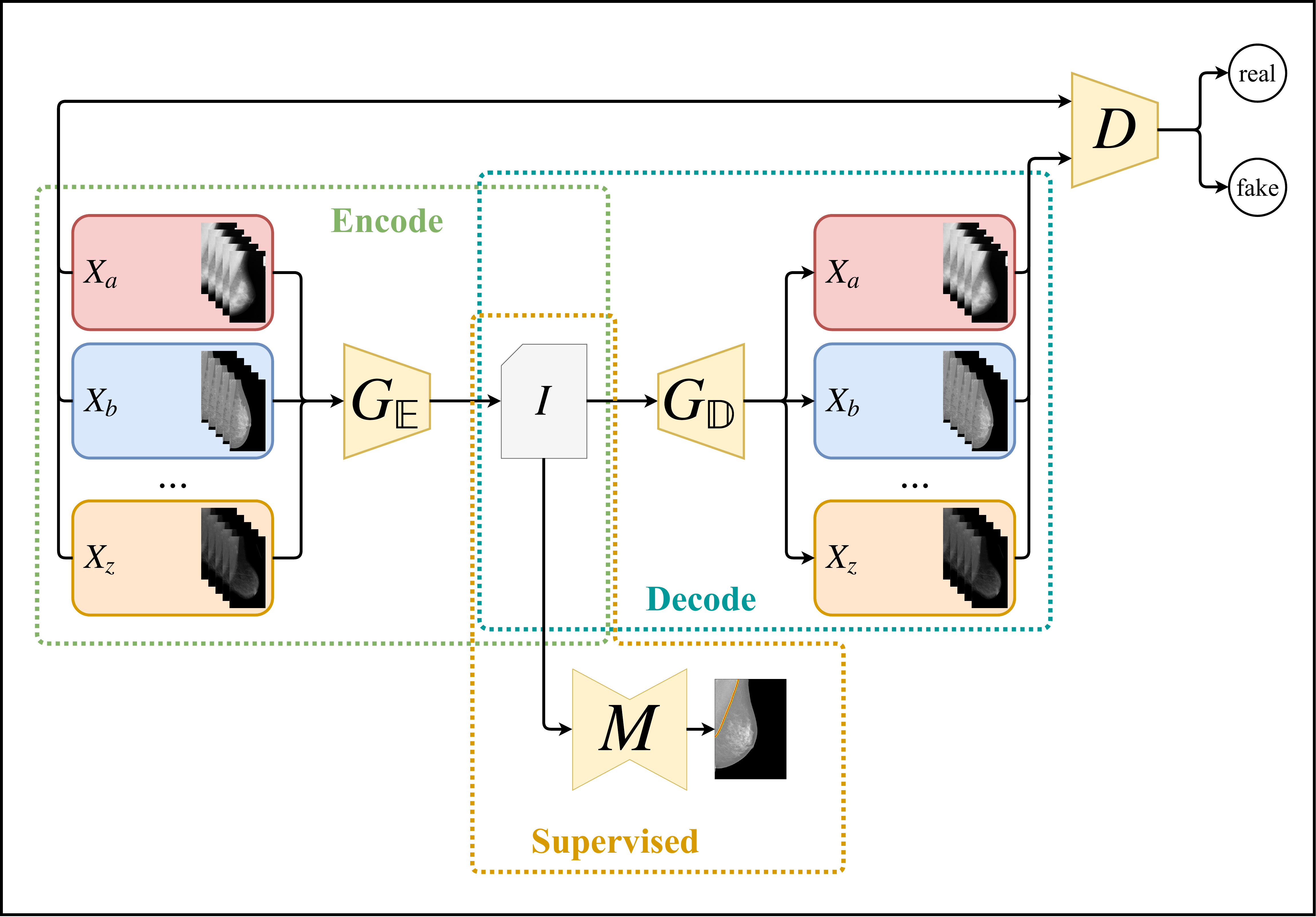}
    \caption{Simplified view of the CoDAGAN architecture for visual DA. A single $G$ network divided into encoder ($G_{\mathbb{E}}$) and decoder ($G_{\mathbb{D}}$) layers performs translations between the datasets. The discriminator $D$ evaluates if the fake images generated by $G$ according to the style of the target dataset are likely samples to have been drawn from the target distribution. A single supervised model $M$ is trained on the isomorphic representation $I$.}
    \label{fig:codagan_architecture}
\end{figure*}

In the medical community, labeled data is often limited and there are large amounts of unlabeled datasets that can be used for unsupervised learning. To make matters worse, the generalization of DNNs is normally limited by the variability of the training data, which is a major hamper, as different digitization techniques and devices used to acquire different datasets tend to produce biomedical images with distinct visual features \cite{Ghafoorian:2017}. Therefore, the study of methods that can use both labeled and unlabeled data is an active research area in both Computer Vision and Biomedical Image Processing.

Domain Adaptation (DA) \cite{Zhang:2017} methods are often used to improve the generalization of DNNs over images in a supervised manner. The most popular method for deep DA is Transfer Learning via Fine-Tuning pre-trained DNNs from larger datasets, such as ImageNet \cite{Deng:2009}. However, Fine-Tuning is a Fully Supervised Domain Adaptation (FSDA) method, only capable of learning from labeled data, ignoring the potentially larger amounts of unlabeled data available. Therefore, during the last years, several approaches have been proposed for Unsupervised Domain Adaptation (UDA) \cite{Zhang:2018,Liu:2016} and Semi-Supervised Domain Adaptation (SSDA) \cite{Wu:2016}. 

This paper proposes a Deep Learning-based DA method that works for the whole spectrum of UDA, SSDA and FSDA, being able to learn from both labeled and unlabeled data. An overview of the proposed approach, named Conditional Domain Adaptation Generative Adversarial Network (CoDAGAN), is presented in Figure~\ref{fig:codagan_architecture}.

\textbf{Contributions:} The main contribution of this work is a novel method that allows multiple datasets to be used conjointly in the training procedure. Most of the other modern methods in the visual DA literature \cite{Hoffman:2016,Hoffman:2018,Murez:2018,Wu:2018,Oliveira:2018} only allow for the pairwise training of one source and one target domain. In contrast to these pairwise methods, CoDAGANs learn to perform supervised inference over a common isomorphic representation built upon samples drawn from marginal distributions of multiple domains, as shown in Figure~\ref{fig:codagan_architecture}. In other words, CoDAGANs are able to take into account samples from distinct datasets with varying distributions drawn from the joint domain distribution to improve generalization.

Other sections in this paper are organized as follows. Section~\ref{sec:related} presents the previous works that paved the way for the proposal of CoDAGANs and gives an overview of visual DA. Section~\ref{sec:coda} describes the CoDAGAN modules, architecture, subroutines and loss components. Section~\ref{sec:setup} shows the experimental setup discussed in this paper, including datasets, hyperparameters, experimental protocol, evaluation metrics and baselines. Section~\ref{sec:results} introduces and discusses the results found during the exploratory tests of CoDAGANs for UDA, SSDA and FSDA in a quantitative and qualitative manner. At last, Section~\ref{sec:conclusion} finalizes this work with our final remarks and conclusions regarding the methods and experiments shown in this work. 
\section{Background and Related Work}
\label{sec:related}

This section presents recent developments in the literature regarding Semantic Segmentation (Section~\ref{sec:segmentation}), Adversarial Learning (Section~\ref{sec:gans}), Image Translation (Section~\ref{sec:image_translation}), Visual Domain Adaptation (Section~\ref{sec:domain_adaptation}) and Image Translation for DA (Section~\ref{sec:image_translation_da}). All of these works were crucial for the making of this paper.

\subsection{Deep Semantic Segmentation}
\label{sec:segmentation}

Since the resurgence of Neural Network technology as Deep Learning in the early 2010's, they have been adapted to perform segmentation tasks. Initial approaches used traditional CNNs \cite{Krizhevsky:2012} for image classification in dense labeling tasks by applying them to image patches. Patch-based approaches were observed to be unreasonably slow in these tasks, resulting in the proposal of Fully Convolutional Networks (FCNs) \cite{Long:2015}, which provided an end-to-end pipeline deep segmentation. These networks have similar architectures to traditional CNNs, which allows for transferring the knowledge from sparse to dense labeling scenarios. All pretrained convolutional layers in a CNN can be repurposed in an FCN architecture by replacing fully connected layers with additional trainable convolutions, which are then trained from scratch. FCNs also introduced the concept of skip connections, allowing for the fusion of low-level pixel activations with high semantic level information, yielding better fine-grained segmentations. One should notice that FCNs use the same loss functions as CNNs for image classification, as dense labeling can be seen as a collection of sparse labels for each pixel in an image. Therefore, Cross Entropy is the most common loss for supervised semantic segmentation and it can be expressed by:
\begin{equation}
    \mathcal{L}_{sup}(Y, \hat{y}) = -Y \log{(\hat{y})} - (1 - Y)\log{(1 - \hat{y})} \text{,}
    \label{eq:cross_entropy}
\end{equation}
where $Y$ represents the pixel-wise semantic map and $\hat{y}$ the probabilities for each class for a given sample.

\begin{figure}[!t]
    \centering
    \renewcommand{\currprop}{\columnwidth}
    \includegraphics[trim=0.1cm 0.1cm 0.1cm 0.1cm,clip,width=\currprop]{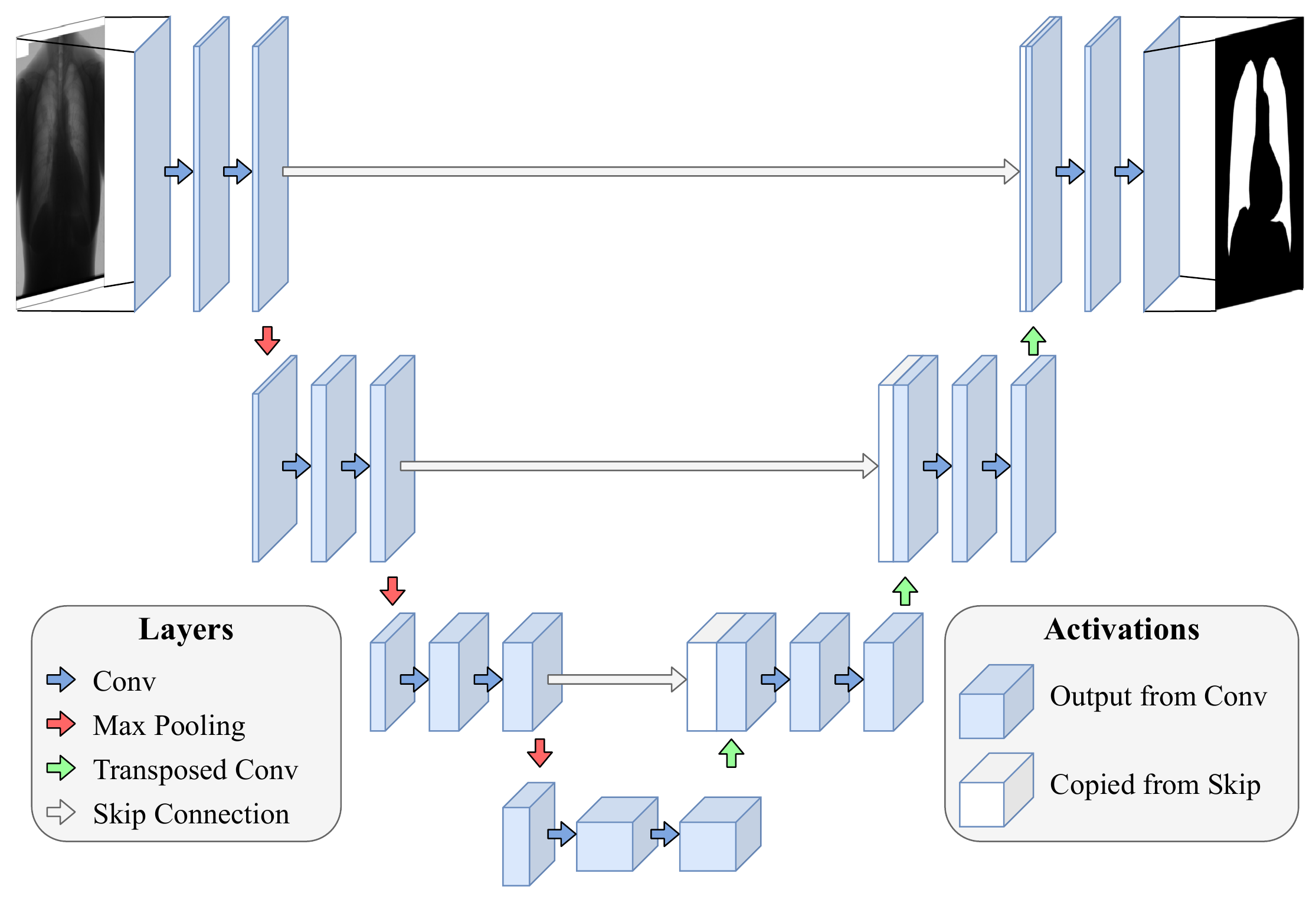}%
    \caption{Simplified U-Net architecture. Adapted from Ronneberger \textit{et al.} \cite{Ronneberger:2015}.}
    \label{fig:unet}
\end{figure}

Deconvolutional Networks \cite{Ronneberger:2015,Badrinarayanan:2017} based on transposed convolutions have grown in popularity in recent years, allowing for learnable upscaling of low spatial-resolution activations in the middle of the network. These are Encoder-Decoder networks, as they are often composed of downscaling convolution blocks followed by symmetric upscaling transposed convolutions. U-Nets \cite{Ronneberger:2015} built upon the idea of skip connections from FCNs, relying heavily on them for the decoding (upscaling) procedure. Most of these Encoder-Decoder networks \cite{Ronneberger:2015,Badrinarayanan:2017} are based on VGG architectures \cite{Simonyan:2014}, composed only of $3 \times 3$ convolutions that halve the spatial resolution of the feature map at each convolutional block. 
U-Nets rely heavily on Skip Connections, employing them on each Encoder/Decoder block pair, as shown in Figure~\ref{fig:unet}.

\subsection{Generative Adversarial Networks}
\label{sec:gans}

\begin{figure}[!t]
    \centering
    \renewcommand{\currprop}{0.35\textheight}
    \includegraphics[height=\currprop]{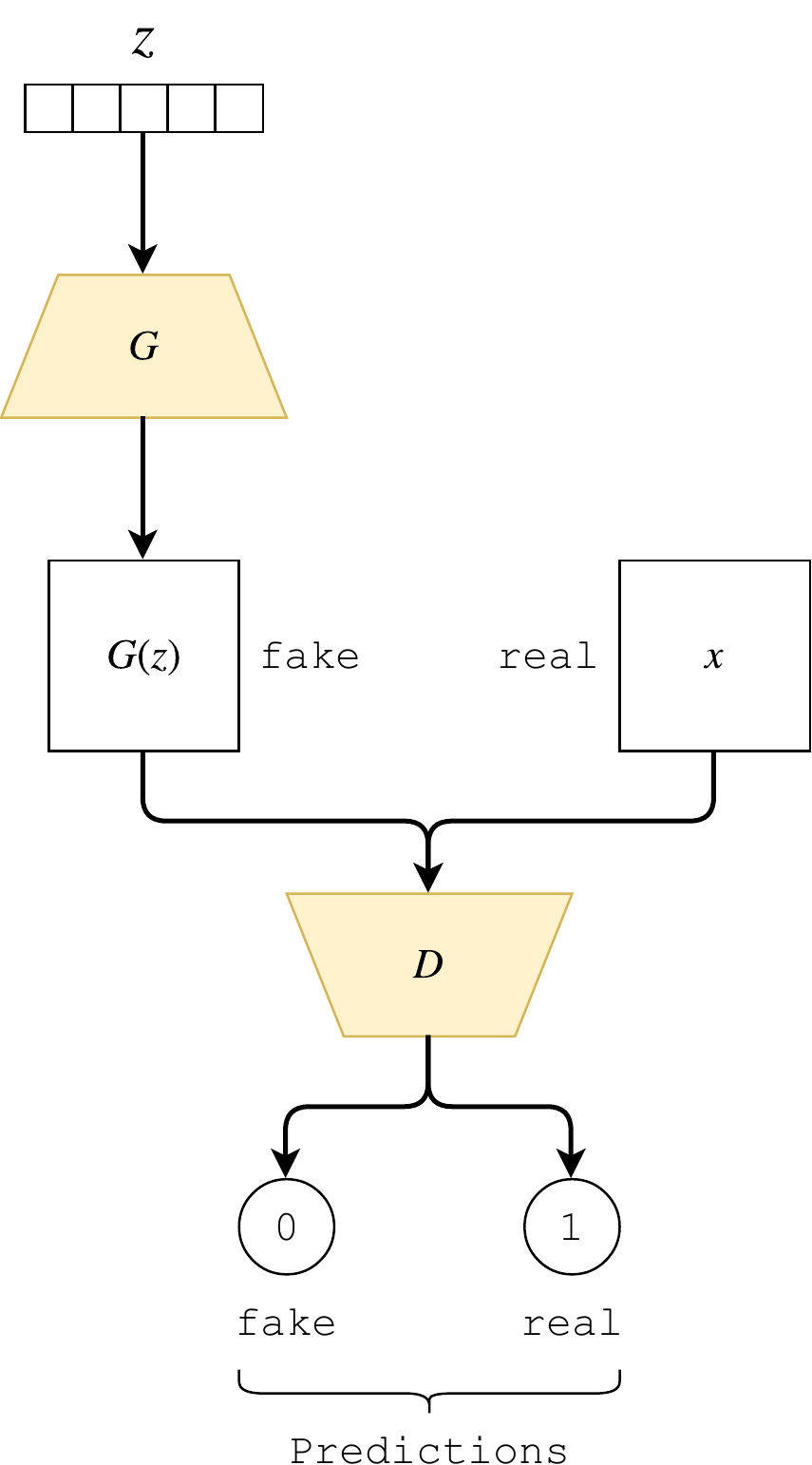}
    \caption{Traditional Architecture for Generative Adversarial Networks.}
    \label{fig:gans}
\end{figure}

Generative Adversarial Networks (GANs) \cite{Goodfellow:2014} have been an active and proliferous subject of research during the last years, being arguably the main go-to solution to deep generative modeling. Traditional GANs are composed of two networks trained conjointly: a generator ($G$) and a discriminator ($D$), as can be seen in Figure~\ref{fig:gans}.

$D$ is trained to correctly classify real samples $x \sim p_{x}$ drawn from the training dataset from fake samples $G(z) \sim p_{fake}$ created by the generator according to a random vector $z$ drawn from a noise distribution $p_{z}$. $G$ is trained to fool $D$ by approximating $p_{fake}$ from the true data distribution $p_{x}$. Loss functions for optimization for $D$ ($\mathcal{L}_{adv}^{D}$) and $G$ ($\mathcal{L}_{adv}^{G}$) are given by Equations~\ref{eq:gan_d} and~\ref{eq:gan_g}, respectively:
\begin{align*}
    \mathcal{L}_{adv}^{D} = -\mathbb{E}_{x}\left[ \log{(D(x))} \right] - \mathbb{E}_{z}\left[ \log{(1 - D(G(x)))} \right] \text{,} \numberthis
    \label{eq:gan_d}
\end{align*}
\begin{equation}
    \mathcal{L}_{adv}^{G} = -\mathcal{L}_{adv}^{D} \text{.} \numberthis
    \label{eq:gan_g}
\end{equation}
As objectives for $G$ and $D$ are exact opposites from each other, the networks can converge together. Training procedures for $G$ and $D$ are executed intermittently for a simultaneous convergence of the networks. Since this first proposal for adversarial learning of generative distributions, several advances have been made regarding training stability \cite{Goodfellow:2014,Mao:2017,Gulrajani:2017,Arjovsky:2017,Karras:2018}.

If trained properly, $G$ is able to receive new random vectors $z^{1}, z^{2}, ..., z^{N}$ and generate fake samples $G(z)^{1}, G(z)^{2}, ..., G(z)^{N}$ drawn from the approximate distribution $p_{fake}$. 
Later iterations on the research in generative modelling proposed changes in the architectures, input data and losses in order to adapt GANs to tasks such as deep convolutional architectures \cite{Radford:2015}, conditional training \cite{Mirza:2014}, unsupervised mapping of latent variables \cite{Chen:2016} and modeling joint distributions using only marginal samples -- as in Coupled GANs (CoGANs) \cite{Liu:2016}. 
Current state-of-the-art GANs are able to perform tasks as diverse as: 1) generating high resolution images with visual quality reasonably close to real ones \cite{Karras:2018}; 2) single-image superresolution \cite{Yuan:2018}; and 3) image-to-image translation between different domains \cite{Isola:2017, Zhu:2017:Cycle, Liu:2017, Huang:2018}. Some of these tasks have been more recently tackled by Variational Autoencoder (VAE) \cite{Kingma:2013} architectures as well, both alone and conjointly with GANs \cite{Liu:2017, Zhu:2017:Bicycle, Huang:2018}.

\subsection{Image-to-Image Translation}
\label{sec:image_translation}

Image-to-Image Translation Networks are Generative Adversarial Networks (GANs) \cite{Goodfellow:2014} capable of transforming samples from one image domain into images from another. Access to paired images from the two domains simplifies the learning process considerably, as losses can be devised using only pixel-level or patch-level comparisons between the original and translated images \cite{Isola:2017}. Paired Image-to-Image Translation can be achieved, therefore, by Conditional GANs (CGANs) \cite{Mirza:2014} coupled with simple regression models \cite{Chen:2017}. In order to achieve image translation, the adversarial components presented in Equations~\ref{eq:gan_d} and~\ref{eq:gan_g} are added to a paired regression loss $\mathcal{L}_{pair}(X_{A}^{(i)}, X_{B}^{(i)})$ between a pair of samples of index $i$ for datasets $X_{A}$ and $X_{B}$ from domains $A$ and $B$:
\begin{equation}
    \mathcal{L}_{pair}(X_{A}^{(i)}, X_{B}^{(i)})\ =\ \mathbb{E} \left\| X_{A}^{(i)} - G(X_{B}^{(i)}) \right\| \text{.}
    \label{eq:paired_regression_loss}
\end{equation}
This regression loss is usually the L1 loss, as it tends to produce less blurry results than the Mean Squared Error (MSE) loss \cite{Isola:2017}.

Requiring paired samples reduces the applicability of image-to-image translation to a very small and limited subset of image domains where there is the possibility of generating paired datasets. This limitation motivated the creation of Unpaired Image-to-Image Translation methods \cite{Zhu:2017:Cycle,Liu:2017,Huang:2018}. These networks are based on the concept of Cycle-Consistency, which models the translation process between two image domain as an invertible process represented by a cycle.

\begin{figure}[!t]
    \centering
    \subfloat[]{
        \includegraphics[page=1, trim=1.85cm 2.25cm 1.15cm 1.25cm, clip, width=0.95\columnwidth]{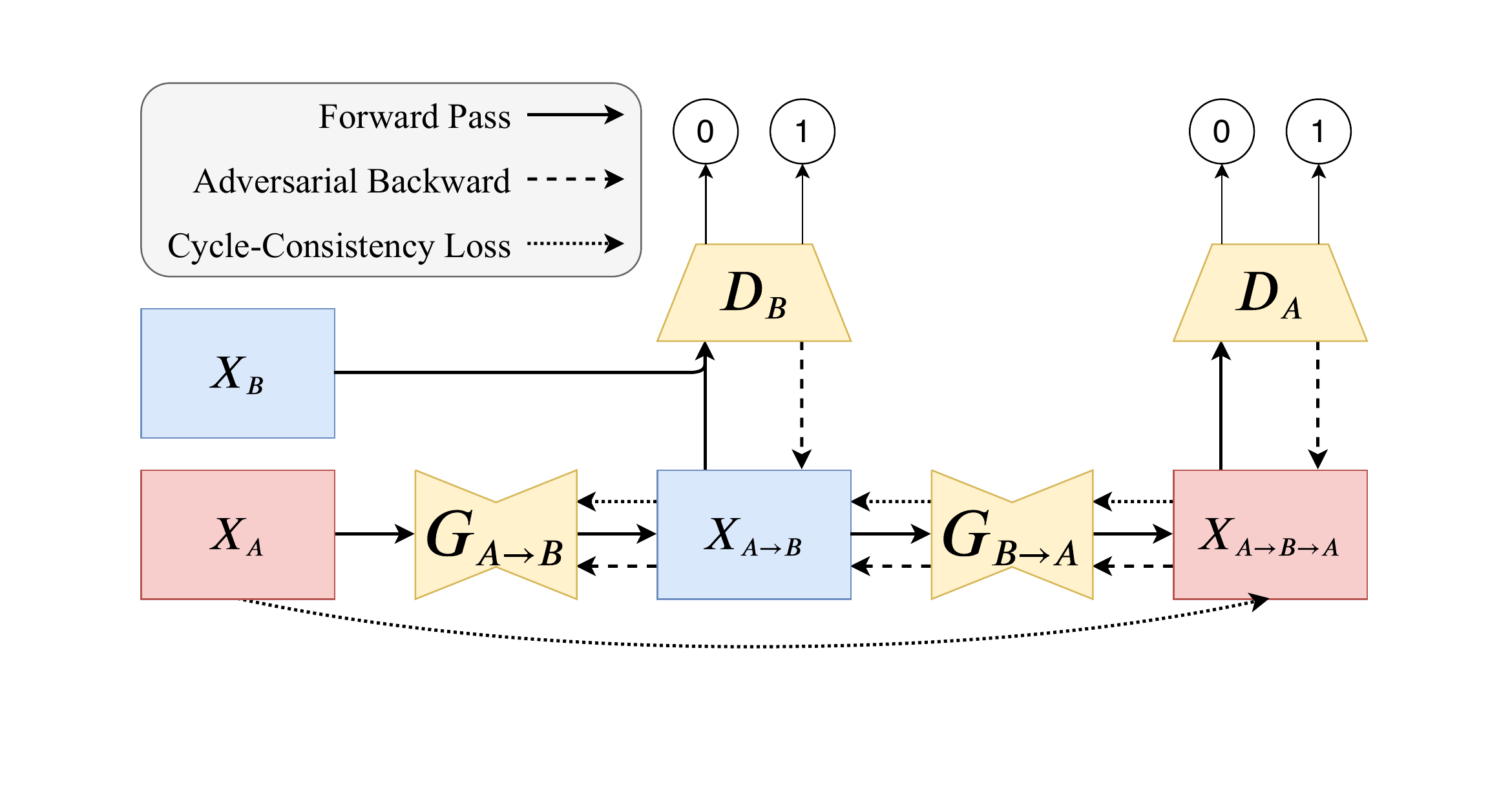}%
    }
    \hfil
    \subfloat[]{
        \includegraphics[page=1, trim=1.85cm 2.25cm 1.15cm 1.25cm, clip, width=0.95\columnwidth]{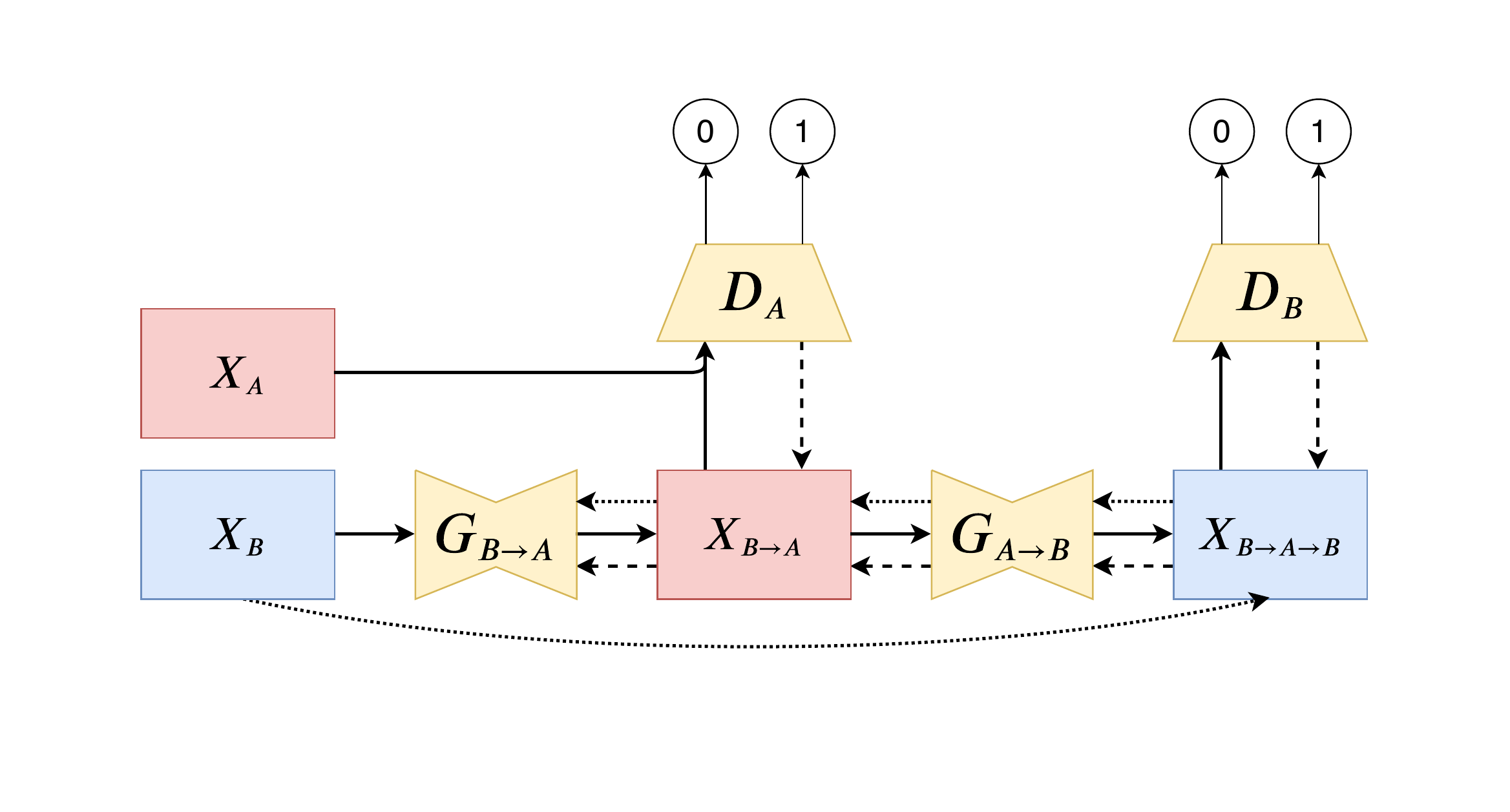}%
    }
    \caption{Example of GAN architecture based on Cycle Consistency. Traditionally, two generators ($G_{A \rightarrow B}$ and $G_{B \rightarrow A}$) and two discriminators ($D_{A}$ and $D_{B}$) are trained in order to achieve unsupervised image-to-image translation and an objective loss can be devised by comparing the pairs of samples $\{ X_{A}, X_{A \rightarrow B \rightarrow A} \}$ and $\{ X_{B}, X_{B \rightarrow A \rightarrow B} \}$.}
    \label{fig:cycle_consistency}
\end{figure}

These networks are based on the concept of Cycle-Consistency, which models the translation process between two image domain as an invertible process represented by a cycle, as can be seen in Figure~\ref{fig:cycle_consistency}. This cyclic structure allows for Cycle-Consistent losses to be used together with the adversarial loss components of traditional GANs.

A Cycle-Consistent loss can be formulated as follows: let $A$ and $B$ be two image domains containing unpaired image samples $X_{A}$ and $X_{B}$. Consider then two functions $G_{A \rightarrow B}$ and $G_{B \rightarrow A}$ that perform the translations $A \rightarrow B$ and $B \rightarrow A$ respectively. Then a loss $\mathcal{L}_{cyc}$ can be devised by comparing the pairs of images $\{ X_{A}, X_{A \rightarrow B \rightarrow A} \}$ and $\{ X_{B}, X_{B \rightarrow A \rightarrow B} \}$. In other words, the relations $X_{A} \approx G_{B \rightarrow A}(G_{A \rightarrow B}(X_{A}))$ and $X_{B} \approx G_{A \rightarrow B}(G_{B \rightarrow A}(X_{B}))$ should be maintained in the translation process. The counterparts of the generative networks in GANs are discriminative networks, which are trained to identify if an image is natural from the domain or translated samples originally from other domains. $D_{A}$ and $D_{B}$ are referred to as the discriminative networks for datasets $A$ and $B$, respectively. Discriminative networks are normally traditional supervised networks, such as CNNs \cite{Krizhevsky:2012,Simonyan:2014}, which are trained in the classification task of distinguishing real images from fake images generated by the generators. The loss $\mathcal{L}_{cyc}$ is usually the same L1 regression loss $\mathcal{L}_{pair}$ used in the Paired Image-to-Image Translation, but due to the lack of paired $X_{A}^{(i)}$ and $X_{B}^{(i)}$ samples, the regression is computed instead using the reconstruction $X_{A \rightarrow B \rightarrow A}^{(i)}$:
\begin{equation}
    \mathcal{L}_{cyc}(X_{A}^{(i)}, X_{A\rightarrow B \rightarrow A}^{(i)})\ =\ \mathbb{E} \left\| X_{A}^{(i)} - X_{A\rightarrow B \rightarrow A}^{(i)} \right\| \text{.}
    \label{eq:unpaired_regression_loss}
\end{equation}
The case for translations $B \rightarrow A \rightarrow B$ is analogous to the case of $A \rightarrow B \rightarrow A$.

Some efforts have been spent in proposing Unpaired Image Translation GANs for multi-domain scenarios, as the case of StarGANs \cite{Choi:2017}, but these networks do not explicitly present isomorphic representations of the data, as UNIT and MUNIT architectures do. Other advantages of UNIT and MUNIT over StarGANs is that they also compute reconstruction losses on the isomorphic representations, beside the traditional Cycle-Consistency between real and reconstructed images. CoDAGANs were built to be agnostic to the image translation network used as basis for the implementation, being able to transform any Image Translation GAN that has an isomorphic representation of the data into a multi-domain architecture with only minor changes to the generator and discriminator networks.


\subsection{Visual Domain Adaptation}
\label{sec:domain_adaptation}

DNNs often require a large amount of labeled training data in order to converge properly for performing supervised tasks in visual domains, such as classification \cite{Krizhevsky:2012}, detection \cite{Ren:2017} and segmentation \cite{Long:2015,Ronneberger:2015,Badrinarayanan:2017}. Due to this hunger for data, Transfer Learning has become a common procedure and received unprecedented attention in the realm of Deep Learning research, mainly using fine-tuning for adapting DNNs pretrained in larger datasets to perform similar tasks in smaller datasets. The larger set is usually a massive database, such as ImageNet \cite{Deng:2009} and is called the source dataset, while the smaller set is called the target dataset, being composed of the samples from the domain upon which inference will be performed.

\begin{figure*}[!t]
    \centering
    \renewcommand{\currprop}{0.30\textwidth}
    \subfloat[]{
        \includegraphics[width=\currprop]{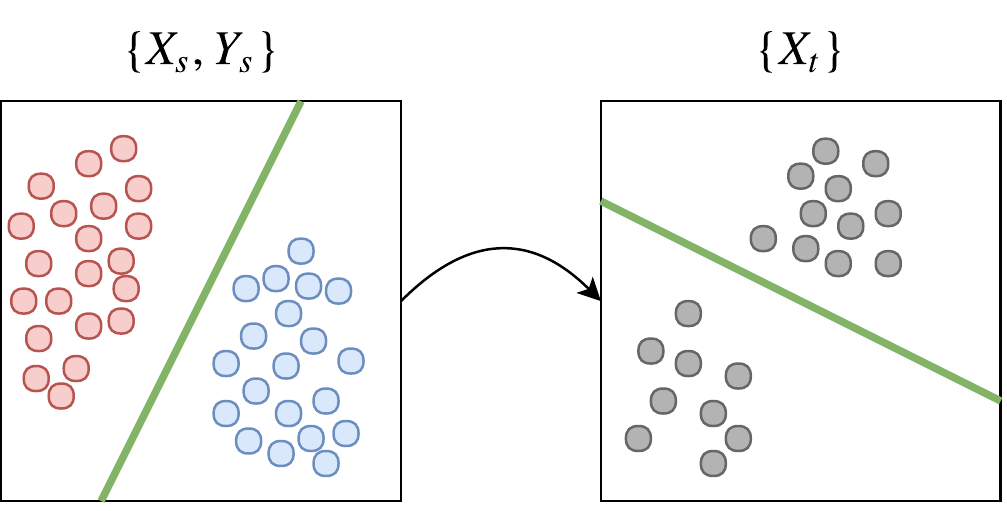}%
        \label{fig:domain_adaptation_uda}
    }
    \hfill
    \subfloat[]{
        \includegraphics[width=\currprop]{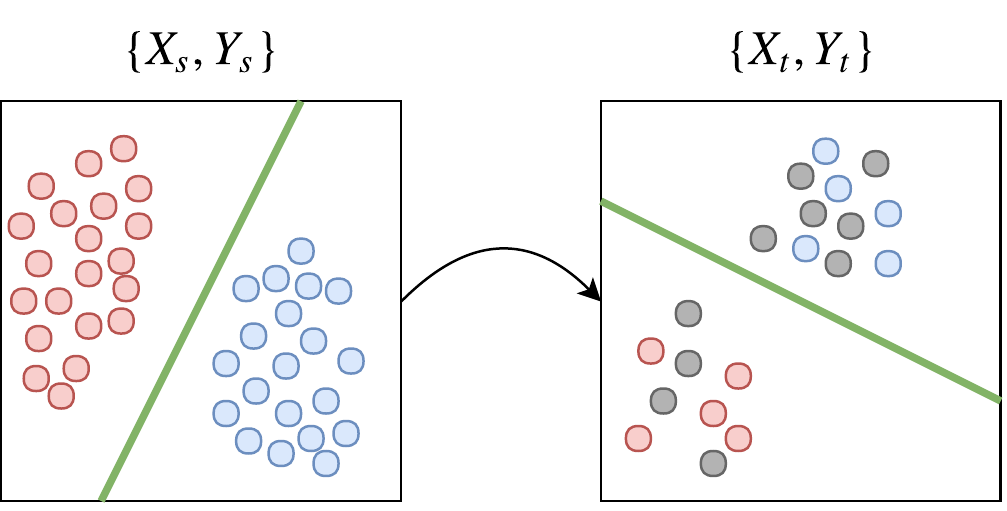}%
        \label{fig:domain_adaptation_ssda}
    }
    \hfill
    \subfloat[]{
        \includegraphics[width=\currprop]{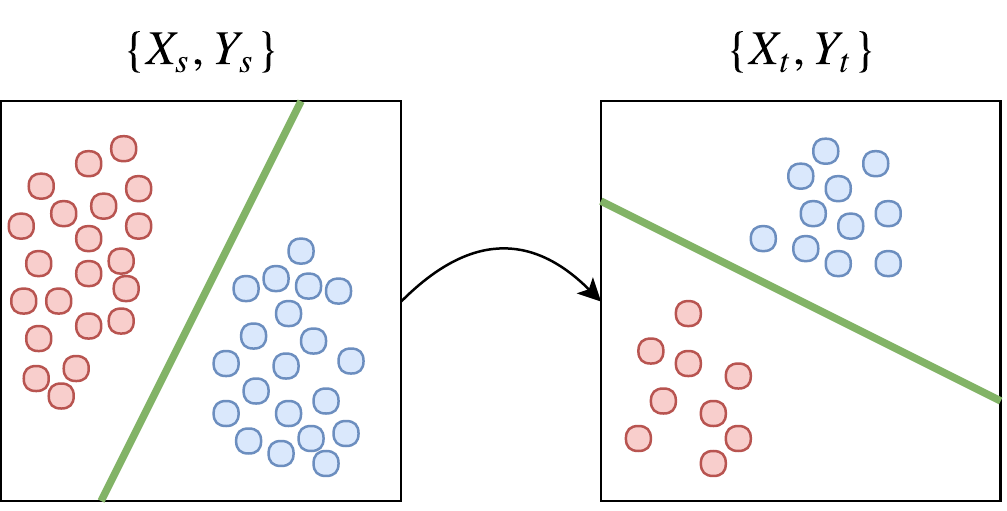}%
        \label{fig:domain_adaptation_fsda}
    }
    \caption{Examples of UDA (a), SSDA (b) and FSDA (c) in a classification scenario. $X_{s}$ and $Y_{s}$ are, respectively, the source dataset data and labels, while $X_{t}$ and $Y_{t}$ represent the target dataset data and labels. The green line shows a decision boundary between the classes in the dataset.}
    \label{fig:domain_adaptation}
\end{figure*}

Domain Adaptation can be done for fully labeled (FSDA), partially labeled (SSDA) and unlabeled (UDA) datasets. In UDA scenarios, no labels $Y_{t}$ are available for the target set, while SSDA tasks have both labeled and unlabeled samples on the target domain. FSDA contains only labeled data in the target domain and it is the most common practice nowadays among deep DA methods due to the simplicity of fine-tuning pretrained DNNs to perform new tasks. Computer Vision-related domains have a lot to benefit from fine-tuning, as most off-the-shelf large labeled datasets are from competitions for traditional Computer Vision tasks \cite{Deng:2009,Everingham:2015,Lin:2014}. Examples of UDA, SSDA and FSDA can be seen in Figure~\ref{fig:domain_adaptation}.

Zhang \textit{et al.} \cite{Zhang:2017} describes a taxonomy for DA tasks comprising most of the spectrum of deep and shallow knowledge transfer techniques. This taxonomy describes several classes of problems with variations in feature and label spaces between source and target domains, data labeling, balanced/unbalanced data and sequential/non-sequential data. CoDAGANs cannot be put in one single category in Zhang \textit{et al.}'s \cite{Zhang:2017} taxonomy, as they allow for a dataset to be source and target at the same time and are suitable for UDA, SSDA and FSDA, being able to learn from both unsupervised and supervised data. CoDAGANs can also be seen as a Domain Generalization method, as using distinct data sources leads to more generalizable models.

Several other surveys \cite{Patel:2015,Shao:2015,Csurka:2017,Wang:2018} specifically conducted on Visual DA methods assess that UDA and SSDA have been successfully achieved for image classification, mainly through the use of distribution alignment methods \cite{Geng:2011,Chen:2018,Li:2018}.
However, as discussed in all these surveys, UDA algorithms for segmentation, detection, reconstruction, and tracking still are in their infancy. As far as the authors are aware, the few proposed approaches for deep DA in dense labeling tasks have been tackled mainly using synthetic data for specific problems, such as outdoor scene segmentation \cite{Chen:2016,Hoffman:2016,Murez:2018,Wu:2018,Zhang:2019}, depth estimation \cite{Bousmalis:2017}. 

One of the most influential of these methods was proposed by Hoffman \textit{et al.} \cite{Hoffman:2016}, henceforth referred to as FCNs in the Wild. FCNs in the Wild adapt the traditional Fully Convolutional Networks \cite{Long:2015} for performing semantic segmentation in the same tasks -- outdoor scene segmentation, as in the original paper -- but in distinct datasets from the training ones, such as images from cities in other parts of the world and/or in other seasons. This approach uses a joint objective function $\mathcal{L}$ composed of a supervised segmentation loss $\mathcal{L}_{seg}$ in the source domain $\mathcal{S}$, an adversarial component $\mathcal{L}_{da}$ to optimize the domain alignment between $\mathcal{S}$ and the target $\mathcal{T}$ and, a multiple instance loss $\mathcal{L}_{mi}$ based on per-sample class histograms computed in $\mathcal{S}$. FCNs in the Wild report results in the synthetic GTA5 \cite{Richter:2016} and SYNTHIA \cite{Ros:2016} datasets -- used as source data -- and in the real-world CityScapes \cite{Cordts:2016} and the newly introduced Berkeley Deep Driving Segmentation (BDDS) \cite{Hoffman:2016} dataset as targets. DA results from GTA5$\rightarrow$CityScapes achieve a mean Intersection over Union (mIoU) value of 27.1\%, while SYNTHIA$\rightarrow$Cityscapes adaptation yields 17.0\%.

Most deep approaches for UDA and SSDA \cite{Long:2017,Li:2018} rely on distribution matching to perform knowledge transfer by comparing the statistical moments of the two domains, mainly in variations of the Maximum Mean Discrepancy (MMD) \cite{Borgwardt:2006} metric. MMD-based approaches are a kind of Feature Representation Learning, which only takes into account feature space, ignoring label space, thus, it's fully unsupervised. 
MMD methods are the most common approaches for matching distinct distributions in deep image classification tasks, but these approaches have not been tested on dense labeling scenarios.


Traditional DA techniques perform knowledge transfer between a single pair of datasets: a source and a target datasets. In many cases it is advantageous to acquire as much data as possible from multiple sources, mainly when there is a lack of labels. Multi-source methods \cite{Sun:2011,Gong:2013,Fang:2013,Caseiro:2015,Ming:2015} try to infer a joint probability distribution $p_{X_{1}, X_{2}, ..., X_{N}}$ from a multitude of source data $X_{1}, X_{2}, ..., X_{N}$, each one with its own marginal probability distribution $p_{X_{1}}, p_{X_{2}}, ..., p_{X_{N}}$. These methods must infer joint distributions for the domains based only on the marginal distributions of the source data. 

\subsection{Image Translation for Domain Adaptation}
\label{sec:image_translation_da}

Since the introduction of Image-to-Image Translation GANs, several works \cite{Liu:2016,Bousmalis:2017,Hoffman:2018,Oliveira:2018,Murez:2018,Wu:2018,Zou:2018} have used these architectures to perform Domain Adaptation between image domains. In the following paragraphs, when available, we will mainly focus on the experiments of the literature in dense labeling tasks.

As far as the authors are aware, the first use of Image-to-Image Translation for Domain Adaptation purposes was shown by CoGANs \cite{Liu:2016}. This work showed UDA for digit classification between the MNIST \cite{Lecun:1998} and USPS \cite{Hull:1994} datasets. While MNIST contains well-behaved, preprocessed and high-contrast handwritten digit samples, USPS better mimics a real-world scenario for digit classification. Thus, being able to adapt a digit classifier from MNIST to USPS without using labels from the target set is a challenging problem. One should notice that CoGANs still did not present UDA results in dense labeling tasks.

\begin{figure}[!t]
    \centering
    \subfloat[]{
        \includegraphics[page=1, trim=1.85cm 0.95cm 1.15cm 1.25cm, clip, width=0.95\columnwidth]{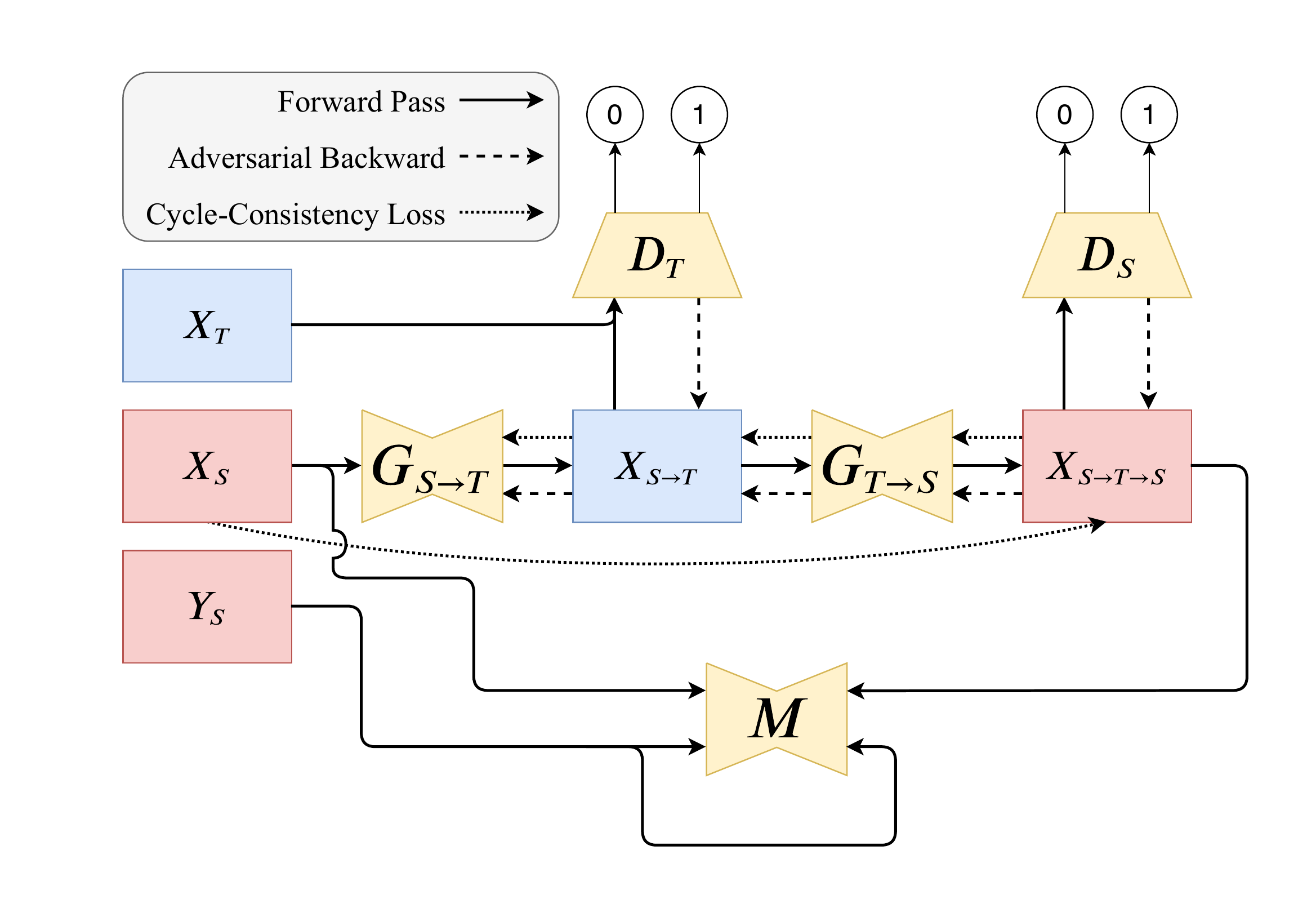}%
        \label{fig:cycle_consistency_da_aba}
    }
    \hfil
    \subfloat[]{
        \includegraphics[page=1, trim=1.85cm 0.95cm 1.15cm 1.25cm, clip, width=0.95\columnwidth]{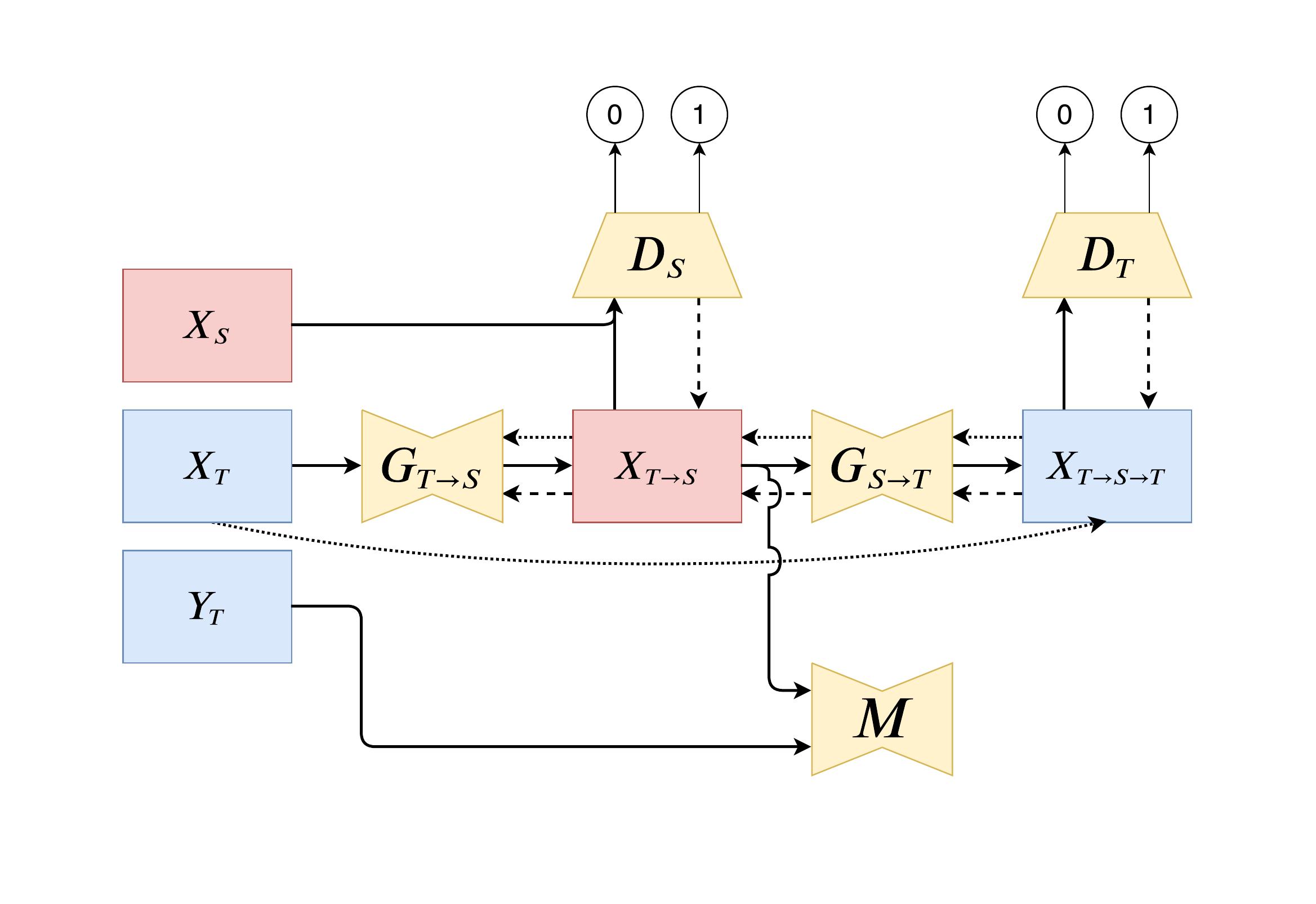}%
        \label{fig:cycle_consistency_da_bab}
    }
    \caption{Simplified scheme for a Cycle-Consistency DA approach in translations $S \rightarrow T \rightarrow S$ (a) and $T \rightarrow S \rightarrow T$ (b). A supervised model $M$ performs the supervised learning by using the labels $Y_{S}$ in the source domain $S$ and, in the case of FSDA or SSDA, also using the target domain labels $Y_{T}$, when available.}
    \label{fig:cycle_consistency_da}
\end{figure}

Cycle-Consistent Adversarial Domain Adaptation (CyCADA) \cite{Hoffman:2018} was built upon CycleGANs to perform UDA in dense labeling tasks -- more specifically semantic segmentation. As most other papers in the area, CyCADA relies on synthetic data from realistic 3D simulations such as third person games to acquire labeled data for outdoor scene classification. It is much less time-consuming to annotate synthetic images from these simulations in an automated or semi-automated manner than to label entire datasets from scratch with pixel-level annotations, such as Pascal VOC \cite{Everingham:2015}. CyCADA achieves UDA by attaching an FCN to the end of a CycleGAN, as shown in Figure~\ref{fig:cycle_consistency_da}, limiting it to adapting between a pair of source and target domains $\left\{\mathcal{S}, \mathcal{T}\right\}$. One should notice in this architecture that in the case of total lack of target labels $Y_{T}$ -- that is, in a UDA scenario -- semantic consistency gradients are successfully fed to $G_{T \rightarrow S}$ due to its proximity to $M$, but very small gradient intensities flow from $M$ to $G_{T \rightarrow S}$ in $S \rightarrow T \rightarrow S$ (Figure~\ref{fig:cycle_consistency_da_aba}). This might represent an imbalance in the training of $G_{S \rightarrow T}$ and $G_{T \rightarrow S}$, which is not desirable for DA.

The final loss for CyCADA ($\mathcal{L}_{CyCADA}$) is composed of a Cycle-Consistency loss $\mathcal{L}_{cyc}$, adversarial loss component $\mathcal{L}^{G}_{adv}$ for the pair of generators, adversarial loss component $\mathcal{L}^{D}_{adv}$ for the pair of discriminators and a supervised Semantic Consistency loss $\mathcal{L}_{sem}$. CyCADA reports successful UDA results between the synthetic GTA5 \cite{Richter:2016} dataset and the real-world CityScapes dataset \cite{Cordts:2016}. CyCADA reports mIoU results of 35.4\%, frequency weighted Intersection over Union (fwIoU) of 73.8\% and Pixel Accuracy of 83.6\% in translations between GTA5$\rightarrow$CityScapes.


Similarly to CyCADA, I2IAdapt \cite{Murez:2018} uses CycleGANs coupled with segmentation architectures to perform UDA for dense labeling tasks. Again the GTA5 and CityScapes datasets are used as source and target data in I2IAdapt, comparing the results with simply testing the pretrained DNN in the target domain, yielding considerable improvements. Their best configuration with a DenseNet \cite{Huang:2017} backbone achieves 35.7\% of mIoU on CityScapes.

The Dual Channel-wise Alignment Network (DCAN) \cite{Wu:2018} also follows close architectural choices to CyCADA and I2IAdapt, attaching a segmentation architecture to the target end of a translation architecture. DCAN was trained on two synthetic datasets (GTA5 and SYNTHIA \cite{Ros:2016}) and in one real-world dataset (CityScapes). Wu \textit{et al.} \cite{Wu:2018} report mIoU values of 38.9\% for GTA5$\rightarrow$CityScapes and 41.7\% for SYNTHIA$\rightarrow$CityScapes, surpassing the baselines by between 8\% and 9\% and other similar methods by a small percentage.

Oliveira \textit{et al.} \cite{Oliveira:2018} used Unpaired Image-to-Image Translation to perform UDA, SSDA and FSDA between CXR datasets. As the previously mentioned approaches \cite{Hoffman:2018,Murez:2018,Wu:2018}, Oliveira \textit{et al.}'s approach is only able to perform DA between a single pair of domains due to the supervised DNN being added to one of the ends of a Cycle Consistent GAN. Oliveira \textit{et al.} \cite{Oliveira:2018} report Jaccard results in the Montgomery dataset \cite{Jaeger:2014} ranging from 88.2\% in the UDA scenario to of 93.18\% in the FSDA scenario, surpassing both Fine-Tuning and From Scratch training in scarce label scenarios.


As shown, most methods simply combine the supervised learning from classification or segmentation schemes with a supervised or unsupervised image translation architecture to perform UDA, attaching an FCN-like architecture at one end of the image translation. From now on these models will be referenced as Domain-to-Domain (D2D) methods due to their limitations in allowing only pairwise training. CoDAGANs apply a similar framework to D2D in order to perform UDA, SSDA and FSDA, mixing the unsupervised learning of Cycle-Consistent GANs with the supervised pixel-wise learning of an Encoder-Decoder architecture. Two crucial distinctions between D2D methods and CoDAGANs must be addressed, though: 1) only one Encoder, one Decoder and one Discriminator are used in the image translation process, as different domains are recognized by $G$ and $D$ via One-Hot Encoding, allowing for multi-target domain adaptation; 2) supervised learning is performed only on the bottleneck of $G$, not in end of the translation process, allowing all domains to share a single isomorphic space $I$. These differences allow for drawing supervised and unsupervised knowledge from several distinct datasets, depending on their label availability.
\section{Proposed Method: CoDAGANs}
\label{sec:coda}

\begin{figure*}[ht]
    \centering
    \renewcommand{\currprop}{0.95\textwidth}
    \includegraphics[page=1, clip, trim=1.5cm 1.5cm 1.5cm 1.5cm, width=\currprop]{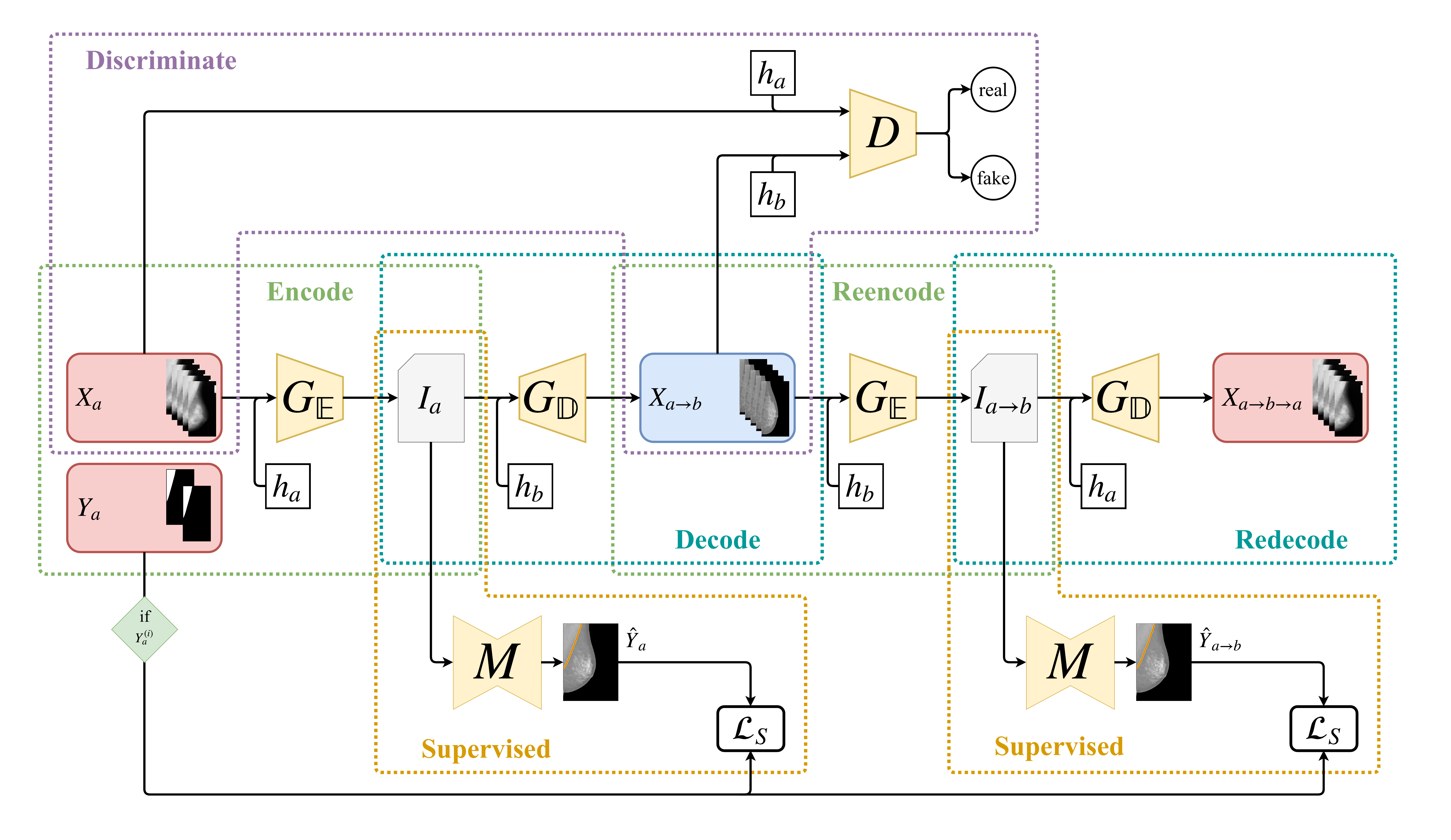}
    \caption{Training procedure for CoDAGANs. This figure exemplifies a translation $a \rightarrow b \rightarrow a$, but the translation $b \rightarrow a \rightarrow b$ -- which is performed simultaneously to the procedure for $a \rightarrow b \rightarrow a$ -- is analogous. Notice that the reconstruction losses are omitted from this view of our architecture for simplification. 
    }
    \label{fig:codagan_training}
\end{figure*}

CoDAGANs combine unsupervised and supervised learning to perform UDA, SSDA or FSDA between two or more image sets. These architectures are based on adaptations of preexisting Unsupervised Image-to-Image Translation networks \cite{Zhu:2017:Cycle,Liu:2017,Huang:2018}, adding supervision to the process in order to perform Transfer Learning. The generator networks ($G$) in Image Translation GANs are implemented usually using Encoder-Decoder architectures as U-Nets \cite{Ronneberger:2015}. At the end of the Encoder ($G_{\mathbb{E}}$) there is a middle-level representation $I$ that can be trained to be isomorphic in these architectures. $I$ serves as input of the Decoder ($G_{\mathbb{D}}$). Isomorphism allows for learning a supervised model $M$ based on $I$ that is capable of inferring over several datasets. This unsupervised translation process followed by a supervised learning model can be seen in Figure~\ref{fig:codagan_training}.

For this work we employed the Unsupervised Image-to-Image Translation Network (UNIT) and Multimodal Unsupervised Image-to-Image Translation (MUNIT) Network as a basis for the generation of $I$. On top of that, we added the supervised model $M$ -- which is based on a U-net \cite{Ronneberger:2015} -- and made some considerable changes to the translation approaches, mainly regarding the architecture and conditional distribution modelling of the original GANs, as discussed in Section~\ref{sec:conditional}. The exact architecture for $G$ depends on the basis translation network chosen for the adaptation. In our case, both UNIT and MUNIT use VAE-like architectures \cite{Kingma:2013} for $G$, containing downsampling ($G_{\mathbb{E}}$), upsampling ($G_{\mathbb{D}}$) and residual layers.

The shape of $I$ depends on the architecture choice for $G$. UNIT, for example, assumes a single latent space between the image domains, while MUNIT separates the content of an image from its style. CoDAGANs feeds the whole latent space to the supervised model when it is based on UNIT and only content information when it is built upon MUNIT, as the style vector has no spatial resolution and as we intend to ignore style and preserve content.

A training iteration on a CoDAGAN follows the sequence presented in Figure~\ref{fig:codagan_training}. The generator network $G$ -- such as U-nets \cite{Ronneberger:2015} and Variational Autoencoders \cite{Kingma:2013} -- is an Encoder-Decoder architecture. However, instead of mapping the input image into itself or into a semantic map as its Encoder-Decoder counterparts, it is capable of translating samples from one image dataset into synthetic samples from another dataset. The encoding half of this architecture ($G_{\mathbb{E}}$) receives images from the various datasets and creates an isomorphic representation somewhere between the image domains in a high dimensional space. This code will be henceforth described as $I$ and is expected to correlate important features in the domains in an unsupervised manner \cite{Liu:2016}. Decoders ($G_{\mathbb{D}}$) in CoDAGAN generators are able to read $I$ and produce synthetic images from the same domain or from other domains used in the learning process. This isomorphic representation is an integral part of both UNIT \cite{Liu:2017} and MUNIT \cite{Huang:2018} translations, as they also enforce good reconstructions for $I$ in the learning process. It also plays an essential role in CoDAGANs, as all supervised learning is performed on $I$.

As shown in Figure~\ref{fig:codagan_training}, CoDAGANs include five unsupervised subroutines: a) Encode, b) Decode, c) Reencode, d) Redecode and e) Discriminate; and two f) Supervision subroutines, which are the only labeled ones. These subroutines will be detailed further in the following paragraphs.

\paragraph{Encode} First, a pair of datasets $a$ (source) and $b$ (target) are randomly selected among the potentially large number of datasets used in training. A minibatch $X_{a}$ of images from $a$ is then appended to a code $h_{a}$ generated by a One Hot Encoding scheme, aiming to inform the encoder $G_{\mathbb{E}}$ of the samples' source dataset. The 2-uple $\{X_{a}, h_{a}\}$ is passed to the encoder $G_{\mathbb{E}}$, producing an intermediate isomorphic representation $I_{a}$ for the input $X_{a}$ according to the marginal distributions computed by $G_{\mathbb{E}}$ for dataset $a$.

\paragraph{Decode} The information flow is then split into two distinct branches: 1) $I_{a}$ is fed to the supervised model $M$; 2) $I_{a}$ is appended to a code $h_{b}$ and passed through the decoder $G_{\mathbb{D}}$ conditioned to dataset $b$. The function $G_{\mathbb{D}}(I_{a}, h_{b})$ produces $X_{a \rightarrow b}$, which is a translation of images in the minibatch $X_{a}$ with the style of dataset $b$.

\paragraph{Reencode} The Reencode procedure performs the same operation of generating an isomorphic representation as the Encode subroutine, but receiving as input the synthetic image $X_{a \rightarrow b}$. More specifically, the reencoded isomorphic representation $I_{a \rightarrow b}$ is generated by $G_{\mathbb{E}}(X_{a \rightarrow b}, h_{b})$.

\paragraph{Redecode} Again the architecture splits into two branches: 1) $I_{a \rightarrow b}$ is passed to $M$ in order to produce the prediction $\hat{Y}_{a \rightarrow b}$; 2) the isomorphic representation is decoded as in $G_{\mathbb{D}}(I_{a \rightarrow b}, h_{b})$, producing the reconstruction $X_{a \rightarrow b \rightarrow a}$, which can be compared to $X_{a}$ via a Cycle-Consistency loss $\mathcal{L}_{cyc}$ (Equation~\ref{eq:unpaired_regression_loss}).

\paragraph{Discriminate} At the end of Decode, the synthetic image $X_{a \rightarrow b}$ is produced. The original samples $X_{a}$ and the translated images $X_{a \rightarrow b}$ are merged in a single batch and passed to $D$, which uses the adversarial loss component $\mathcal{L}^{D}_{adv}$ (Equation~\ref{eq:gan_d}) in order to classify between real and synthetic samples. In Routines when the generators are being updated instead of the discriminators, the adversarial loss $\mathcal{L}^{G}_{adv}$ (Equation~\ref{eq:gan_g}) is computed instead.

\paragraph{Supervision} At the end of Encode and Reencode subroutines, for each sample $X_{a}^{(i)}$ which has a corresponding label $Y_{a}^{(i)}$, the isomorphisms $I_{a}^{(i)}$ and $I_{a \rightarrow b}^{(i)}$ are both fed to the same supervised model $M$. The model $M$ perform the desired supervised task, generating the predictions $\hat{Y}_{a}^{(i)}$ and $\hat{Y}_{a \rightarrow b}(i)$. Both these predictions can be compared in a supervised manner to $Y_{a}^{(i)}$ by using $\mathcal{L}_{S}$ (Equation~\ref{eq:cross_entropy}), if there are labels for the image $i$ in this minibatch. As there are always at least some labeled samples in this scenario, $M$ is trained to perform inference on isomorphic encodings of both originally labeled data ($M(I_{a}) = \hat{Y}_{a} \approx Y_{a}$) and data translated by the CoDAGAN for the style of other datasets ($M(I_{a \rightarrow b}) = \hat{Y}_{a \rightarrow b} \approx Y_{a}$). 

If domain shift is computed and adjusted properly during the training procedure, the properties $X_{a} \approx X_{a \rightarrow b \rightarrow a}$ and $I_{a} \approx I_{a \rightarrow b}$ are achieved, satisfying Cycle-Consistency and Isomorphism, respectively. After training, it does not matter which input dataset among the training ones is conditionally fed to $G_{\mathbb{E}}$ to the generation of isomorphism $I$, as samples from all datasets should all belong to the same joint distribution in $I$-space. Therefore any learning performed on $I_{a}$ and $I_{a \rightarrow b}$ is universal to all datasets used in the training procedure. Instead of performing only the translation $a \rightarrow b \rightarrow a$ for the randomly chosen datasets $a$ and $b$, all mentioned subroutines are run simultaneously for both $a \rightarrow b \rightarrow a$ and $b \rightarrow a \rightarrow b$, as in UNIT \cite{Liu:2017} and MUNIT \cite{Huang:2018}. Translations $b \rightarrow a \rightarrow b$ are analogous to the $a \rightarrow b \rightarrow a$ case described previously.

One should notice that $G_{\mathbb{E}}$ performs spatial downsample, while $G_{\mathbb{D}}$ performs upsample, consequently the model $M$ should take into account the amount of downsampling layers in $G_{\mathbb{E}}$. More specifically, we removed the first two layers of U-Net \cite{Ronneberger:2015} when using them as the model $M$, resulting in an asymmetrical U-Net to compensate for $G_{\mathbb{E}}$ downsamplings. The amount of input channels of $M$ must also be compatible with the amount of output channels in $G_{\mathbb{E}}$. Another constraint for the architecture of the pair $\{G_{\mathbb{E}}, G_{\mathbb{D}}\}$ is that the upsampling performed by $G_{\mathbb{D}}$ should always compensate the downsampling factor of $G_{\mathbb{E}}$, characterizing $G$ as a whole as a symmetric Encoder-Decoder network.

The discriminator $D$ for CoDAGANs is basically the same as the discriminator from the original Cycle-Consistency network, that is, a basic CNN that classifies between real and fake samples. The only addition to $D$ is conditional training in order for the discriminator to know the domain the sample is supposed to belong to. This allows $D$ to use its marginal distribution for each dataset for determining the likelihood of veracity for the sample. It is important to notice that our model is agnostic to the choice of Unsupervised Image-to-Image Translation architecture, therefore future advances in this area based on Cycle-Consistency should be equally portable to perform DA and further benefit CoDAGAN's performance.



\subsection{Conditional Dataset Encoding}
\label{sec:conditional}

Conditional dataset training allows CoDAGANs to process data and perform transfer from several distinct source/target datasets. Fully or partially labeled datasets act as source datasets for the method, while unlabeled data is used both to enforce isomorphism in $I$ and to yield adequate image translations between domains. Partially labeled and unlabeled data are, therefore, the target datasets for in this architecture.

\begin{figure}[ht!]
    \centering
    \includegraphics[page=1, trim=0.45cm 0cm 0cm 0cm, clip, width=\columnwidth]{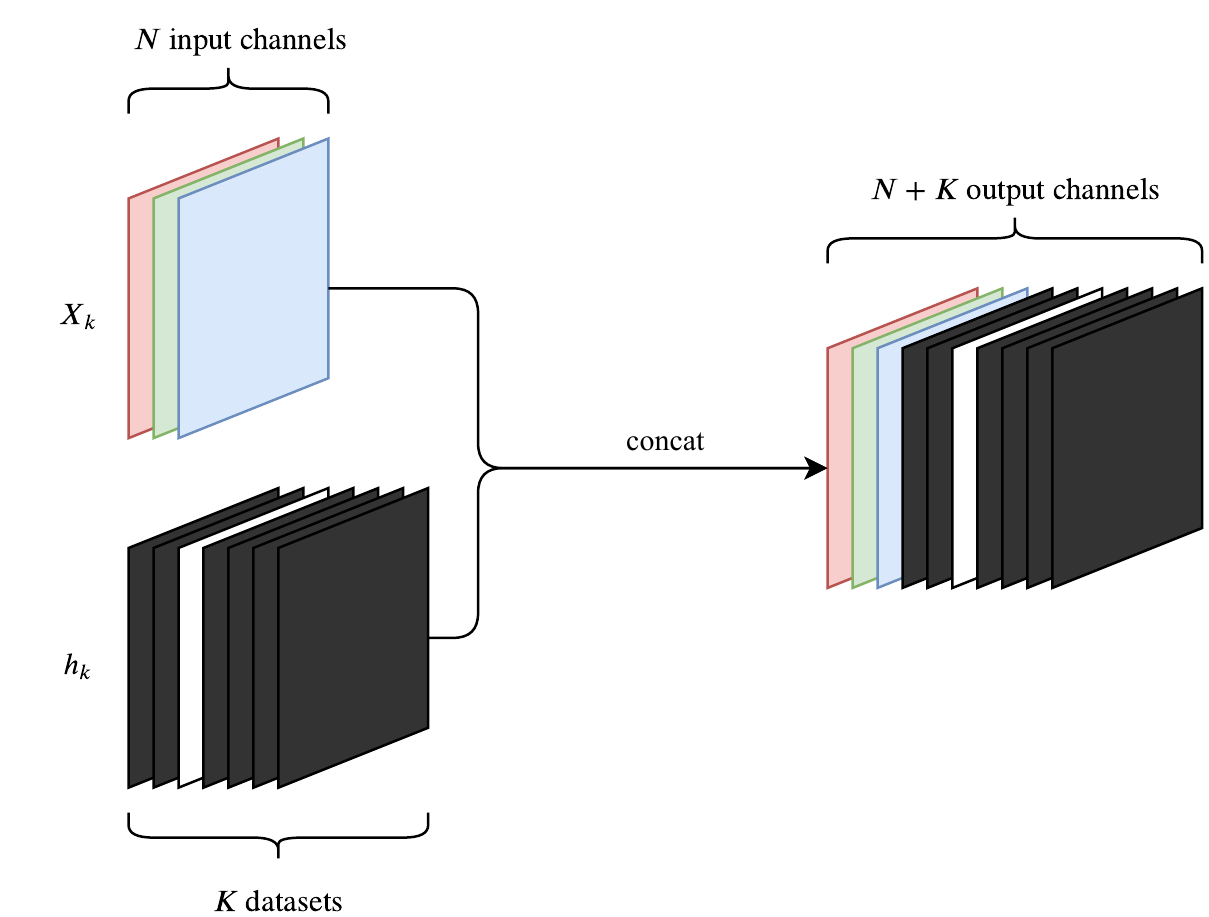} 
    \caption{Illustration of One-Hot-Encoding on image channels in order to encode dataset information.}
    \label{fig:one_hot}
\end{figure}

\begin{figure}[!t]
    \centering
    \renewcommand{\currprop}{0.95\columnwidth}
    \subfloat[]{
        \includegraphics[page=1, clip, trim=1.3cm 8cm 1.8cm 8cm, width=\currprop]{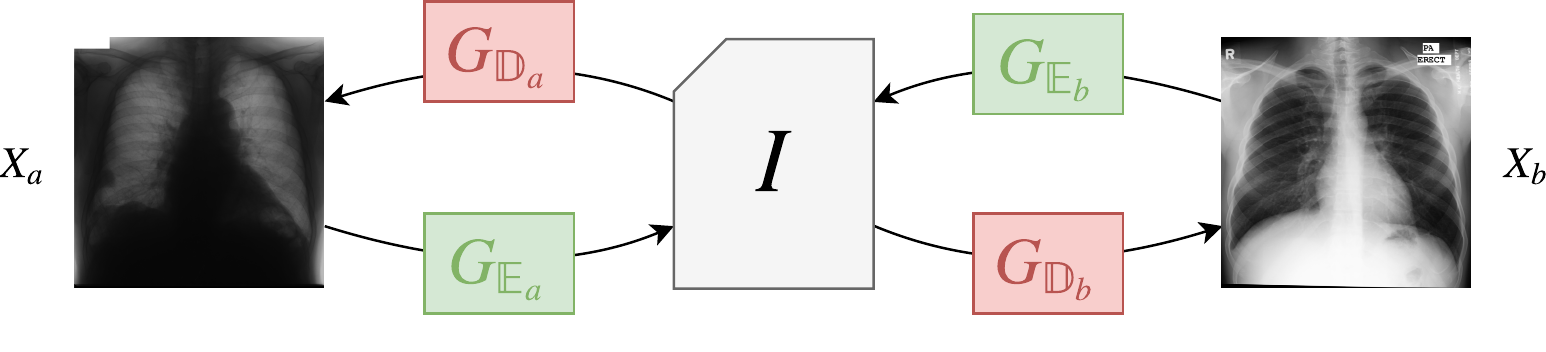}%
    }
    \hfil
    \subfloat[]{
        \includegraphics[page=1, clip, trim=1.3cm 1.5cm 1.8cm 1.5cm, width=\currprop]{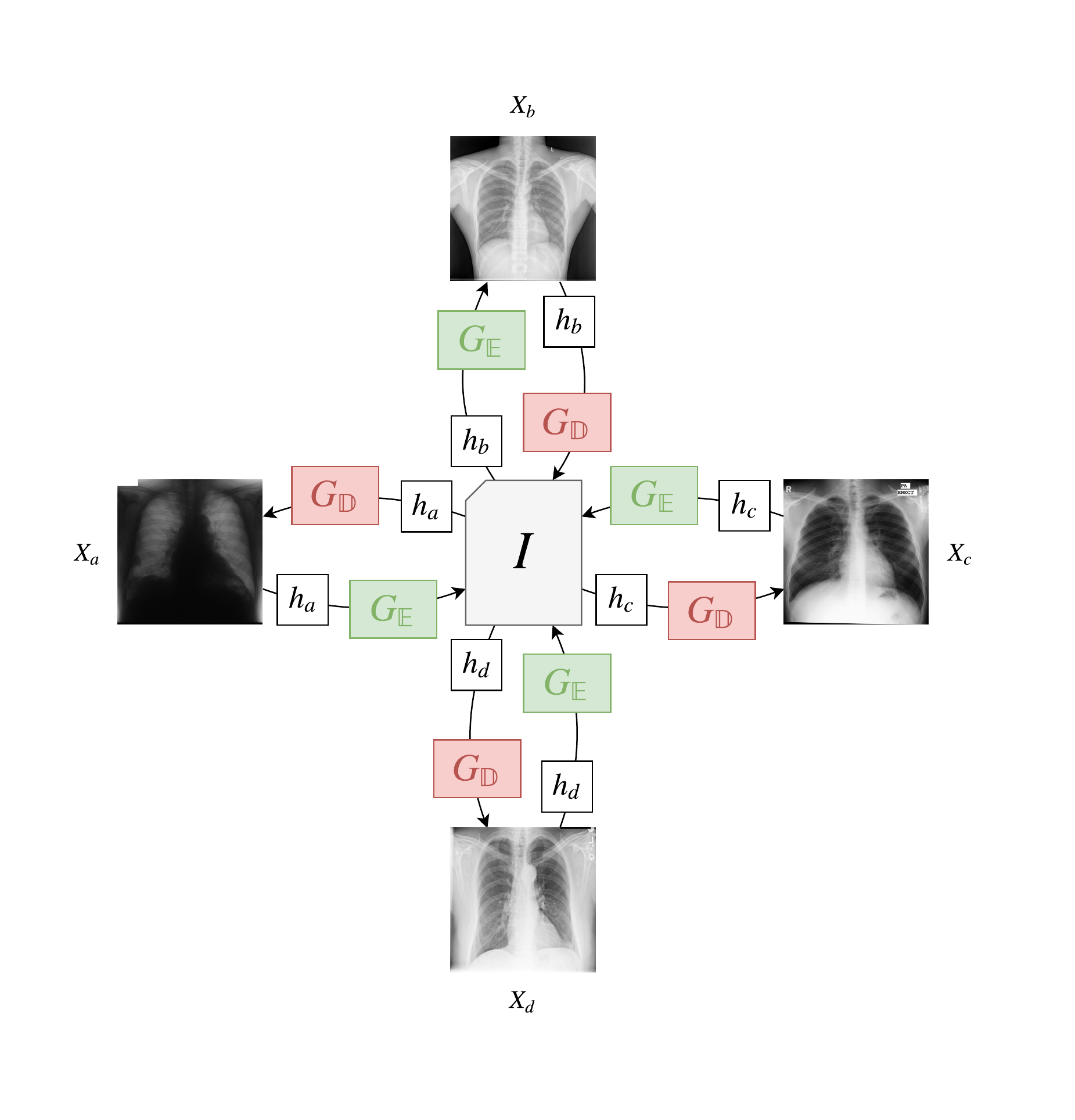}%
    }
    \caption{Comparison between D2D architectures and CoDAGANs regarding architectural choices for computing the isomorphic representation. While D2D use an Encoder/Decoder pair for each domain, CoDAGANs use One Hot Encoding in order to allow training with more than two domains without scalability hurdles.}
    \label{fig:conditional_dataset}
\end{figure}

While D2D approaches use a coupled architecture composed of 2 encoders ($G_{\mathbb{E}_{a}}$ and $G_{\mathbb{E}_{b}}$) and 2 decoders ($G_{\mathbb{D}_{a}}$ and $G_{\mathbb{D}_{b}}$) for learning a joint distribution over datasets $a$ and $b$, CoDAGANs use only one generator $G$ composed of one encoder and one decoder ($G_{\mathbb{E}}$ and $G_{\mathbb{D}}$). Additionally to the data $X_{k}$ from some dataset $k$, $G_{\mathbb{E}}$ is conditionally fed a One Hot Encoding $h_{k}$, as in $I = G_{\mathbb{E}}(X_{k}, h_{k})$. The addition of the data in $X_{k}$ to the code $h_{k}$ is achieved by simple concatenation, as shown in Figure~\ref{fig:one_hot}. The code $h_{k}$ forces the generator to encode the data according to the marginal distribution optimized for dataset $k$, conditioning the method to the visual style of these data, as exemplified in Figures~\ref{fig:codagan_training} and~\ref{fig:conditional_dataset}. The code $h_{l}$ for a second dataset $l$ is received by the decoder, as in $\hat{X}_{k \rightarrow l} = G_{\mathbb{D}}(I, h_{l})$, in order to produce the translation $\hat{X}_{k \rightarrow l}$ to dataset $l$.

\subsection{Training Routines in CoDAGANs}
\label{sec:training_routines}

In each iteration of a traditional GAN there are two routines for training the networks: 1) freezing the discriminator and updating the generator (\textit{Gen Update}); and 2) freezing the generator and updating the discriminator (\textit{Dis Update}). Performing these routines intermittently allows the networks to converge together in unsupervised settings. CoDAGANs add a new supervised routine to this scheme in order to perform UDA, SSDA and FSDA: \textit{Model Update}. The subroutines described in Section~\ref{sec:coda} that compose the three routines of CoDAGANs are presented in Table~\ref{tab:routines}

\begin{table}[!t]
    \centering
    \caption{Subroutines for each routine of CoDAGANs.}
    \label{tab:routines}
    \begin{tabular}{|c|c|c|c|}
        \hline
        \backslashbox{\textbf{Subroutine}}{\textbf{Routine}} & \textbf{$\bm{G}$ Update} & \textbf{$\bm{D}$ Update} & \textbf{$\bm{M}$ Update} \\ \hhline{=|=|=|=}
        \textbf{Encode}       & \checkmark & \checkmark & \checkmark \\ \hline
        \textbf{Decode}       & \checkmark & \checkmark & \checkmark \\ \hline
        \textbf{Reencode}     & \checkmark & X          & \checkmark \\ \hline
        \textbf{Redecode}     & \checkmark & X          & \checkmark \\ \hline
        \textbf{Discriminate} & X          & \checkmark & X          \\ \hline
        \textbf{Supervision}  & X          & X          & \checkmark \\ \hline
    \end{tabular}
\end{table}

Since the first proposal of GANs \cite{Goodfellow:2014}, stability has been considered a major problem in GAN training. Adversarial training is known to be more susceptible to convergence problems \cite{Goodfellow:2014,Salimans:2016} than traditional training procedures for DNNs due to problems as: more complex objectives composed of two or more (often contradictory) terms, discrepancies between the capacities of $G$ and $D$, mode collapse etc. Therefore, in order to achieve more stable results, we split the training procedure of CoDAGANs into two phases: a) \textit{Full Training} and b) \textit{Supervision Tuning}; which will be explained on the following paragraphs.

\paragraph{\textit{Full Training}} During the first 75\% of the epochs in a CoDAGAN training procedure, \textit{Full Training} is performed. This training phase is composed of the procedures \textit{Dis Update}, \textit{Gen Update} and \textit{Model Update}, executed in this order. That is, for each iteration in an epoch of the \textit{Full Training} phase, first the discriminator $D$ is optimized, followed by an update of $G$ and finishing with the update of the supervised model. During this phase adversarial training enforces the creation of good isomorphic representations by $G$ and translations between the domains. At the same time, the model $M$ uses the existing (and potentially scarce) label information in order to improve the translations performed by $G$ by adding semantic meaning to the translated visual features in the samples.

\paragraph{\textit{Supervision Tuning}} The last 25\% of the network epochs are trained in the \textit{Supervision Tuning} setting. This phase removes the unstable adversarial training by freezing $G$ and performing only the \text{Model Update} procedure, effectively tuning the supervised model to a stationary isomorphic representation. Freezing $G$ has the effect of removing the instability generated by the adversarial training in the translation process, as it is harder for $M$ to converge properly while the isomorphic input $I$ is constantly changing its visual properties due to changes in the weights of $G$.




\subsection{CoDAGAN Loss}
\label{sec:coda_loss}

Both UNIT \cite{Liu:2017} and MUNIT \cite{Huang:2018} optimize conjointly GAN-like adversarial loss components and Cycle-Consistency reconstruction losses. Cycle-Consistency losses ($\mathcal{L}_{cyc}$) are used in order to provide unsupervised training capabilities to these translation methods, allowing for the use of unpaired image datasets, as paired samples from distinct domains are often hard or impossible to create. Cycle Consistency is often achieved via Variational inference, which tries to find an upper bound to the Maximum Likelihood Estimation (MLE) of high dimensional data \cite{Kingma:2013}. Variational losses allow VAEs to generate new samples learnt from an approximation to the original data distribution as well as reconstruct images from these distributions. Optimizing an upper bound to the MLE allows VAEs to produce samples with high likelihood regarding the original data distribution, but still possessing low visual quality.

Adversarial losses ($\mathcal{L}_{adv}$) are often complementarily used with reconstruction losses in order to yield high visual quality and detailed images, as GANs are widely observed to take bigger risks in generating samples than simple regression losses \cite{Isola:2017}. Simpler approaches to image generation tend to average the possible outcomes of new samples, producing low quality images, therefore GANs produce less blurry and more realistic images than non-adversarial approaches in most settings. Unsupervised Image-to-Image Translation architectures normally use a weighted sum of these previously discussed losses as their total loss function ($\mathcal{L}_{tot}$), as in:
\begin{equation}
    \label{eq:total_loss_unsupervised}
    \begin{split}
        \mathcal{L}_{tot} = & \lambda_{cyc} \left[ \mathcal{L}_{cyc}(X_{a}, X_{a \rightarrow b \rightarrow a}) + \mathcal{L}_{cyc}(X_{b}, X_{b \rightarrow a \rightarrow b}) \right] + \\
                          & \lambda_{adv} \left[ \mathcal{L}_{adv}(X_{b}, X_{a \rightarrow b}) + \mathcal{L}_{adv}(X_{a}, X_{b \rightarrow a}) \right]\text{.}
    \end{split}
\end{equation}

More details on UNIT and MUNIT loss components can be found in their respective original papers \cite{Liu:2017,Huang:2018}. One should notice that we only presented the architecture-agnostic routines and loss components for CoDAGANs in the previous subsections, as the choice of Unsupervised Image-to-Image Translation basis network might introduce new objective terms and/or architectural changes. MUNIT, for instance, computes reconstruction losses to both the pair of images $\{ X_{a}, X_{a \rightarrow b \rightarrow a} \}$ and the pair of isomorphic representations $\{ I_{a}, I_{a \rightarrow b} \}$, which are separated into style and content components in this architecture.

CoDAGANs add a new supervised component $\mathcal{L}_{sup}$ to the completely unsupervised loss $\mathcal{L}_{tot}$ of Unsupervised Image-to-Image Translation methods. The supervised component for CoDAGANs is the default cost function for supervised classification/segmentation tasks, the Cross-Entropy loss (Equation~\ref{eq:cross_entropy}). 
The full objective $\mathcal{L}_{CoDA}$ for CoDAGANs is, therefore, defined by:
\begin{align*}
    \centering
    \label{eq:total_loss_coda}
    \mathcal{L}_{CoDA} = \lambda_{cyc} [&\mathcal{L}_{cyc}(X_{a}, X_{a \rightarrow b \rightarrow a})\ +\ \mathcal{L}_{cyc}(X_{b}, X_{b \rightarrow a \rightarrow b})]\ + \\
                         \lambda_{adv} [&\mathcal{L}_{adv}(X_{b}, X_{a \rightarrow b})              \ +\ \mathcal{L}_{adv}(X_{a}, X_{b \rightarrow a})]\ + \\
                         \lambda_{sup} [&\mathcal{L}_{sup}(Y_{a}, M(I_{a}))                         \ +\ \mathcal{L}_{sup}(Y_{b}, M(I_{b}))\ + \\
                                      &\mathcal{L}_{sup}(Y_{a}, M(I_{a \rightarrow b}))           \ +\ \mathcal{L}_{sup}(Y_{b}, M(I_{b \rightarrow a}))] \text{.}
                         \numberthis
\end{align*}
\section{Experimental Setup}
\label{sec:setup}

All code was implemented using the PyTorch\footnotemark \footnotetext{\url{https://pytorch.org/}} Deep Learning framework. We used the MUNIT/UNIT implementation from Huang \textit{et al.} \cite{Huang:2018}\footnotemark \footnotetext{\url{https://github.com/nvlabs/MUNIT}} as a basis and some segmentation architectures from the pytorch-semantic-segmentation\footnotemark \footnotetext{\url{https://github.com/zijundeng/pytorch-semantic-segmentation/}} project. All tests were conducted on NVIDIA Titan X Pascal GPUs with 12GB of memory. Our implementation can be found in this project's website\footnotemark \footnotetext{\url{http://www.patreo.dcc.ufmg.br/codagans/}}.


\subsection{Hyperparameters}
\label{sec:hyperparameters}

Architectural choices and hyperparameters can be further analysed according to the codes and configuration files in the project's website, but the main ones are described in the following paragraphs. CoDAGANs were run for 400 epochs in our experiments, as this was empirically found to be a good stopping point for convergence in these networks. Learning rate was set to $1 \times 10^{-4}$ with L2 normalization by weight decay with value $1 \times 10^{-5}$ and we used the Adam solver \cite{Kingma:2014}. $G_{\mathbb{E}}$ is composed of two downsampling layers followed by two residual layers for both UNIT \cite{Liu:2017} and MUNIT \cite{Huang:2018} based implementations, as these configurations were observed to simultaneously yield satisfactory results and have small GPU memory requirements. The first downsampling layer contains 32 convolutional filters, doubling this number for each subsequent layer. $D$ was implemented using a Least Squares Generative Adversarial Network (LSGAN) \cite{Mao:2017} objective with only two layers, although differently from MUNIT, we do not employ multiscale discriminators due to GPU memory constraints. Also distinctly from MUNIT and UNIT, we do not employ the VGG-based \cite{Simonyan:2014} perceptual loss -- further detailed by Huang \textit{et al.} \cite{Huang:2018} -- due to the dissimilarities between the domains wherein these networks were pretrained and the biomedical images used in our work.


\subsection{Datasets}
\label{sec:datasets}

We tested our methodology in a total of 16 datasets, 8 of them being Chest X-Ray (CXR) datasets, 6 of them being Mammographic X-Ray (MXR) datasets and 2 of them being composed of Dental X-Ray (DXR) images. The chosen CXR datasets are the Japanese Society of Radiological Technology (JSRT) \cite{Shiraishi:2000}\footnotemark \footnotetext{\url{http://db.jsrt.or.jp/eng.php}}, OpenIST\footnotemark \footnotetext{\url{https://github.com/pi-null-mezon/OpenIST}}, Shenzhen and Montgomery sets \cite{Jaeger:2014}\footnotemark \footnotetext{\url{https://lhncbc.nlm.nih.gov/publication/pub9931}}, Chest X-Ray 8 \cite{Wang:2017}\footnotemark \footnotetext{\url{https://nihcc.app.box.com/v/ChestXray-NIHCC/folder/37178474737}}, PadChest \cite{Bustos:2019} \footnotemark \footnotetext{\url{http://bimcv.cipf.es/bimcv-projects/padchest/}}, NLMCXR \cite{Demner:2015} \footnotemark \footnotetext{\url{https://openi.nlm.nih.gov/}} and the Optical Coherence Tomography and Chest X-Ray Images (OCT CXR) \cite{Kermany:2018} \footnotemark \footnotetext{\url{https://data.mendeley.com/datasets/rscbjbr9sj/3}} dataset. The MXR datasets used in this work are INbreast \cite{Moreira:2012}\footnotemark \footnotetext{\url{http://medicalresearch.inescporto.pt/breastresearch/index.php/Get_INbreast_Database}}, the Mammographic Image Analysis Society (MIAS) dataset \cite{Suckling:2015}\footnotemark \footnotetext{\url{https://www.repository.cam.ac.uk/handle/1810/250394}}, the Digital Database for Screening Mammography (DDSM) \cite{Heath:2000}\footnotemark \footnotetext{\url{http://marathon.csee.usf.edu/Mammography/Database.html}}, the Breast Cancer Digital Repository (BCDR) \cite{Lopez:2012}\footnotemark \footnotetext{\url{https://bcdr.eu/patient/list#}}, and LAPIMO \cite{Matheus:2011} \footnotemark \footnotetext{\url{http://lapimo.sel.eesc.usp.br/bancoweb/english/}}. DDSM was split into two groups: 1) samples A, and 2) samples B/C; as these groups were acquired and digitized with different equipments, yielding considerably distinct visual patterns. At last, the two DXR datasets we used in our experiments are the IvisionLab \cite{Silva:2018}\footnotemark \footnotetext{\url{https://github.com/IvisionLab/deep-dental-image}} and the Panoramic X-Ray \cite{Abdi:2015}\footnotemark \footnotetext{\url{https://data.mendeley.com/datasets/hxt48yk462/1}} datasets.

A total of 7 distinct segmentation tasks are compared in our experiments: 1) Pectoral muscle, 2) Breast region in MXRs; 3) Lungs, 4) Heart, 5) Clavicles in CXRs; 6) Mandible and 7) Teeth in DXRs.

\subsection{Experimental Protocol}
\label{sec:protocol}

All datasets were randomly split into training and test sets according to an 80\%/20\% division. Aiming to mimic real-world scenarios wherein the lack of labels is a considerable problem, we did not keep samples for validation purposes. We evaluate results from epochs 360, 370, 380, 390 and 400 for computing the mean and standard deviation values presented in Section~\ref{sec:results} in order to consider the statistical variability of the methods during the last epochs of the training procedure. 

For quantitative assessment we used the Jaccard (Intersection over Union -- IoU) metric, which is a common choice in segmentation and detection tasks and is widely used in all tested domains \cite{Rampun:2017,Van:2006,Silva:2018}. 
Jaccard ($J$) for a binary classification task is given by the following equation:
\begin{equation}
    J = \frac{TP}{TP + FN + FP}\textit{,} 
    \label{eq:jaccard}
\end{equation}
where $TP$, $FN$ and $FP$ refer to True Positives, False Negatives and False Positives, respectively. Jaccard values range between 0 and 1, however we present these metrics as percentages by multiplying them by a factor of 100 in Section~\ref{sec:results}.




\subsection{Baselines}
\label{sec:baselines}

Large datasets as ImageNet \cite{Deng:2009} turned Fine-tuning DNNs into a well known method for Transfer Learning in the Deep Learning literature, as most specific datasets do not possess the large amount of labeled data required for training from scratch in classification tasks. Fine-tuning was later adapted for dense labeling tasks \cite{Long:2015} and is nowadays common procedure in semantic segmentation tasks in the Computer Vision domain. However, Fine-tuning still does not work in UDA, as it necessarily requires labeled data. Therefore, we inserted the use of Pretrained DNNs as baselines both without further training in UDA scenarios and as basis for Fine-tuning in SSDA and FSDA scenarios. Still in the field of classical approaches do Transfer Learning, we add as a baseline to our experimental procedure training a DNN From Scratch with the smaller amount of labeled data available for targets datasets in SSDA and FSDA scenarios.

Our main baseline was the D2D approach proposed by Oliveira \textit{et al.} \cite{Oliveira:2018}, as it uses a Cycle Consistent GAN with a similar architecture as CoDAGANs. However, instead of performing the supervised learning at the one of the ends of the translation procedure -- as most of the literature does \cite{Hoffman:2018,Murez:2018,Wu:2018,Oliveira:2018} -- CoDAGANs optimize the supervised loss using the isomorphic representation as input.





\section{Results and Discussion}
\label{sec:results}

Quantitative results presented in this section are divided by domain and type of analysis. Sections~\ref{sec:results_mxr} and~\ref{sec:results_cxr} presents the UDA, SSDA and FSDA result regarding segmentation of anatomical structures in MXRs and CXRs, respectively. DXRs are evaluated qualitatively, as there is only one labeled dataset per task. In order to mimic the lack of labels in the tasks while still being able to evaluate the performance of our method in UDA and SSDA scenarios, we tested six labels configurations in CXRs and MXRs. Experiments with only source labels (UDA) are referred to as $E_{0\%}$ and experiments with the whole range of labels available for training (FSDA) are denominated $E_{100\%}$. Between $E_{0\%}$ and $E_{100\%}$, we limited the amount of target labels to 2.5\% ($E_{2.5\%}$), 5\% ($E_{5\%}$), 10\% ($E_{10\%}$) and 50\% ($E_{50\%}$), emulating four SSDA scenarios. Each table in Sections~\ref{sec:results_mxr} and~\ref{sec:results_cxr} attribute one uppercase letter for each dataset, so that they can be more easily be referenced during the discussion.

Qualitative analysis in both unlabeled and labeled data are presented in Section~\ref{sec:results_qualitative} for all domains. Section~\ref{sec:results_qualitative_i} discusses the activations of channels in isomorphic representations, where supervised learning is performed and Domain Generalization is enforced. At last, Section~\ref{sec:results_dimensionality} discusses the distributions of samples across different domains and datasets computed from the isomorphic space $I$.


\subsection{Quantitative Results for MXR Samples}
\label{sec:results_mxr}

Jaccard average values and standard deviations for MXR tasks are shown in Tables~\ref{tab:results_pectoral} and~\ref{tab:results_breast} for pectoral muscle and breast region segmentation, respectively. The first lines in the tables present the label configurations used in the experiments at each column. Results are shown separately for datasets INbreast (A), MIAS (B), DDSM B/C (C), and DDSM A (D) in Table~\ref{tab:results_pectoral} and for datasets INbreast (A) and MIAS (B) in Table~\ref{tab:results_breast}. Objective results for datasets BCDR (E) and LAPIMO (F) for pectoral muscle and datasets (C)-(F) in breast region segmentation are not possible due to the complete lack of labels in these tasks. We reinforce that only two CoDAGANs ($\bm{CoDA_{M}}$ using MUNIT \cite{Huang:2018} and $\bm{CoDA_{U}}$ based on UNIT \cite{Liu:2017}) were trained for all datasets in each task, as CoDAGANs allow for multi-source and multi-target DA. Thus, repeated columns indicating the results for $\bm{CoDA_{M}}$ and $\bm{CoDA_{U}}$ are simply reporting the results of the same models for different datasets. All methods beside $\bm{CoDA_{M}}$ and $\bm{CoDA_{U}}$ indicate whether the source or target data used in the training, as they are neither multi-source nor multi-target, limiting them to pairwise training.

\begin{table*}[!t]
    \centering
    \caption{Jaccard results (in \%) for pectoral muscle segmentation DA to and/or from six distinct MXR datasets: INbreast (A), MIAS (B), DDSM B/C (C), DDSM A (D), BCDR (E) and LAPIMO (F).}
    \label{tab:results_pectoral}
    \begin{tabular}{|c|c|c|c|c|c|c|c|}
        \hline
        \multicolumn{2}{|c|}{\textbf{Experiments}}                                       & \textbf{$\bm{E_{0\%}}$}   & \textbf{$\bm{E_{2.5\%}}$} & \textbf{$\bm{E_{5\%}}$}   & \textbf{$\bm{E_{10\%}}$}  & \textbf{$\bm{E_{50\%}}$}  & \textbf{$\bm{E_{100\%}}$} \\ \hhline{|==|=|=|=|=|=|=|}
        \multicolumn{2}{|c|}{\textbf{\% Labels INbreast (A)}}                            & 100.00\%                  & 100.00\%                  & 100.00\%                  & 100.00\%                  & 100.00\%                  & 100.00\%                  \\ \hline
        \multicolumn{2}{|c|}{\textbf{\% Labels MIAS (B)}}                                & 0.00\%                    & 2.50\%                    & 5.00\%                    & 10.00\%                   & 50.00\%                   & 100.00\%                  \\ \hline
        \multicolumn{2}{|c|}{\textbf{\% Labels DDSM\_BC (C)}}                            & 0.00\%                    & 0.00\%                    & 0.00\%                    & 0.00\%                    & 0.00\%                    & 0.00\%                    \\ \hline
        \multicolumn{2}{|c|}{\textbf{\% Labels DDSM\_A (D)}}                             & 0.00\%                    & 0.00\%                    & 0.00\%                    & 0.00\%                    & 0.00\%                    & 0.00\%                    \\ \hline
        \multicolumn{2}{|c|}{\textbf{\% Labels BCDR (E)}}                                & 0.00\%                    & 0.00\%                    & 0.00\%                    & 0.00\%                    & 0.00\%                    & 0.00\%                    \\ \hline
        \multicolumn{2}{|c|}{\textbf{\% Labels LAPIMO (F)}}                              & 0.00\%                    & 0.00\%                    & 0.00\%                    & 0.00\%                    & 0.00\%                    & 0.00\%                    \\ \hhline{|==|=|=|=|=|=|=|}
        \multirow{13}{*}{\textbf{(A)}} & $\bm{CoDA_{M}}$                                 & 91.95 $\pm$ 0.81          & 92.57 $\pm$ 0.31          & \textbf{92.61 $\pm$ 0.44} & \textbf{92.00 $\pm$ 0.90} & \textbf{90.66 $\pm$ 0.53} & 88.58 $\pm$ 1.76          \\ \cline{2-8} 
                                       & $\bm{CoDA_{U}}$                                 & 91.18 $\pm$ 0.36          & 90.61 $\pm$ 0.89          & 91.03 $\pm$ 1.43          & 91.23 $\pm$ 1.51          & 90.36 $\pm$ 0.98          & 89.98 $\pm$ 0.37          \\ \cline{2-8} 
                                       & \textbf{$\bm{D2D_{M}}$ (A)$\rightarrow$(B)} & 93.27 $\pm$ 0.51          & \textbf{92.67 $\pm$ 0.64} & 87.54 $\pm$ 10.00         & 90.58 $\pm$ 2.14          & 84.46 $\pm$ 3.36          & \textbf{90.11 $\pm$ 1.06} \\ \cline{2-8} 
                                       & \textbf{$\bm{D2D_{U}}$ (A)$\rightarrow$(B)} & 92.27 $\pm$ 0.55          & 83.17 $\pm$ 9.27          & 89.56 $\pm$ 2.21          & 23.82 $\pm$ 10.21         & 81.64 $\pm$ 6.22          & 86.81 $\pm$ 2.98          \\ \cline{2-8} 
                                       & \textbf{$\bm{D2D_{M}}$ (A)$\rightarrow$(C)} & 93.43 $\pm$ 0.24          & --                        & --                        & --                        & --                        & --                        \\ \cline{2-8} 
                                       & \textbf{$\bm{D2D_{U}}$ (A)$\rightarrow$(C)} & 93.64 $\pm$ 0.50          & --                        & --                        & --                        & --                        & --                        \\ \cline{2-8} 
                                       & \textbf{$\bm{D2D_{M}}$ (A)$\rightarrow$(D)} & 93.72 $\pm$ 0.96          & --                        & --                        & --                        & --                        & --                        \\ \cline{2-8} 
                                       & \textbf{$\bm{D2D_{U}}$ (A)$\rightarrow$(D)} & 88.06 $\pm$ 2.30          & --                        & --                        & --                        & --                        & --                        \\ \cline{2-8} 
                                       & \textbf{$\bm{D2D_{M}}$ (A)$\rightarrow$(E)} & \textbf{93.87 $\pm$ 0.71} & --                        & --                        & --                        & --                        & --                        \\ \cline{2-8} 
                                       & \textbf{$\bm{D2D_{U}}$ (A)$\rightarrow$(E)} & 92.55 $\pm$ 0.57          & --                        & --                        & --                        & --                        & --                        \\ \cline{2-8} 
                                       & \textbf{$\bm{D2D_{M}}$ (A)$\rightarrow$(F)} & 92.29 $\pm$ 2.06          & --                        & --                        & --                        & --                        & --                        \\ \cline{2-8} 
                                       & \textbf{$\bm{D2D_{U}}$ (A)$\rightarrow$(F)} & 91.47 $\pm$ 0.55          & --                        & --                        & --                        & --                        & --                        \\ \cline{2-8} 
                                       & \textbf{From Scratch in (A)}                    & 93.25 $\pm$ 0.75          & --                        & --                        & --                        & --                        & \textbf{--}               \\ \hhline{|==|=|=|=|=|=|=|}
        \multirow{6}{*}{\textbf{(B)}}  & $\bm{CoDA_{M}}$                                 & \textbf{67.61 $\pm$ 2.07} & 69.92 $\pm$ 2.42          & \textbf{72.31 $\pm$ 0.65} & 75.66 $\pm$ 0.98          & 78.24 $\pm$ 0.23          & \textbf{79.08 $\pm$ 0.78} \\ \cline{2-8} 
                                       & $\bm{CoDA_{U}}$                                 & 60.01 $\pm$ 2.77          & 61.81 $\pm$ 3.26          & 71.33 $\pm$ 1.81          & \textbf{76.67 $\pm$ 0.15} & \textbf{78.37 $\pm$ 0.54} & 78.49 $\pm$ 1.38          \\ \cline{2-8} 
                                       & \textbf{$\bm{D2D_{M}}$ (A)$\rightarrow$(B)} & 0.00 $\pm$ 0.00           & 0.00 $\pm$ 0.00           & 22.73 $\pm$ 17.62         & 17.72 $\pm$ 15.05         & 35.46 $\pm$ 15.24         & 64.46 $\pm$ 5.61          \\ \cline{2-8} 
                                       & \textbf{$\bm{D2D_{U}}$ (A)$\rightarrow$(B)} & 41.06 $\pm$ 19.00         & 36.72 $\pm$ 15.07         & 59.67 $\pm$ 3.59          & 59.63 $\pm$ 12.37         & 62.69 $\pm$ 9.92          & 75.95 $\pm$ 2.57          \\ \cline{2-8} 
                                       & \textbf{Pretrained (A)$\rightarrow$(B)}   & 40.49                     & \textbf{72.11 $\pm$ 0.16} & 60.46 $\pm$ 3.76          & 71.89 $\pm$ 1.22          & 75.52 $\pm$ 0.34          & 78.35 $\pm$ 1.20          \\ \cline{2-8} 
                                       & \textbf{From Scratch in (B)}                    & --                        & 58.51 $\pm$ 5.44          & 51.90 $\pm$ 1.38          & 63.32 $\pm$ 5.62          & 77.79 $\pm$ 0.44          & 78.08 $\pm$ 0.46          \\ \hhline{|==|=|=|=|=|=|=|}
        \multirow{5}{*}{\textbf{(C)}}  & $\bm{CoDA_{M}}$                                 & \textbf{89.99 $\pm$ 0.80} & \textbf{90.73 $\pm$ 0.83} & \textbf{91.49 $\pm$ 0.36} & \textbf{92.34 $\pm$ 0.57} & \textbf{92.80 $\pm$ 0.40} & \textbf{92.50 $\pm$ 0.48} \\ \cline{2-8} 
                                       & $\bm{CoDA_{U}}$                                 & 82.45 $\pm$ 4.01          & 86.21 $\pm$ 3.13          & 89.90 $\pm$ 2.10          & 90.71 $\pm$ 0.72          & 91.21 $\pm$ 0.63          & 92.24 $\pm$ 0.65          \\ \cline{2-8} 
                                       & \textbf{$\bm{D2D_{M}}$ (A)$\rightarrow$(C)} & 0.03 $\pm$ 0.01           & --                        & --                        & --                        & --                        & --                        \\ \cline{2-8} 
                                       & \textbf{$\bm{D2D_{U}}$ (A)$\rightarrow$(C)} & 0.64 $\pm$ 0.90           & --                        & --                        & --                        & --                        & --                        \\ \cline{2-8} 
                                       & \textbf{Pretrained (A)$\rightarrow$(C)}   & 78.22                     & --                        & --                        & --                        & --                        & --                        \\ \hhline{|==|=|=|=|=|=|=|}
        \multirow{5}{*}{\textbf{(D)}}  & $\bm{CoDA_{M}}$                                 & \textbf{49.38 $\pm$ 5.21} & \textbf{50.00 $\pm$ 4.37} & \textbf{49.20 $\pm$ 1.84} & \textbf{54.37 $\pm$ 2.21} & \textbf{77.59 $\pm$ 0.75} & \textbf{69.76 $\pm$ 3.89} \\ \cline{2-8} 
                                       & $\bm{CoDA_{U}}$                                 & 23.83 $\pm$ 2.15          & 26.13 $\pm$ 2.74          & 42.93 $\pm$ 4.73          & 35.10 $\pm$ 4.81          & 31.23 $\pm$ 3.76          & 56.46 $\pm$ 5.78          \\ \cline{2-8} 
                                       & \textbf{$\bm{D2D_{M}}$ (A)$\rightarrow$(D)} & 0.41 $\pm$ 0.27           & --                        & --                        & --                        & --                        & --                        \\ \cline{2-8} 
                                       & \textbf{$\bm{D2D_{U}}$ (A)$\rightarrow$(D)} & 0.74 $\pm$ 0.94           & --                        & --                        & --                        & --                        & --                        \\ \cline{2-8} 
                                       & \textbf{Pretrained (A)$\rightarrow$(D)}   & 22.20                     & --                        & --                        & --                        & --                        & --                        \\ \hline
    \end{tabular}
\end{table*}

\begin{table*}[!t]
    \centering
    \caption{Jaccard results (in \%) for breast region segmentation DA to and/or from six distinct MXR datasets: INbreast (A), MIAS (B), DDSM B/C (C), DDSM A (D), BCDR (E) and LAPIMO (F).}
    \label{tab:results_breast}
    \begin{tabular}{|c|c|c|c|c|c|c|c|}
        \hline
        \multicolumn{2}{|c|}{\textbf{Experiments}}                                       & \textbf{$\bm{E_{0\%}}$}   & \textbf{$\bm{E_{2.5\%}}$} & \textbf{$\bm{E_{5\%}}$}   & \textbf{$\bm{E_{10\%}}$}  & \textbf{$\bm{E_{50\%}}$}  & \textbf{$\bm{E_{100\%}}$} \\ \hhline{|==|=|=|=|=|=|=|}
        \multicolumn{2}{|c|}{\textbf{\% Labels INbreast (A)}}                            & 100.00\%                  & 100.00\%                  & 100.00\%                  & 100.00\%                  & 100.00\%                  & 100.00\%                  \\ \hline
        \multicolumn{2}{|c|}{\textbf{\% Labels MIAS (B)}}                                & 0.00\%                    & 2.50\%                    & 5.00\%                    & 10.00\%                   & 50.00\%                   & 100.00\%                  \\ \hline
        \multicolumn{2}{|c|}{\textbf{\% Labels DDSM\_BC (C)}}                            & 0.00\%                    & 0.00\%                    & 0.00\%                    & 0.00\%                    & 0.00\%                    & 0.00\%                    \\ \hline
        \multicolumn{2}{|c|}{\textbf{\% Labels DDSM\_A (D)}}                             & 0.00\%                    & 0.00\%                    & 0.00\%                    & 0.00\%                    & 0.00\%                    & 0.00\%                    \\ \hline
        \multicolumn{2}{|c|}{\textbf{\% Labels BCDR (E)}}                                & 0.00\%                    & 0.00\%                    & 0.00\%                    & 0.00\%                    & 0.00\%                    & 0.00\%                    \\ \hline
        \multicolumn{2}{|c|}{\textbf{\% Labels LAPIMO (F)}}                              & 0.00\%                    & 0.00\%                    & 0.00\%                    & 0.00\%                    & 0.00\%                    & 0.00\%                    \\ \hhline{|==|=|=|=|=|=|=|}
        \multirow{13}{*}{\textbf{(A)}} & $\bm{CoDA_{M}}$                                 & 98.69 $\pm$ 0.06          & 98.48 $\pm$ 0.15          & 98.59 $\pm$ 0.09          & 97.98 $\pm$ 0.36          & \textbf{98.27 $\pm$ 0.41} & \textbf{98.11 $\pm$ 0.20} \\ \cline{2-8} 
                                       & $\bm{CoDA_{U}}$                                 & 98.29 $\pm$ 0.13          & 98.12 $\pm$ 0.10          & 98.37 $\pm$ 0.15          & 97.79 $\pm$ 0.64          & 97.89 $\pm$ 0.27          & 98.04 $\pm$ 0.18          \\ \cline{2-8} 
                                       & \textbf{$\bm{D2D_{M}}$ (A)$\rightarrow$(B)} & 98.90 $\pm$ 0.09          & \textbf{98.93 $\pm$ 0.16} & 98.27 $\pm$ 0.45          & 98.36 $\pm$ 0.82          & 97.36 $\pm$ 1.60          & 85.89 $\pm$ 7.74          \\ \cline{2-8} 
                                       & \textbf{$\bm{D2D_{U}}$ (A)$\rightarrow$(B)} & 98.74 $\pm$ 0.14          & 98.26 $\pm$ 0.31          & \textbf{98.92 $\pm$ 0.19} & \textbf{98.65 $\pm$ 0.09} & 97.80 $\pm$ 0.52          & 95.01 $\pm$ 1.51          \\ \cline{2-8} 
                                       & \textbf{$\bm{D2D_{M}}$ (A)$\rightarrow$(C)} & 99.00 $\pm$ 0.08          & --                        & --                        & --                        & --                        & --                        \\ \cline{2-8} 
                                       & \textbf{$\bm{D2D_{U}}$ (A)$\rightarrow$(C)} & 98.90 $\pm$ 0.09          & --                        & --                        & --                        & --                        & --                        \\ \cline{2-8} 
                                       & \textbf{$\bm{D2D_{M}}$ (A)$\rightarrow$(D)} & 98.87 $\pm$ 0.15          & --                        & --                        & --                        & --                        & --                        \\ \cline{2-8} 
                                       & \textbf{$\bm{D2D_{U}}$ (A)$\rightarrow$(D)} & 98.83 $\pm$ 0.14          & --                        & --                        & --                        & --                        & --                        \\ \cline{2-8} 
                                       & \textbf{$\bm{D2D_{M}}$(A)$\rightarrow$(E)}  & \textbf{99.02 $\pm$ 0.05} & --                        & --                        & --                        & --                        & --                        \\ \cline{2-8} 
                                       & \textbf{$\bm{D2D_{U}}$ (A)$\rightarrow$(E)} & 98.90 $\pm$ 0.05          & --                        & --                        & --                        & --                        & --                        \\ \cline{2-8} 
                                       & \textbf{$\bm{D2D_{M}}$ (A)$\rightarrow$(F)} & 98.11 $\pm$ 1.65          & --                        & --                        & --                        & --                        & --                        \\ \cline{2-8} 
                                       & \textbf{$\bm{D2D_{U}}$ (A)$\rightarrow$(F)} & 98.69 $\pm$ 0.15          & --                        & --                        & --                        & --                        & --                        \\ \cline{2-8} 
                                       & \textbf{From Scratch in (A)}                    & 98.75 $\pm$ 0.11          & --                        & --                        & --                        & --                        & --                        \\ \hhline{|==|=|=|=|=|=|=|}
        \multirow{6}{*}{\textbf{(B)}}  & $\bm{CoDA_{M}}$                                 & 68.96 $\pm$ 1.57          & 88.13 $\pm$ 3.81          & 91.97 $\pm$ 0.68          & 93.11 $\pm$ 2.01          & 96.53 $\pm$ 0.45          & 97.19 $\pm$ 0.10          \\ \cline{2-8} 
                                       & $\bm{CoDA_{U}}$                                 & 69.72 $\pm$ 0.30          & 90.12 $\pm$ 0.60          & 91.97 $\pm$ 3.61          & 95.28 $\pm$ 0.25          & 95.86 $\pm$ 0.50          & 97.21 $\pm$ 0.15          \\ \cline{2-8} 
                                       & \textbf{$\bm{D2D_{M}}$ (A)$\rightarrow$(B)} & 5.02 $\pm$ 0.21           & 69.26 $\pm$ 14.74         & 5.91 $\pm$ 2.64           & 13.78 $\pm$ 8.17          & 50.20 $\pm$ 20.80         & 58.70 $\pm$ 29.26         \\ \cline{2-8} 
                                       & \textbf{$\bm{D2D_{U}}$ (A)$\rightarrow$(B)} & 9.63 $\pm$ 2.49           & 15.72 $\pm$ 2.39          & 66.02 $\pm$ 26.18         & 90.70 $\pm$ 1.82          & 95.93 $\pm$ 0.77          & 80.00 $\pm$ 13.50         \\ \cline{2-8} 
                                       & \textbf{Pretrained (A)$\rightarrow$(B)}   & \textbf{75.53}            & \textbf{91.94 $\pm$ 0.28} & 93.21 $\pm$ 0.35          & 94.06 $\pm$ 1.66          & 96.64 $\pm$ 0.13          & \textbf{97.40 $\pm$ 0.08} \\ \cline{2-8} 
                                       & \textbf{From Scratch in (B)}                    & --                        & 91.38 $\pm$ 0.28          & \textbf{93.27 $\pm$ 0.25} & \textbf{95.69 $\pm$ 0.22} & \textbf{97.10 $\pm$ 0.25} & 97.37 $\pm$ 0.15          \\ \hline
    \end{tabular}
\end{table*}

Bold values in these tables indicate the best results for the corresponding dataset indicated in the first column of these tables. As there are four datasets being evaluated in Table~\ref{tab:results_pectoral}, there are four bold values for each experiment. Analogously, Table~\ref{tab:results_breast} only has two bold values per column because only two datasets are being objectively evaluated in breast region segmentation. In both tables INbreast was used as source dataset, providing 100\% of its labels in all experiments. MIAS was used as both source ($E_{0\%}$) and target ($E_{2.5\%}$ to $E_{100\%}$) dataset, depending on the label configuration of the experiment. As DDSM does not possess pixel-level labels, we created some ground truths only for a small subset of images from this dataset for the pectoral muscle segmentation task in order to objectively evaluate the UDA. One should notice that these ground truths were used only on the test procedure, but not in training, as all cases presented in Tables~\ref{tab:results_pectoral} and~\ref{tab:results_breast} show DDSM with 0\% of labeled data. Thus DDSM is used only as a source dataset in our experiments. Breast region segmentation analysis on DDSM was only performed qualitatively, as there are no ground truths for this task.

\subsubsection{Pectoral Muscle Segmentation in MXR Images}
\label{sec:results_pectoral}

For the completely unlabeled case $E_{0\%}$ in pectoral muscle segmentation, $\bm{CoDA_{M}}$ and $\bm{CoDA_{U}}$ achieved $J$ values of 67.61\% and 60.01\% for the MIAS target dataset, while the best baseline achieved 41.06\%. SSDA and FSDA experiments ($E_{2.5\%}$ to $E_{100\%}$) regarding the MIAS dataset show that CoDAGANs achieve considerably better results than all baselines in all but one case. These results evidenced the higher instability of training pairwise translation architectures compared to conditional training. Across the training procedure, Jaccard values for D2D fluctuated by several percentage units, yielding standard deviations of one magnitude or more larger than CoDAGANs.

In the case of pectoral muscle for DDSM B/C (C), UDA using $\bm{CoDA_{M}}$ and $\bm{CoDA_{U}}$ achieved 89.99\% and 82.45\%, with the D2D baseline achieving worse than random results, evidencing its lack of capability to translate between domains (A) and (C). The best baseline in this case was simply the use of Pretrained DNNs in (A) and testing on (C), which achieved 78.22\%. Segmentation results for DDSM A (D) were considerably worse for all methods and experiments, as samples from this subset of images showed an extremely lower contrast compared to the samples of DDSM B/C. Even in this suboptimal case, CoDAGANs achieved much better results than the baseline in UDA. Preprocessing using adaptive histogram equalization in DDSM A (C) samples might improve results, although more empirical evidence is required. 
As there were only few samples labeled from (C) and (D), only UDA was possible for these datasets in D2D and pretrained baselines, as all labels were kept for testing. However, one can easily see that experiments $E_{2.5\%}$ to $E_{100\%}$ show better results in (C) and (D) as the number of labels from (B) increases, achieving a $J$ of 79.08\% with all (B) labels being used in training. This is due to two factors: 1) the larger number of labels achieved with the combination of (A) and (B); and 2) the more similar visual patterns between (B)$\rightarrow$(C) and (B)$\rightarrow$(D). Similarly to DDSM (C)--(D), MIAS (B) is an older originally analog that was later digitized, while INbreast (A) is a Full Field Digital Mammography (FFDM) dataset.

\subsubsection{Breast Region Segmentation in MXR Images}
\label{sec:results_breast}

Breast region segmentation (Table~\ref{tab:results_breast}) proved to be an easier task, with most methods achieving Jaccard values higher than 90\%. Pretrained DNNs and From Scratch training in SSDA scenarios achieved superior results in breast region segmentation for all experiments in the target MIAS (B) dataset, followed closely by CoDAGANs. D2D, however, grossly underperformed in this relatively easy task for all experiments, reiterating this strategy's instability during training.

The marginally lower performance of CoDAGANs in this task can be attributed to the high transferability of pretrained models, as can be seen in experiment $E_{0\%}$, where pretrained models with no Fine-tuning already achieved a $J$ value of 75.53\%. This easier DA task also benefits from the higher capability of U-Nets to segment details using skip connections between symmetric layers. As CoDAGANs remove the first layers of U-Net's Encoder to fit the smaller spatial dimensions of the isomorphic representation, the last layers of the network do not receive skip connections from the first layers, allowing for fine object details to be lost. This can be seen as a compromise between generalization capability and fine segmentation details.

\subsubsection{MXR Segmentation Confidence Intervals}
\label{sec:results_mxr_ci}

\begin{figure*}[!t]
    \centering
    \renewcommand{\currprop}{0.32\textwidth}
    \subfloat[]{
        \includegraphics[width=\currprop]{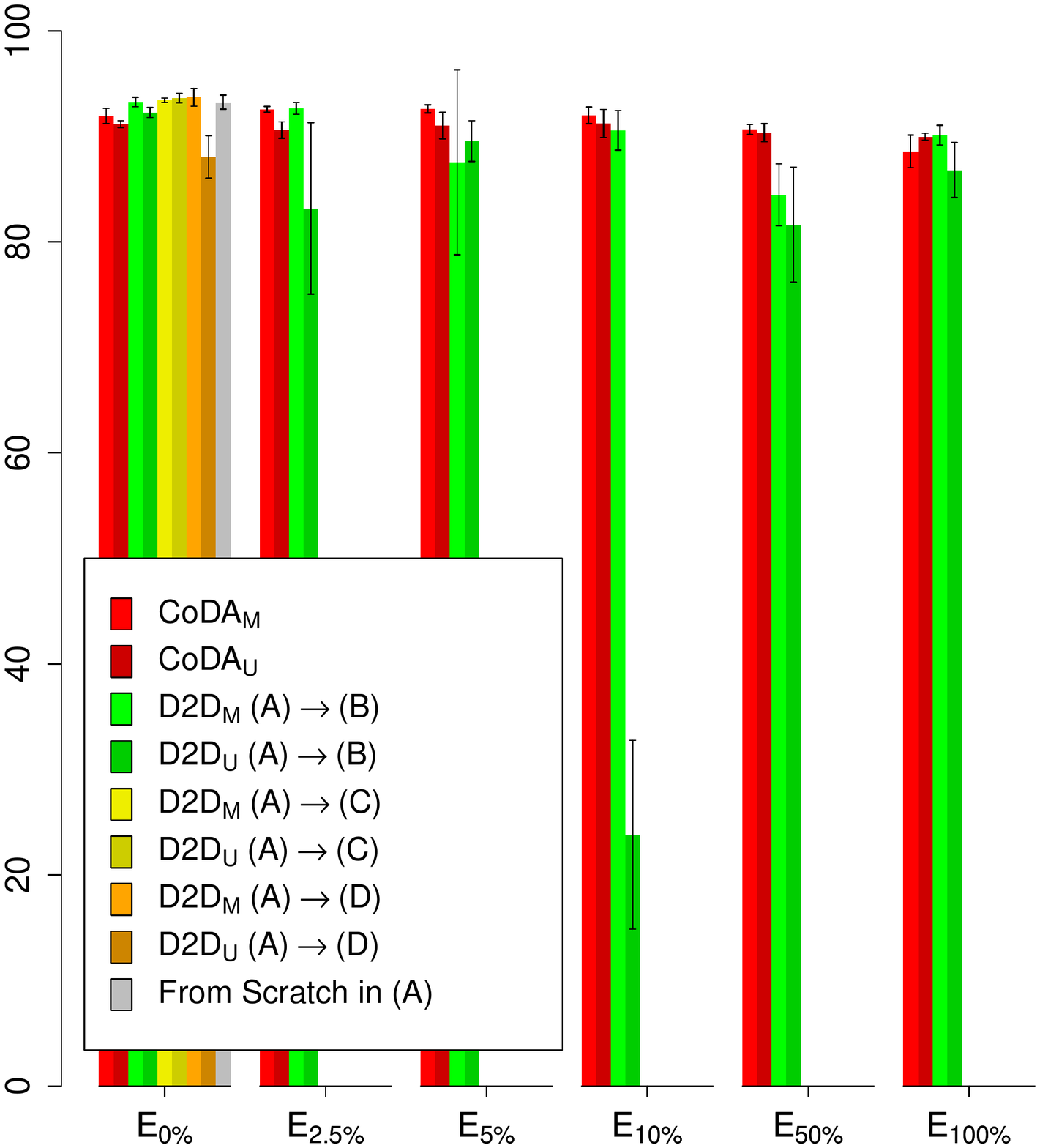}%
        \label{fig:intervals_mxr_a}
    }
    \hfil
    \subfloat[]{
        \includegraphics[width=\currprop]{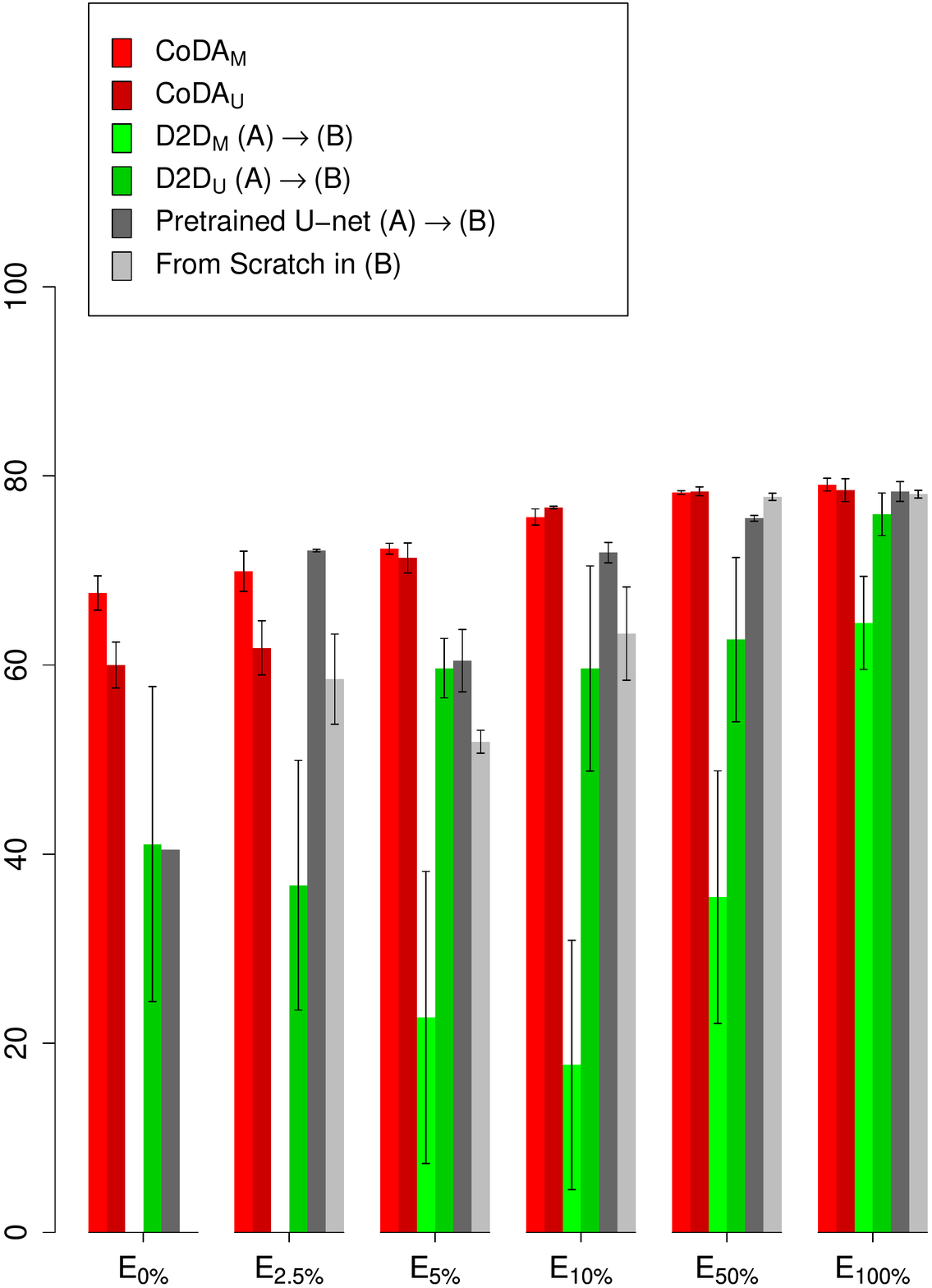}%
        \label{fig:intervals_mxr_b}
    }
    \hfil
    \subfloat[]{
        \includegraphics[width=\currprop]{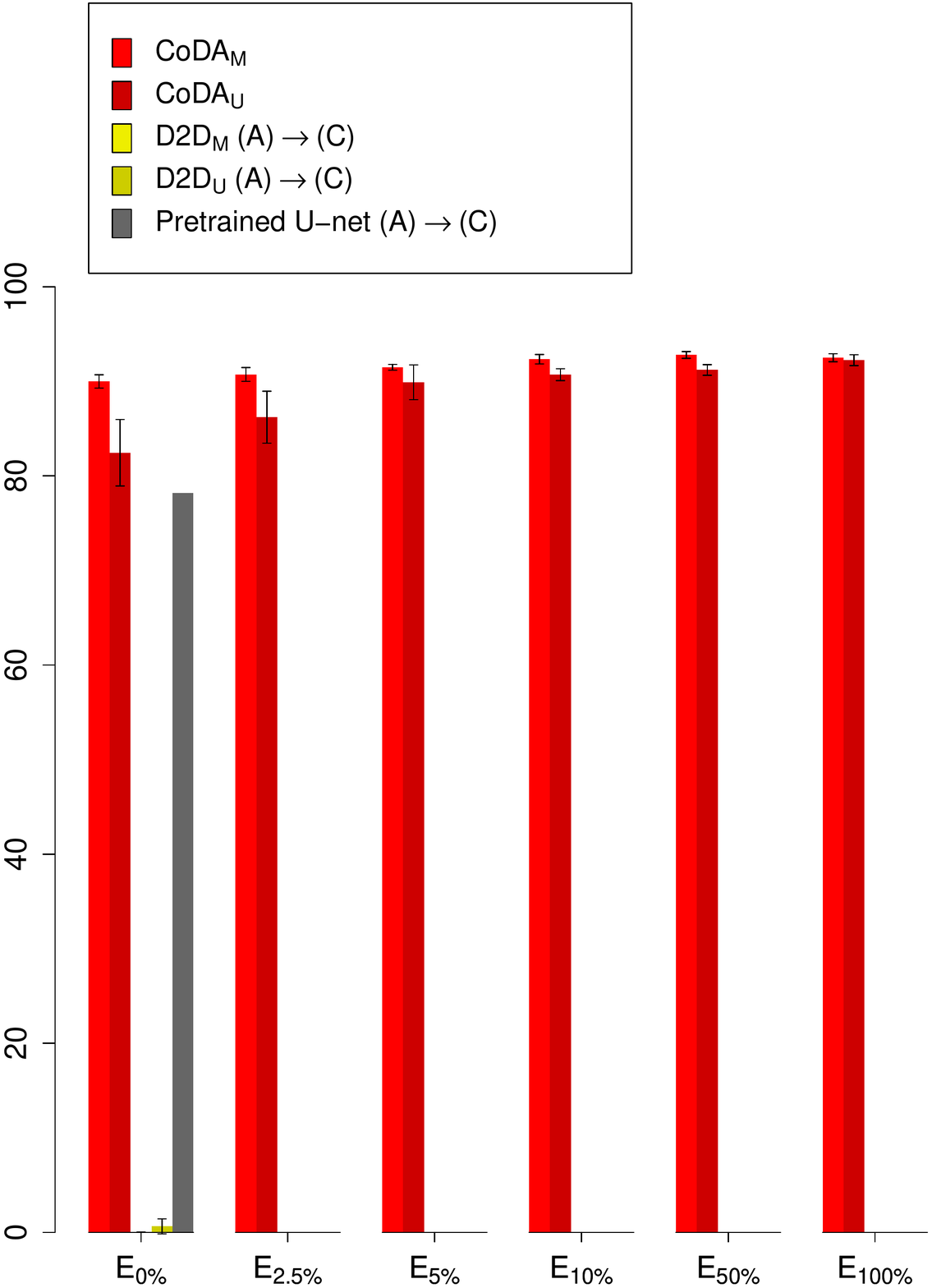}%
        \label{fig:intervals_mxr_c}
    }
    \\
    \subfloat[]{
        \includegraphics[width=\currprop]{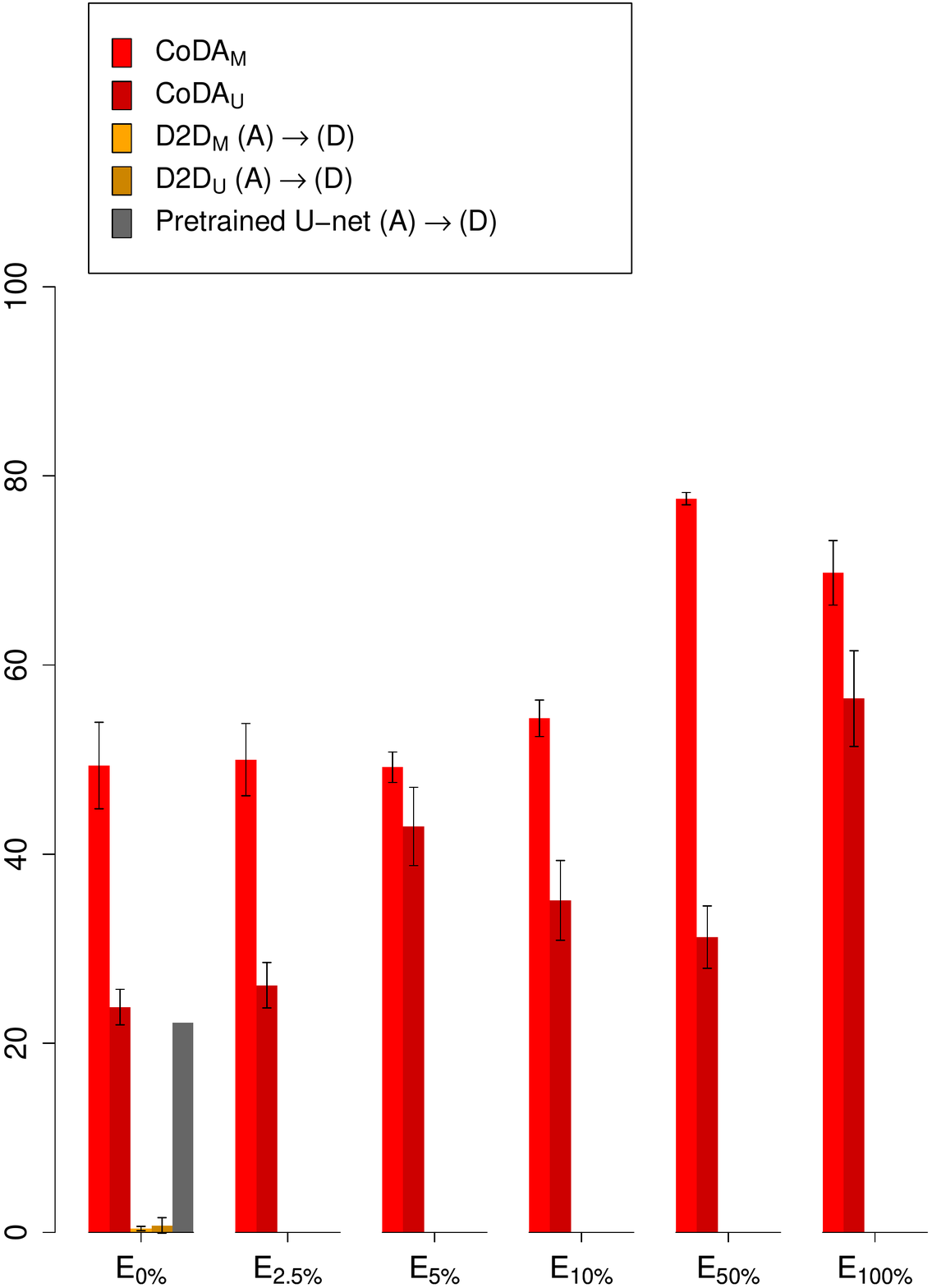}%
        \label{fig:intervals_mxr_d}
    }
    \hfil
    \subfloat[]{
        \includegraphics[width=\currprop]{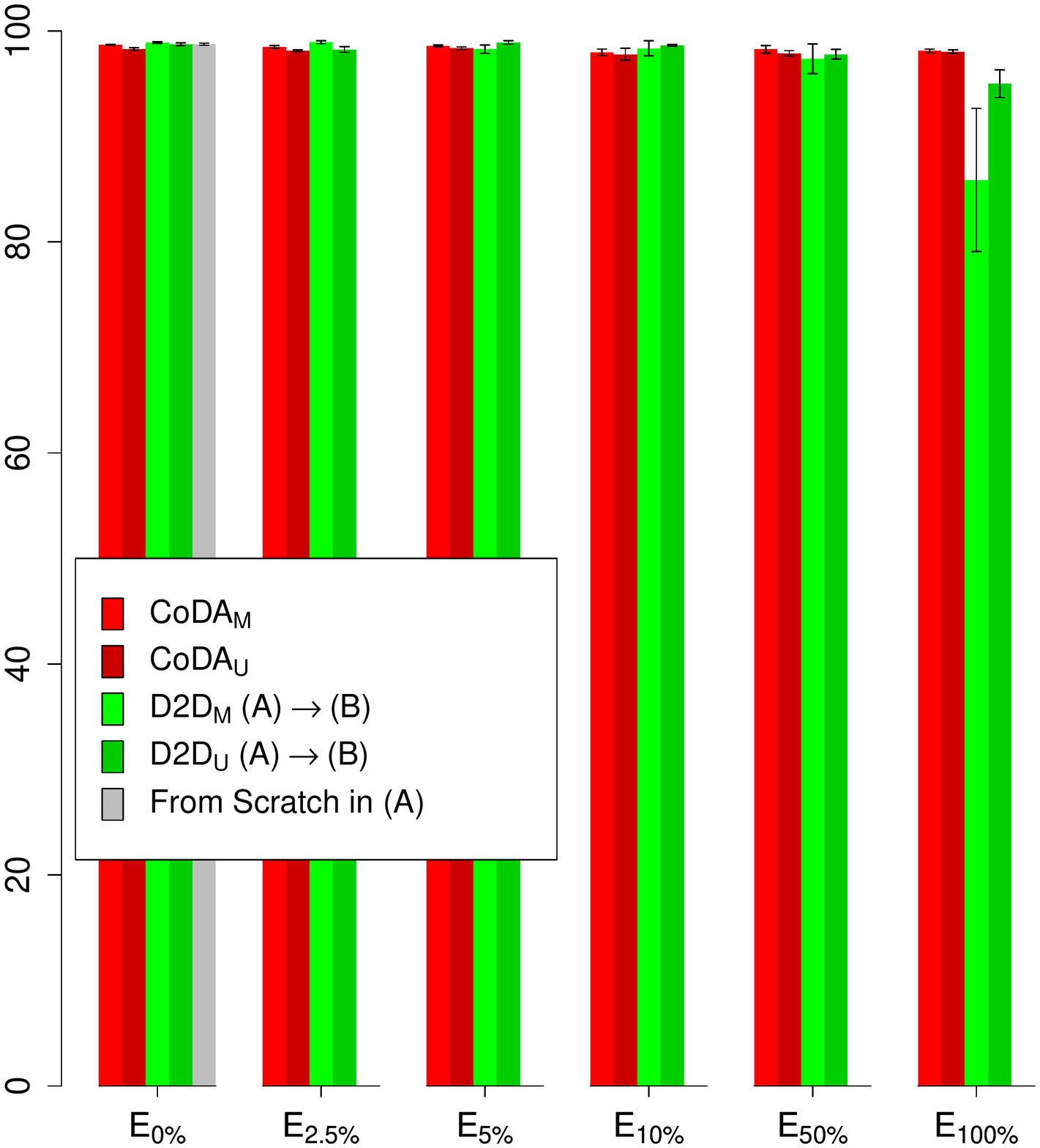}%
        \label{fig:intervals_mxr_e}
    }
    \hfil
    \subfloat[]{
        \includegraphics[width=\currprop]{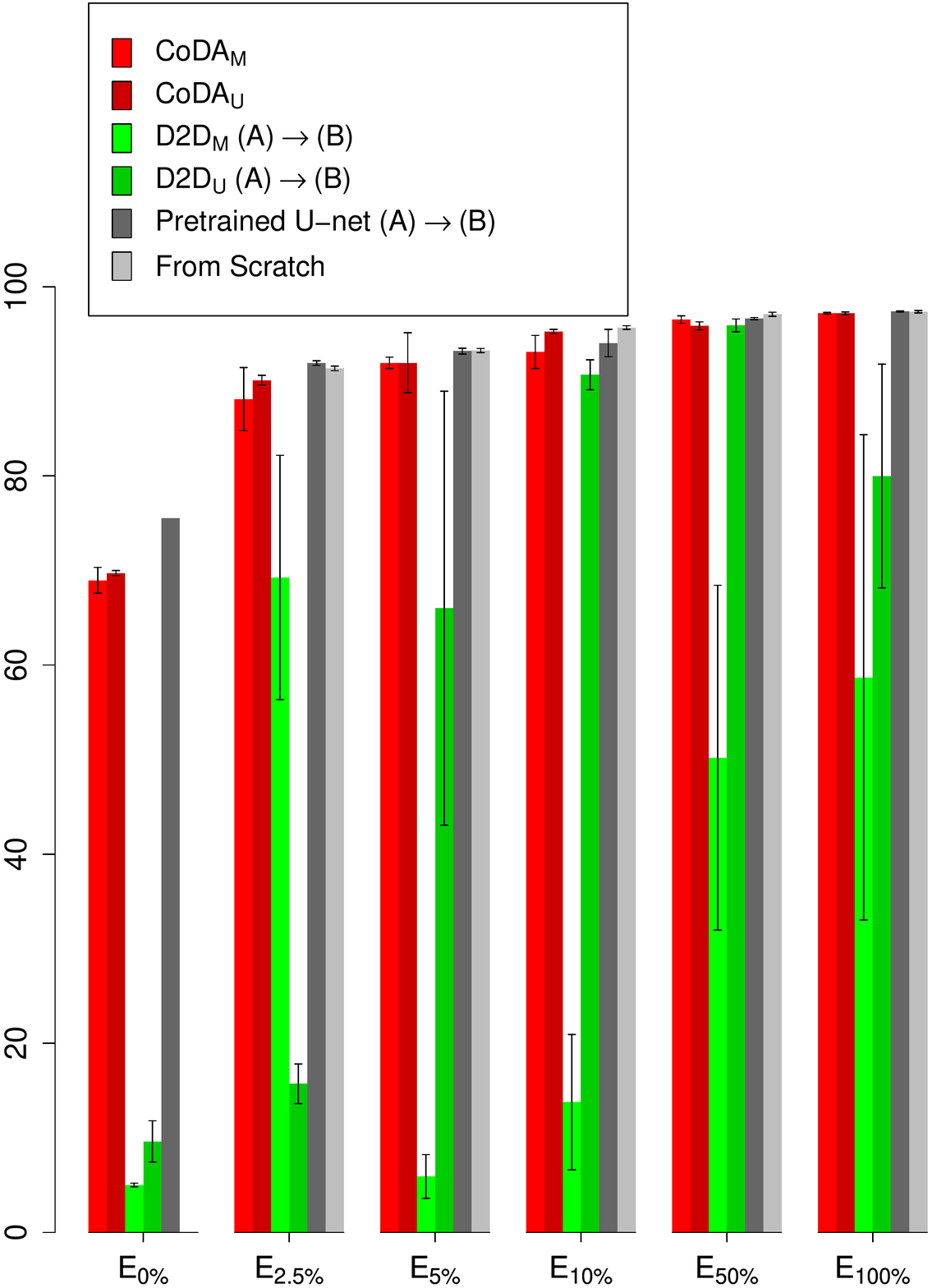}%
        \label{fig:intervals_mxr_f}
    }
    \caption{Confidence Intervals for MXRs according to the values shown in Tables~\ref{tab:results_pectoral} and~\ref{tab:results_breast}.}
    \label{fig:intervals_mxr}
\end{figure*}

Figure~\ref{fig:intervals_mxr} show the $J$ values from Tables~\ref{tab:results_pectoral} and~\ref{tab:results_breast} with confidence intervals for $p \le 0.05$ using a t-Student distribution.

A first noticeable trait in Figures~\ref{fig:intervals_mxr_a} and~\ref{fig:intervals_mxr_e} is that CoDAGANs maintained their capability to perform inference on the INbreast source dataset for both pectoral muscle and breast region experiments when labels from other sources are added to the procedure. D2D tends to get more unstable when the plots get closer to FSDA ($E_{100\%}$) due to the incongruities in labeling styles from the different datasets.

Figures~\ref{fig:intervals_mxr_b},~\ref{fig:intervals_mxr_c} and~\ref{fig:intervals_mxr_d} clearly show that CoDAGANs outperforms all baselines in UDA for the MIAS (B), DDSM B/C (C) and DDSM A (D) datasets by a large margin for pectoral muscle segmentation. All of these discrepancies between CoDAGANs and baselines are statistically significant, showing a clear superiority of CoDAGANs in UDA scenarios in this task. Figure~\ref{fig:intervals_mxr_f} show the UDA, SSDA and FSDA results for CoDAGANs and baselines on the target MIAS (B) dataset in the task of breast region segmentation. CoDAGANs yield considerably higher results than D2D, even though a Pretrained U-Net surpassed all methods in this task for UDA. Domain shifts between the MIAS and INbreast datasets are probably considerably small. Pretrained U-Nets might not be universally better than CoDAGANs in UDA, though, as it is usually unnable to compensate for large domain shifts. This trend was shown in Figures~\ref{fig:intervals_mxr_b},~\ref{fig:intervals_mxr_c} and~\ref{fig:intervals_mxr_d} and will be further reinforced in Section~\ref{sec:results_cxr}.

\subsection{Quantitative Results for CXR Samples}
\label{sec:results_cxr}

CXR results can be seen in Tables~\ref{tab:results_lungs} and~\ref{tab:results_heart_clavicles} for lungs, heart and clavicle segmentations. The JSRT (A), OpenIST (B), Shenzhen (C) and Montgomery (D) datasets are objectively evaluated in the lung field segmentation task, as shown in Table~\ref{tab:results_lungs}, while Chest X-Ray 8 (E), PadChest (F), NLMCXR (G) and OCT CXR (H) do not possess pixel-level ground truths for quantitative assessment. In heart and clavicle segmentation, apart from the source JSRT (A) dataset, only OpenIST (B) contains a subset of 15 labeled samples for these two task. Therefore, we reserved the labeled samples for testing and trained on the remaining samples for UDA quantitative assessment, as shown in Table~\ref{tab:results_heart_clavicles}. 
Analogously to Section~\ref{sec:results_mxr}, bold values in Tables~\ref{tab:results_lungs} and~\ref{tab:results_heart_clavicles} represent the best overall results in a given label configuration for a specific dataset.

\begin{table*}[!t]
    \centering
    \caption{Jaccard results (in \%) for lung field segmentation DA to and/or from eight distinct CXR datasets: JSRT (A), OpenIST (B), Shenzhen (C), Montgomery (D), Chest X-ray 8 (E), PadChest (F), NLMCXR (G) and OCT CXR (H).}
    \label{tab:results_lungs}
    \begin{tabular}{|c|c|c|c|c|c|c|c|}
        \hline
        \multicolumn{2}{|c|}{\textbf{Experiments}}                                       & $\bm{E_{0\%}}$            & $\bm{E_{2.5\%}}$          & $\bm{E_{5\%}}$            & $\bm{E_{10\%}}$           & $\bm{E_{50\%}}$           & $\bm{E_{100\%}}$          \\ \hhline{|==|=|=|=|=|=|=|}
        \multicolumn{2}{|c|}{\textbf{\% Labels JSRT (A)}}                                & 100.00\%                  & 100.00\%                  & 100.00\%                  & 100.00\%                  & 100.00\%                  & 100.00\%                  \\ \hline
        \multicolumn{2}{|c|}{\textbf{\% Labels OpenIST (B)}}                             & 0.00\%                    & 2.50\%                    & 5.00\%                    & 10.00\%                   & 50.00\%                   & 100.00\%                  \\ \hline
        \multicolumn{2}{|c|}{\textbf{\% Labels Shenzhen (C)}}                            & 0.00\%                    & 2.50\%                    & 5.00\%                    & 10.00\%                   & 50.00\%                   & 100.00\%                  \\ \hline
        \multicolumn{2}{|c|}{\textbf{\% Labels Montgomery (D)}}                          & 0.00\%                    & 2.50\%                    & 5.00\%                    & 10.00\%                   & 50.00\%                   & 100.00\%                  \\ \hline
        \multicolumn{2}{|c|}{\textbf{\% Labels ChestX-Ray8 (E)}}                         & 0.00\%                    & 0.00\%                    & 0.00\%                    & 0.00\%                    & 0.00\%                    & 0.00\%                    \\ \hline
        \multicolumn{2}{|c|}{\textbf{\% Labels PadChest (F)}}                            & 0.00\%                    & 0.00\%                    & 0.00\%                    & 0.00\%                    & 0.00\%                    & 0.00\%                    \\ \hline
        \multicolumn{2}{|c|}{\textbf{\% Labels NLMCXR (G)}}                              & 0.00\%                    & 0.00\%                    & 0.00\%                    & 0.00\%                    & 0.00\%                    & 0.00\%                    \\ \hline
        \multicolumn{2}{|c|}{\textbf{\% Labels OCT CXR (H)}}                             & 0.00\%                    & 0.00\%                    & 0.00\%                    & 0.00\%                    & 0.00\%                    & 0.00\%                    \\ \hhline{|==|=|=|=|=|=|=|}
        \multirow{17}{*}{\textbf{(A)}} & $\bm{CoDA_{M}}$                                 & 95.27 $\pm$ 0.07          & 94.08 $\pm$ 0.44          & 94.41 $\pm$ 0.86          & 94.74 $\pm$ 0.12          & 94.87 $\pm$ 0.39          & 95.31 $\pm$ 0.17          \\ \cline{2-8} 
                                       & $\bm{CoDA_{U}}$                                 & 95.55 $\pm$ 0.07          & 95.16 $\pm$ 0.12          & 94.77 $\pm$ 0.17          & 94.83 $\pm$ 0.08          & 94.56 $\pm$ 0.99          & 94.97 $\pm$ 0.62          \\ \cline{2-8} 
                                       & \textbf{$\bm{D2D_{M}}$ (A)$\rightarrow$(B)} & 96.39 $\pm$ 0.06          & \textbf{96.51 $\pm$ 0.07} & \textbf{96.54 $\pm$ 0.07} & \textbf{96.43 $\pm$ 0.03} & 93.96 $\pm$ 4.85          & \textbf{96.34 $\pm$ 0.12} \\ \cline{2-8} 
                                       & \textbf{$\bm{D2D_{U}}$ (A)$\rightarrow$(B)} & 96.16 $\pm$ 0.35          & 96.41 $\pm$ 0.06          & 96.43 $\pm$ 0.04          & 96.26 $\pm$ 0.17          & 96.22 $\pm$ 0.29          & 96.20 $\pm$ 0.19          \\ \cline{2-8} 
                                       & \textbf{$\bm{D2D_{M}}$ (A)$\rightarrow$(C)} & \textbf{96.45 $\pm$ 0.02} & 96.44 $\pm$ 0.05          & 96.07 $\pm$ 0.61          & \textbf{96.43 $\pm$ 0.10} & \textbf{96.29 $\pm$ 0.07} & 96.20 $\pm$ 0.10          \\ \cline{2-8} 
                                       & \textbf{$\bm{D2D_{U}}$ (A)$\rightarrow$(C)} & 96.10 $\pm$ 0.51          & 96.34 $\pm$ 0.03          & 96.04 $\pm$ 0.35          & 96.33 $\pm$ 0.07          & 96.23 $\pm$ 0.04          & 96.33 $\pm$ 0.06          \\ \cline{2-8} 
                                       & \textbf{$\bm{D2D_{M}}$ (A)$\rightarrow$(D)} & 96.23 $\pm$ 0.21          & 96.22 $\pm$ 0.08          & 96.20 $\pm$ 0.09          & 96.24 $\pm$ 0.09          & 96.16 $\pm$ 0.21          & 95.84 $\pm$ 0.70          \\ \cline{2-8} 
                                       & \textbf{$\bm{D2D_{U}}$ (A)$\rightarrow$(D)} & 96.26 $\pm$ 0.11          & 96.21 $\pm$ 0.06          & 96.22 $\pm$ 0.13          & 96.21 $\pm$ 0.17          & 96.22 $\pm$ 0.10          & 96.02 $\pm$ 0.13          \\ \cline{2-8} 
                                       & \textbf{$\bm{D2D_{M}}$ (A)$\rightarrow$(E)} & 96.35 $\pm$ 0.19          & --                        & --                        & --                        & --                        & --                        \\ \cline{2-8} 
                                       & \textbf{$\bm{D2D_{U}}$ (A)$\rightarrow$(E)} & 96.42 $\pm$ 0.09          & --                        & --                        & --                        & --                        & --                        \\ \cline{2-8} 
                                       & \textbf{$\bm{D2D_{M}}$ (A)$\rightarrow$(F)} & 96.38 $\pm$ 0.15          & --                        & --                        & --                        & --                        & --                        \\ \cline{2-8} 
                                       & \textbf{$\bm{D2D_{U}}$ (A)$\rightarrow$(F)} & 96.30 $\pm$ 0.09          & --                        & --                        & --                        & --                        & --                        \\ \cline{2-8} 
                                       & \textbf{$\bm{D2D_{M}}$ (A)$\rightarrow$(G)} & 96.11 $\pm$ 0.65          & --                        & --                        & --                        & --                        & --                        \\ \cline{2-8} 
                                       & \textbf{$\bm{D2D_{U}}$ (A)$\rightarrow$(G)} & 96.34 $\pm$ 0.09          & --                        & --                        & --                        & --                        & --                        \\ \cline{2-8} 
                                       & \textbf{$\bm{D2D_{M}}$ (A)$\rightarrow$(H)} & 96.24 $\pm$ 0.56          & --                        & --                        & --                        & --                        & --                        \\ \cline{2-8} 
                                       & \textbf{$\bm{D2D_{U}}$ (A)$\rightarrow$(H)} & 95.91 $\pm$ 1.02          & --                        & --                        & --                        & --                        & --                        \\ \cline{2-8} 
                                       & \textbf{From Scratch in (A)}                    & 95.70 $\pm$ 0.06          & --                        & --                        & --                        & --                        & --                        \\ \hhline{|==|=|=|=|=|=|=|}
        \multirow{6}{*}{\textbf{(B)}}  & $\bm{CoDA_{M}}$                                 & 90.67 $\pm$ 0.80          & 92.58 $\pm$ 0.59          & 92.83 $\pm$ 1.25          & \textbf{93.41 $\pm$ 0.49} & 93.65 $\pm$ 1.07          & 94.71 $\pm$ 0.15          \\ \cline{2-8} 
                                       & $\bm{CoDA_{U}}$                                 & \textbf{91.03 $\pm$ 0.96} & 92.08 $\pm$ 0.37          & \textbf{93.50 $\pm$ 0.41} & 93.32 $\pm$ 0.16          & 94.23 $\pm$ 0.41          & 94.63 $\pm$ 0.25          \\ \cline{2-8} 
                                       & \textbf{$\bm{D2D_{M}}$ (A)$\rightarrow$(B)} & 19.67 $\pm$ 28.59         & \textbf{92.79 $\pm$ 1.68} & 93.41 $\pm$ 0.65          & 92.15 $\pm$ 1.26          & 93.22 $\pm$ 1.79          & 93.46 $\pm$ 1.35          \\ \cline{2-8} 
                                       & \textbf{$\bm{D2D_{U}}$ (A)$\rightarrow$(B)} & 56.82 $\pm$ 31.88         & 70.11 $\pm$ 12.05         & 88.87 $\pm$ 5.33          & 61.40 $\pm$ 28.25         & 93.94 $\pm$ 0.82          & \textbf{94.95 $\pm$ 0.50} \\ \cline{2-8} 
                                       & \textbf{Pretrained (A)$\rightarrow$(B)}   & 7.48                      & 83.91 $\pm$ 0.17          & 90.37 $\pm$ 0.15          & 92.47 $\pm$ 0.16          & 94.37 $\pm$ 0.17          & 94.88 $\pm$ 0.06          \\ \cline{2-8} 
                                       & \textbf{From Scratch in (B)}                    & --                        & 85.69 $\pm$ 0.33          & 88.94 $\pm$ 0.56          & 91.70 $\pm$ 0.22          & \textbf{94.87 $\pm$ 0.10} & 94.35 $\pm$ 0.12          \\ \hhline{|==|=|=|=|=|=|=|}
        \multirow{6}{*}{\textbf{(C)}}  & $\bm{CoDA_{M}}$                                 & 88.69 $\pm$ 0.46          & \textbf{89.88 $\pm$ 0.36} & \textbf{90.75 $\pm$ 0.49} & 90.64 $\pm$ 0.23          & 90.99 $\pm$ 0.89          & 91.84 $\pm$ 0.09          \\ \cline{2-8} 
                                       & $\bm{CoDA_{U}}$                                 & \textbf{88.99 $\pm$ 0.29} & 89.74 $\pm$ 0.12          & 89.90 $\pm$ 0.61          & 90.04 $\pm$ 0.41          & 91.35 $\pm$ 0.49          & 91.61 $\pm$ 0.49          \\ \cline{2-8} 
                                       & \textbf{$\bm{D2D_{M}}$ (A)$\rightarrow$(C)} & 70.01 $\pm$ 8.67          & 89.45 $\pm$ 0.97          & 83.46 $\pm$ 10.87         & \textbf{91.42 $\pm$ 0.57} & 91.61 $\pm$ 0.68          & 92.03 $\pm$ 0.80          \\ \cline{2-8} 
                                       & \textbf{$\bm{D2D_{U}}$ (A)$\rightarrow$(C)} & 55.40 $\pm$ 33.75         & 82.72 $\pm$ 4.42          & 80.83 $\pm$ 11.11         & 89.02 $\pm$ 3.47          & 91.82 $\pm$ 0.31          & 91.99 $\pm$ 0.19          \\ \cline{2-8} 
                                       & \textbf{Pretrained (A)$\rightarrow$(C)}   & 17.19                     & 88.68 $\pm$ 0.16          & 90.82 $\pm$ 0.07          & 91.60 $\pm$ 0.10          & 92.17 $\pm$ 0.15          & \textbf{92.40 $\pm$ 0.03} \\ \cline{2-8} 
                                       & \textbf{From Scratch in (C)}                    & --                        & 89.62 $\pm$ 0.20          & 90.74 $\pm$ 0.35          & 91.79 $\pm$ 0.08          & \textbf{92.32 $\pm$ 0.04} & 92.25 $\pm$ 0.05          \\ \hhline{|==|=|=|=|=|=|=|}
        \multirow{6}{*}{\textbf{(D)}}  & $\bm{CoDA_{M}}$                                 & 81.88 $\pm$ 1.35          & \textbf{87.86 $\pm$ 0.88} & 87.72 $\pm$ 1.60          & 90.48 $\pm$ 0.64          & 93.15 $\pm$ 0.44          & 94.19 $\pm$ 0.31          \\ \cline{2-8} 
                                       & $\bm{CoDA_{U}}$                                 & \textbf{84.58 $\pm$ 1.48} & 87.12 $\pm$ 0.59          & 87.07 $\pm$ 0.78          & 87.75 $\pm$ 0.81          & 92.76 $\pm$ 2.27          & 92.95 $\pm$ 1.83          \\ \cline{2-8} 
                                       & \textbf{$\bm{D2D_{M}}$ (A)$\rightarrow$(D)} & 30.20 $\pm$ 26.08         & 82.60 $\pm$ 4.05          & \textbf{88.34 $\pm$ 2.99} & 80.48 $\pm$ 8.89          & 93.46 $\pm$ 0.81          & 93.51 $\pm$ 0.99          \\ \cline{2-8} 
                                       & \textbf{$\bm{D2D_{U}}$ (A)$\rightarrow$(D)} & 79.44 $\pm$ 5.64          & 64.61 $\pm$ 8.95          & 76.89 $\pm$ 1.52          & 82.33 $\pm$ 5.01          & 94.02 $\pm$ 0.27          & 94.23 $\pm$ 0.66          \\ \cline{2-8} 
                                       & \textbf{Pretrained (A)$\rightarrow$(D)}   & 10.79                     & 81.47 $\pm$ 0.05          & 86.79 $\pm$ 0.07          & 89.60 $\pm$ 0.14          & 94.40 $\pm$ 0.07          & 94.82 $\pm$ 0.05          \\ \cline{2-8} 
                                       & \textbf{From Scratch in (D)}                    & --                        & 76.32 $\pm$ 0.27          & 87.12 $\pm$ 0.20          & \textbf{90.91 $\pm$ 0.25} & \textbf{94.66 $\pm$ 0.09} & \textbf{95.19 $\pm$ 0.12} \\ \hline
    \end{tabular}
\end{table*}

\begin{table}[!t]
    \centering
    \caption{Jaccard results (in \%) for heart and clavicle segmentation DA to and/or from eight distinct CXR datasets: JSRT (A), OpenIST (B), Shenzhen (C), Montgomery (D), Chest X-ray 8 (E), PadChest (F), NLMCXR (G) and OCT CXR (H).}
    \label{tab:results_heart_clavicles}
    \begin{tabular}{|c|c|c|c|}
        \hline
        \multicolumn{2}{|c|}{\textbf{Experiments}}                                       & \textbf{$\bm{E_{0\%}}$ (Heart)} & \textbf{$\bm{E_{0\%}}$ (Clavicles)} \\ \hhline{|==|=|=|}
        \multicolumn{2}{|c|}{\textbf{\% Labels JSRT (A)}}                                & 100.00\%                        & 100.00\%                            \\ \hline
        \multicolumn{2}{|c|}{\textbf{\% Labels OpenIST (B)}}                             & 0.00\%                          & 0.00\%                              \\ \hline
        \multicolumn{2}{|c|}{\textbf{\% Labels Shenzhen (C)}}                            & 0.00\%                          & 0.00\%                              \\ \hline
        \multicolumn{2}{|c|}{\textbf{\% Labels Montgomery (D)}}                          & 0.00\%                          & 0.00\%                              \\ \hline
        \multicolumn{2}{|c|}{\textbf{\% Labels ChestX-Ray8 (E)}}                         & 0.00\%                          & 0.00\%                              \\ \hline
        \multicolumn{2}{|c|}{\textbf{\% Labels PadChest (F)}}                            & 0.00\%                          & 0.00\%                              \\ \hline
        \multicolumn{2}{|c|}{\textbf{\% Labels NLMCXR (G)}}                              & 0.00\%                          & 0.00\%                              \\ \hline
        \multicolumn{2}{|c|}{\textbf{\% Labels OCT CXR (H)}}                             & 0.00\%                          & 0.00\%                              \\ \hhline{|==|=|=|}
        \multirow{17}{*}{\textbf{(A)}} & $\bm{CoDA_{M}}$                                 & 89.86 $\pm$ 0.29                & 77.31 $\pm$ 0.37                    \\ \cline{2-4} 
                                       & $\bm{CoDA_{U}}$                                 & 89.89 $\pm$ 0.32                & 76.03 $\pm$ 1.06                    \\ \cline{2-4} 
                                       & \textbf{$\bm{D2D_{M}}$ (A)$\rightarrow$(B)} & 90.68 $\pm$ 0.19                & 87.76 $\pm$ 0.18                    \\ \cline{2-4} 
                                       & \textbf{$\bm{D2D_{U}}$ (A)$\rightarrow$(B)} & 90.97 $\pm$ 0.10                & \textbf{87.96 $\pm$ 0.14}           \\ \cline{2-4} 
                                       & \textbf{$\bm{D2D_{M}}$ (A)$\rightarrow$(C)} & 91.16 $\pm$ 0.18                & 87.20 $\pm$ 0.22                    \\ \cline{2-4} 
                                       & \textbf{$\bm{D2D_{U}}$ (A)$\rightarrow$(C)} & 90.70 $\pm$ 0.48                & 87.09 $\pm$ 0.61                    \\ \cline{2-4} 
                                       & \textbf{$\bm{D2D_{M}}$ (A)$\rightarrow$(D)} & 90.65 $\pm$ 0.17                & 84.78 $\pm$ 0.53                    \\ \cline{2-4} 
                                       & \textbf{$\bm{D2D_{U}}$ (A)$\rightarrow$(D)} & 90.67 $\pm$ 0.23                & 84.46 $\pm$ 1.24                    \\ \cline{2-4} 
                                       & \textbf{$\bm{D2D_{M}}$ (A)$\rightarrow$(E)} & 90.97 $\pm$ 0.20                & 87.38 $\pm$ 0.21                    \\ \cline{2-4} 
                                       & \textbf{$\bm{D2D_{U}}$ (A)$\rightarrow$(E)} & 91.11 $\pm$ 0.12                & 87.02 $\pm$ 0.63                    \\ \cline{2-4} 
                                       & \textbf{$\bm{D2D_{M}}$ (A)$\rightarrow$(F)} & 89.74 $\pm$ 2.32                & 87.63 $\pm$ 0.55                    \\ \cline{2-4} 
                                       & \textbf{$\bm{D2D_{U}}$ (A)$\rightarrow$(F)} & 90.63 $\pm$ 0.37                & 87.39 $\pm$ 0.47                    \\ \cline{2-4} 
                                       & \textbf{$\bm{D2D_{M}}$ (A)$\rightarrow$(G)} & 91.15 $\pm$ 0.18                & 87.59 $\pm$ 0.06                    \\ \cline{2-4} 
                                       & \textbf{$\bm{D2D_{U}}$ (A)$\rightarrow$(G)} & 90.90 $\pm$ 0.10                & 87.37 $\pm$ 0.59                    \\ \cline{2-4} 
                                       & \textbf{$\bm{D2D_{M}}$ (A)$\rightarrow$(H)} & \textbf{91.33 $\pm$ 0.16}       & 86.46 $\pm$ 0.62                    \\ \cline{2-4} 
                                       & \textbf{$\bm{D2D_{U}}$ (A)$\rightarrow$(H)} & 90.22 $\pm$ 0.94                & 84.49 $\pm$ 0.64                    \\ \cline{2-4} 
                                       & \textbf{From Scratch in (A)}                    & 88.91 $\pm$ 0.48                & 76.07 $\pm$ 0.41                    \\ \hhline{|==|=|=|}
        \multirow{5}{*}{\textbf{(B)}}  & $\bm{CoDA_{M}}$                                 & \textbf{64.63 $\pm$ 1.28}       & 61.94 $\pm$ 1.03                    \\ \cline{2-4} 
                                       & $\bm{CoDA_{U}}$                                 & 63.71 $\pm$ 0.95                & 57.59 $\pm$ 1.26                    \\ \cline{2-4} 
                                       & \textbf{$\bm{D2D_{M}}$ (A)$\rightarrow$(B)} & 54.91 $\pm$ 2.35                & 67.11 $\pm$ 0.96                    \\ \cline{2-4} 
                                       & \textbf{$\bm{D2D_{U}}$ (A)$\rightarrow$(B)} & 64.50 $\pm$ 0.84                & \textbf{68.53 $\pm$ 0.89}           \\ \cline{2-4} 
                                       & \textbf{Pretrained (A)$\rightarrow$(B)}   & 0.0 $\pm$ 0.0                   & 0.24 $\pm$ 0.40                     \\ \hline
    \end{tabular}
\end{table}

\subsubsection{Lung Segmentation in CXR Images}
\label{sec:results_lungs}

In the task of lung segmentation in CXRs (Table~\ref{tab:results_lungs}), baselines showed considerably poor results for target datasets (B)-(D) in UDA experiments. Following the results from Sections~\ref{sec:results_pectoral} and~\ref{sec:results_breast}, D2D with a small amount of target labels proved to be highly unstable, yielding worse results and considerably higher standard deviations, when compared with CoDAGANs. $\bm{CoDA_{M}}$ and $\bm{CoDA_{U}}$ achieve the best UDA results in (B), (C) and (D), surpassing all baselines by a considerable margin, yielding $J$ values of 91.03\%, 88.99\% and 84.58\% for these three datasets, respectively. Pretrained U-Nets yielded worse than random results in these tasks, which can be expĺained by the high domain shift across (A)$\rightarrow$(B), (A)$\rightarrow$(C) and (A)$\rightarrow$(D).

CoDAGANs maintain state-of-the-art results in SSDA experiments with small amount of labels, surpassing baselines in most datasets for $E_{2.5\%}$ and $E_{5\%}$. In $E_{50\%}$ and $E_{100\%}$ state-of-the-art results are achieved mainly by From Scratch training in the target domain due to label abundance. Similarly, D2D methods are only able to achieve stable results, after $E_{10\%}$. As in MXRs, D2D underperformed in UDA settings compared to CoDAGANs, even though it presented considerably better results than Pretrained DNNs.

We also show that the source dataset presented little to no deterioration in segmentation quality when segmented by CoDAGANs compared to D2D and From Scratch training on (A). D2D from translations (A)$\rightarrow$(B) to (A)$\rightarrow$(H) present remarkably similar results in UDA, SSDA and FSDA, achieving state-of-the-art results in all cases. It is noticeable that CoDAGANs achieved no superiority in the source domain, as it aims for generalization and does not focus in fine-grained segmentation. However, the difference of Jaccard values between CoDAGANs and baseline methods that only consider a pair of domains or even only the source domain (From Scratch) remained limited to between 1\% and 2\%.


\subsubsection{Heart and Clavicle Segmentation in CXR Images}
\label{sec:results_heart_clavicles}

As shown in Table~\ref{tab:results_heart_clavicles}, heart and clavicle segmentation proved to be harder tasks than lung field segmentation. Both tasks only count with the JSRT dataset as fully labeled, with OpenIST having only 15 images with pixel-level annotations for both heart and clavicles. We therefore used these samples only for evaluating UDA in a target dataset, as the small number of samples would not allow for proper SSDA and FSDA experiments. $CoDA_{M}$ achieved the best results in heart segmentation on OpenIST with a $J$ value of 64.63\%, closely followed by $D2D_{U}$ with $64.50\%$. Clavicle segmentation topped on $68.53\%$ for D2D and was the only task that clearly showed an underperformance of CoDAGANs compared with D2D, achieving only $61.94\%$. Both D2D and CoDAGANs greatly surpassed the Pretrained U-Net in both tasks for (B), with the pretrained baseline achieving close to $0\%$ in Jaccard.

Table~\ref{tab:results_heart_clavicles} also shows the remarkable stability of D2D for the source dataset (A), evidencing that performing DA using Image Translation does not compromise performance in the source domain. CoDAGANs closely followed the performance of D2D in the source dataset (A) for heart segmentation, but again showed considerably worse performance in clavicle segmentation. This underperformance of CoDAGANs in clavicle segmentation for both datasets is probably explained by the higher imbalance of this task. Clavicles cover a much smaller area in a CXR than lungs or a heart and, therefore, are more susceptible to low performance in segmentation DNNs that contain fewer skip connections, as the case of the truncated asymmetrical U-Net configured to receive data from the isomorphic representation $I$ in CoDAGANs.

\subsubsection{CXR Segmentation Confidence Intervals}
\label{sec:results_cxr_ci}

\begin{figure*}[!t]
    \centering
    \renewcommand{\currprop}{0.32\textwidth}
    \subfloat[]{
        \includegraphics[width=\currprop]{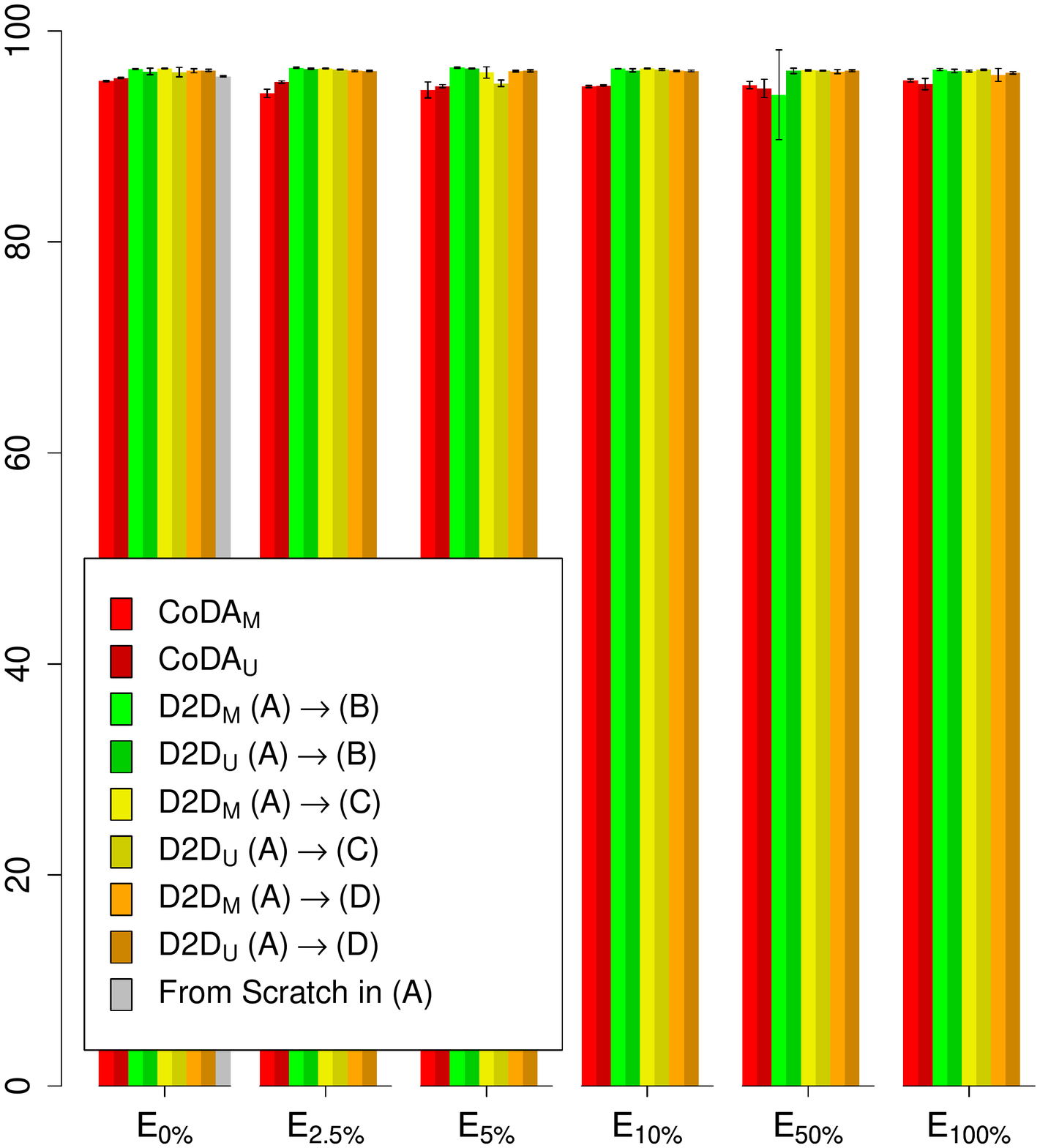}%
        \label{fig:intervals_cxr_a}
    }
    \subfloat[]{
        \includegraphics[width=\currprop]{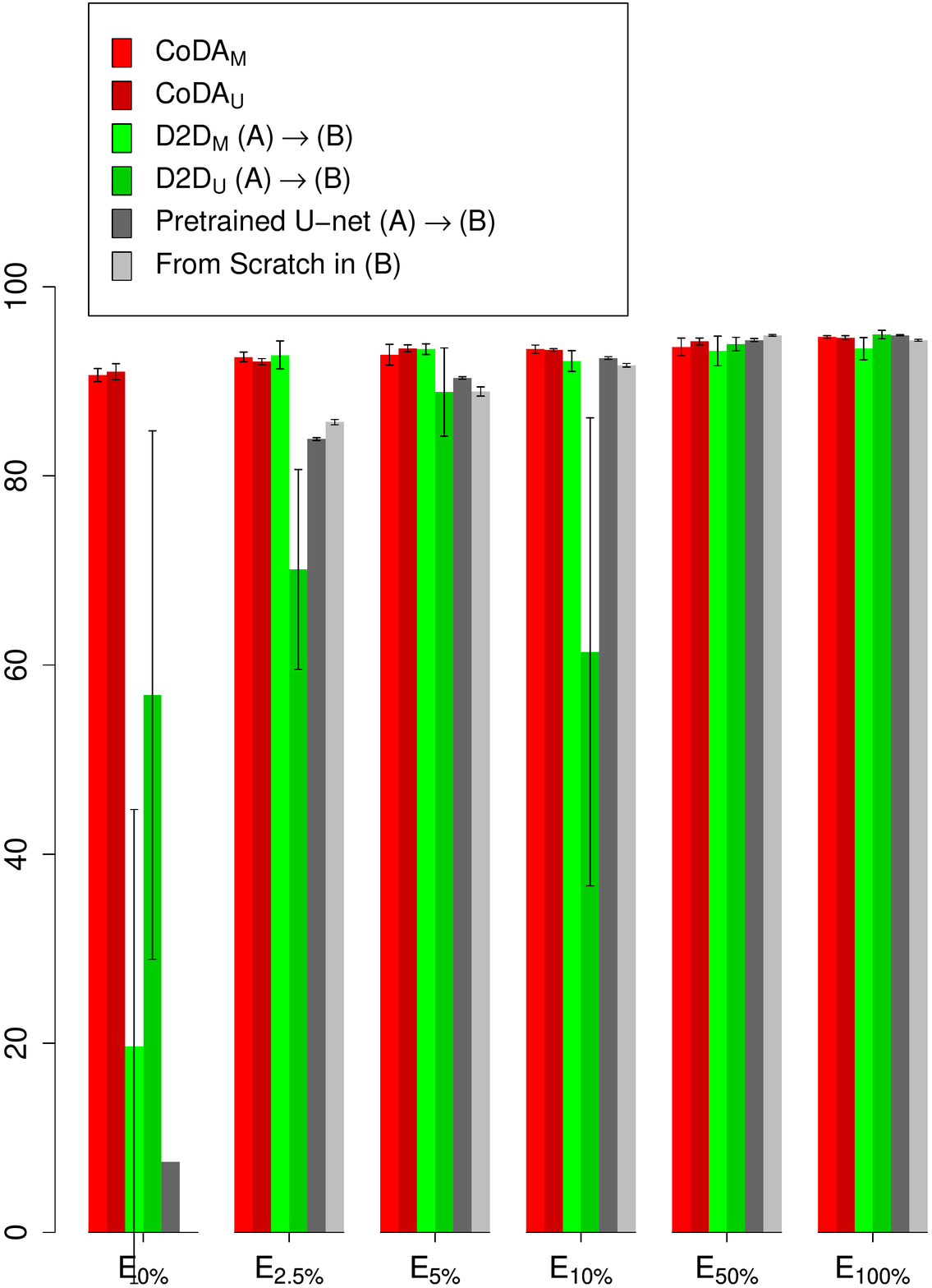}%
        \label{fig:intervals_cxr_b}
    }
    \subfloat[]{
        \includegraphics[width=\currprop]{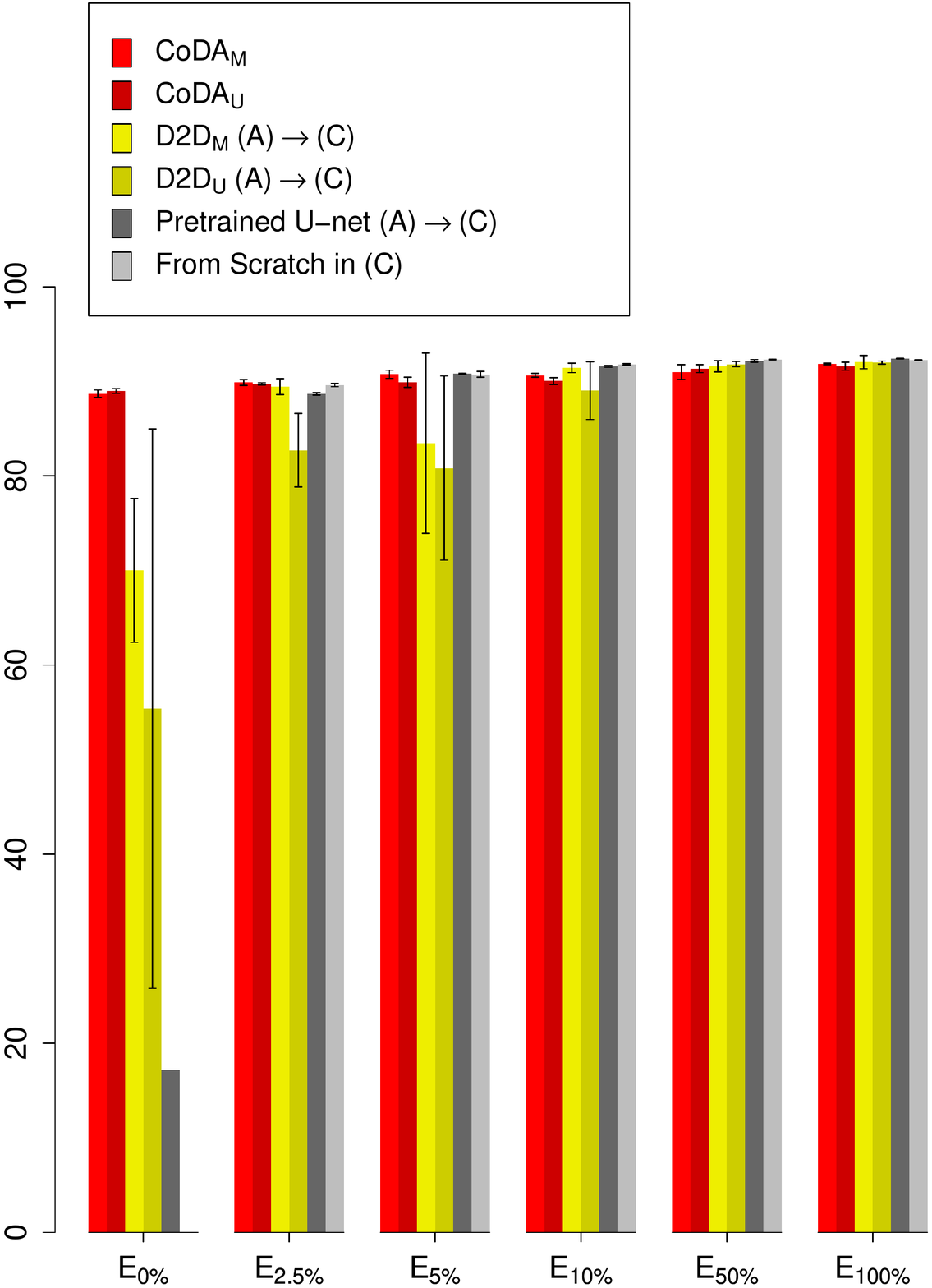}%
        \label{fig:intervals_cxr_c}
    }
    \hfil
    \subfloat[]{
        \includegraphics[width=\currprop]{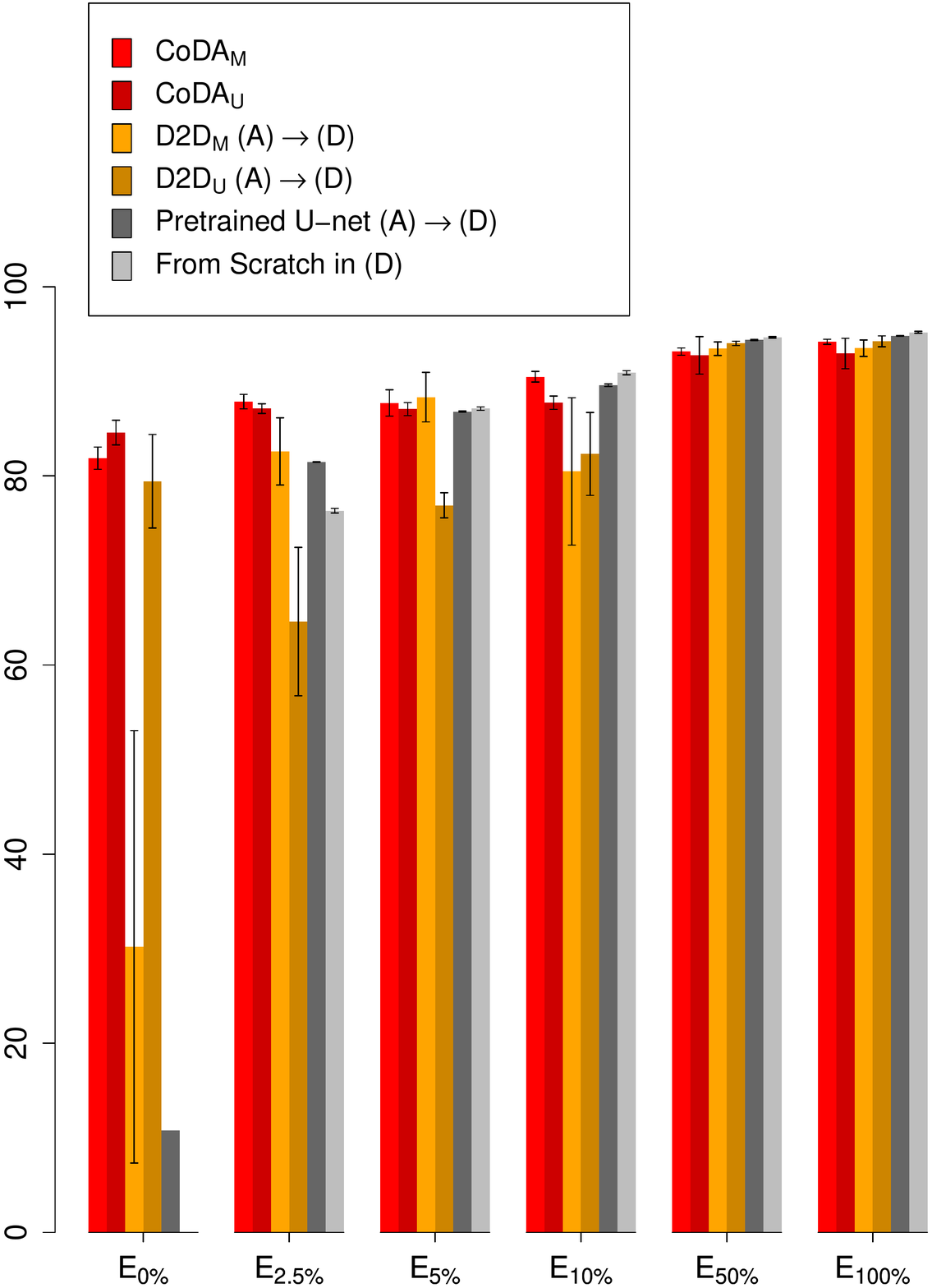}%
        \label{fig:intervals_cxr_d}
    }
    \subfloat[]{
        \includegraphics[width=\currprop]{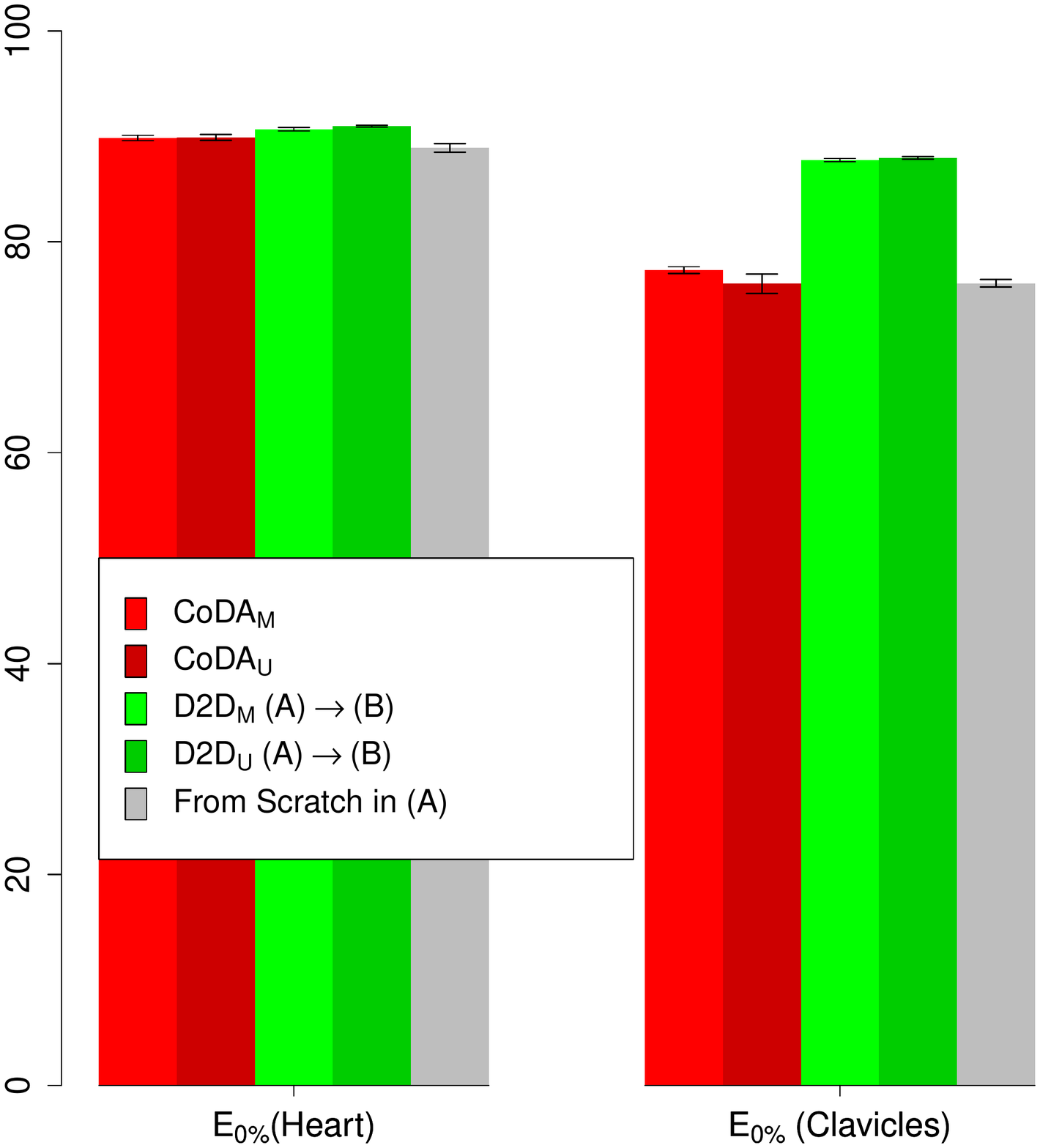}%
        \label{fig:intervals_cxr_e}
    }
    \subfloat[]{
        \includegraphics[width=\currprop]{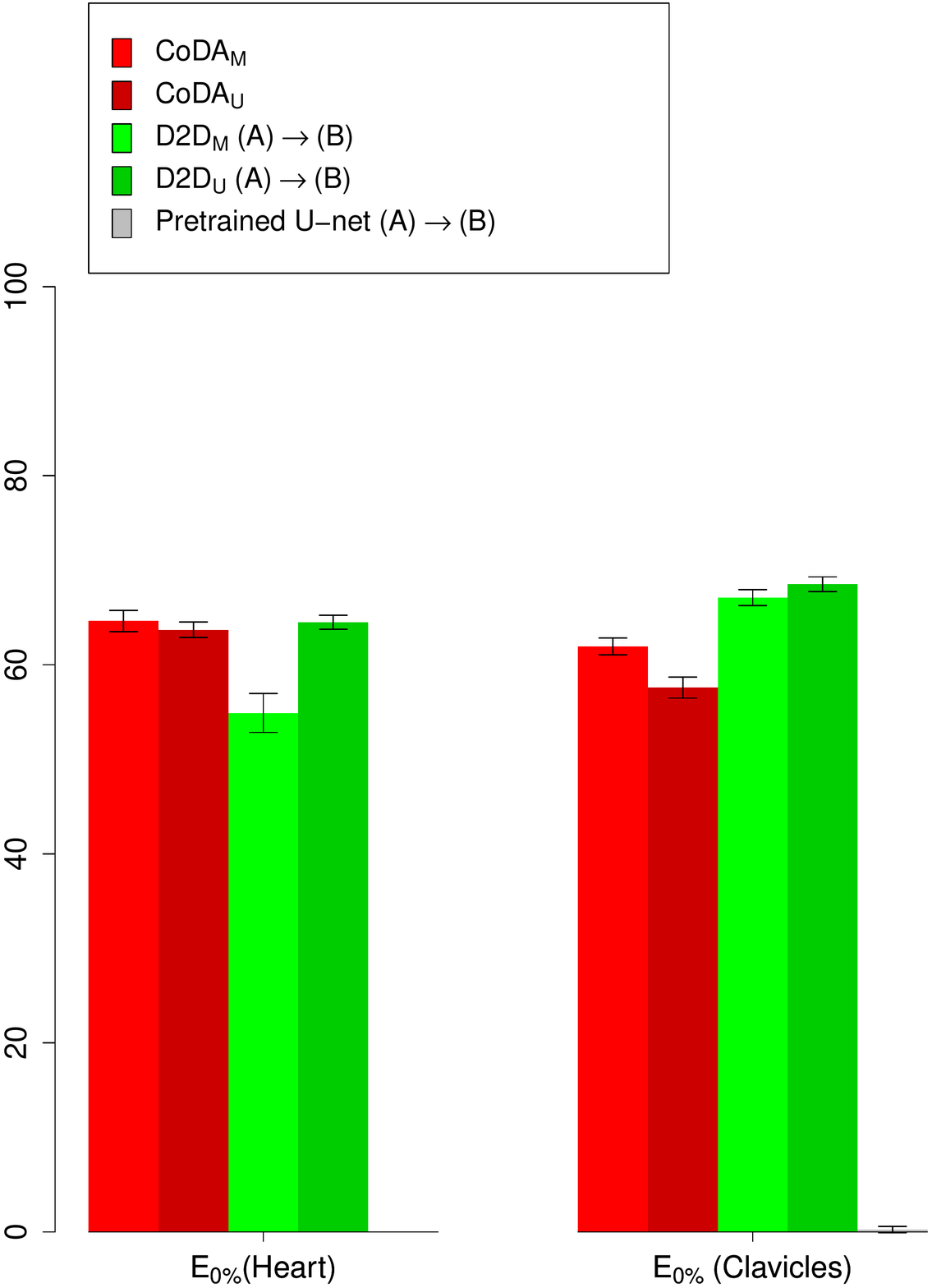}%
        \label{fig:intervals_cxr_f}
    }
    \caption{Confidence Intervals for CXRs according to the values shown in Tables~\ref{tab:results_lungs} and~\ref{tab:results_heart_clavicles}.}
    \label{fig:intervals_cxr}
\end{figure*}

Figure~\ref{fig:intervals_cxr} shows the confidence intervals for $p \le 0.05$ in lung segmentation for both the source JSRT dataset (Figure~\ref{fig:intervals_cxr_a}) and the target image sets (Figures~\ref{fig:intervals_cxr_b},~\ref{fig:intervals_cxr_c} and~\ref{fig:intervals_cxr_d}). Figures~\ref{fig:intervals_cxr_e} and~\ref{fig:intervals_cxr_f} show the results for heart and clavicle segmentation in the source (JSRT) and target (OpenIST) datasets, respectively.

One can see by Figures~\ref{fig:intervals_cxr_a} and~\ref{fig:intervals_cxr_e} that segmentation in the source dataset is preserved even when labels from other datasets are introduced in the training procedure. Figures~\ref{fig:intervals_cxr_b},~\ref{fig:intervals_cxr_c},~\ref{fig:intervals_cxr_b} and~\ref{fig:intervals_cxr_f} show the UDA, SSDA and FSDA efficiency of CoDAGANs in the labeled target datasets, that is, OpenIST, Shenzhen and Montgomery for lung segmentation and only OpenIST for heart and clavicles.

Figure~\ref{fig:intervals_cxr_b} evidences that both training from scratch and transfer learning using MMD, fixed pretrained models and fine-tuning are suboptimal when there is scarce labeled data in the target domain. On $E_{0\%}$ for mammogram segmentation, MMD on hardly improves segmentation results compared to simply using the fixed pretrained networks from other datasets, reinforcing that this method does not properly work for dense labeling problems. CoDAGANs are shown to be significantly more robust in these scarce label scenarios, achieving close to 90\% of jaccard in the target Montgomery dataset even for the completely unlabeled transfer experiment ($E_{0\%}$). This test case shows the inability of MMD and pretrained models to compensate for the domain shift between the two datasets in these dense labeling tasks, with both methods achieving jaccard indices of less than 10\% in $E_{0\%}$. Training from scratch and fine-tuning only achieve competitive results in $E_{50\%}$ and $E_{100\%}$, where there are abundant target labels.



\subsection{Segmentation Qualitative Analysis}
\label{sec:results_qualitative}

\begin{figure*}[!t]
    \centering
    \renewcommand{\currprop}{0.9\columnwidth}
    \subfloat[]{
        \includegraphics[width=\currprop]{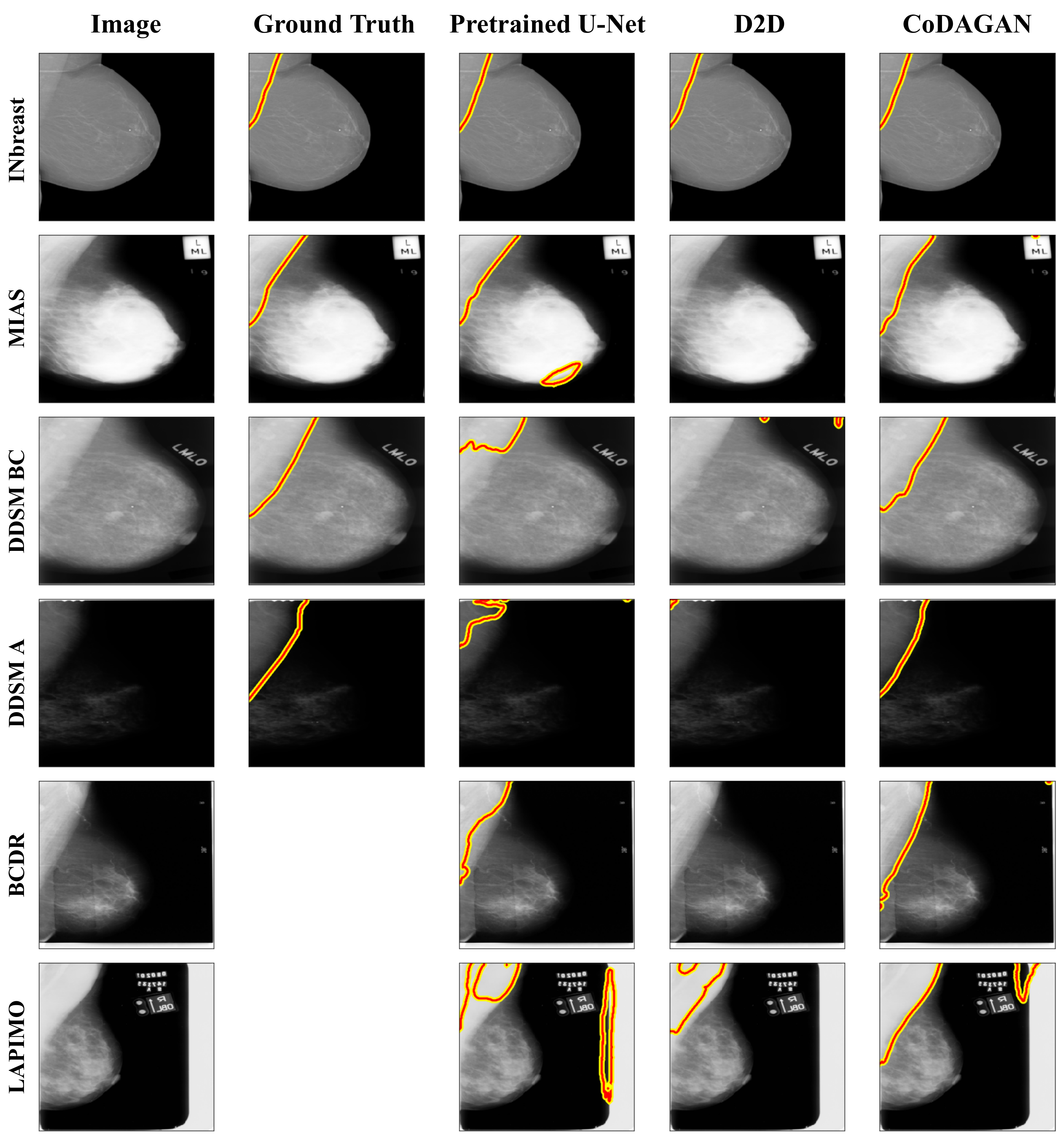}
        \label{fig:segmentation_mxr_pectoral}
    }%
    \hfil
    \subfloat[]{
        \includegraphics[width=\currprop]{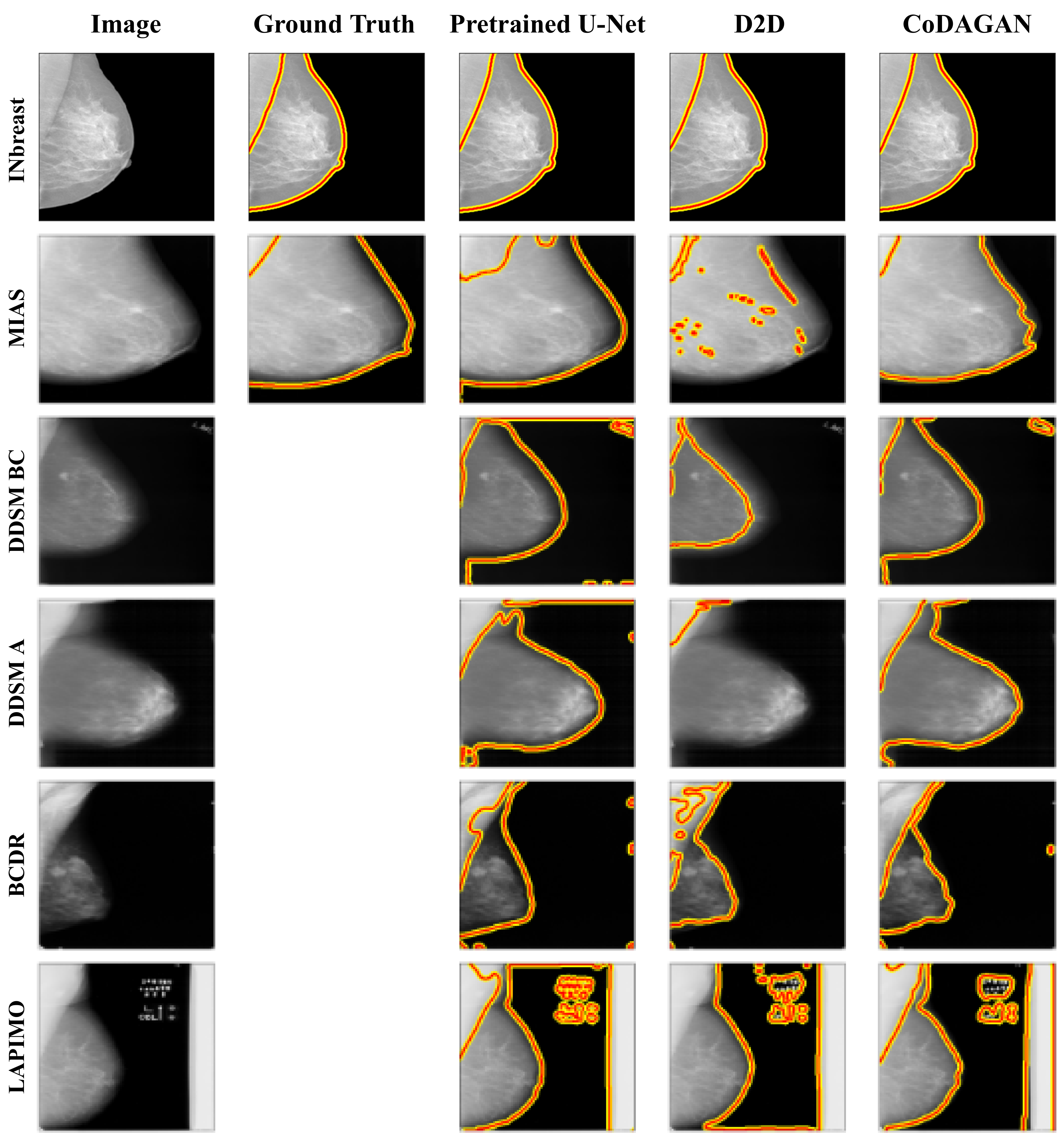}
        \label{fig:segmentation_mxr_breast}
    }%
    \caption{Qualitative segmentation results in mammographic images for two distinct tasks: $\bm{E_{0\%}}$ pectoral (a) and $\bm{E_{0\%}}$ breast region (b).}
    \label{fig:segmentation_mxr}
\end{figure*}

\begin{figure*}[!t]
    \centering
    \renewcommand{\currprop}{0.9\columnwidth}
    \subfloat[]{
        \includegraphics[width=\currprop]{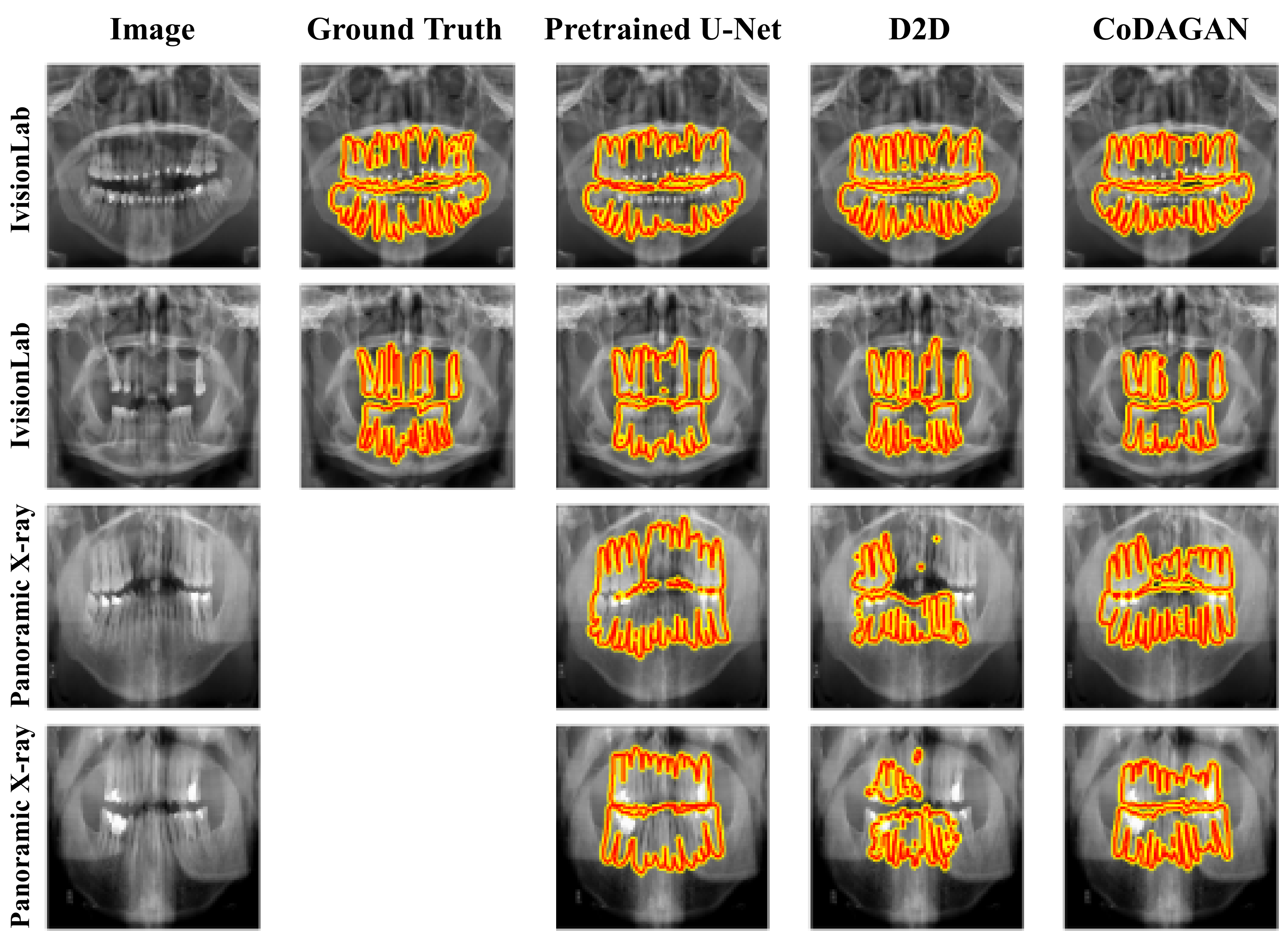}
        \label{fig:segmentation_dxr_teeth}
    }%
    \hfil
    \subfloat[]{
        \includegraphics[width=\currprop]{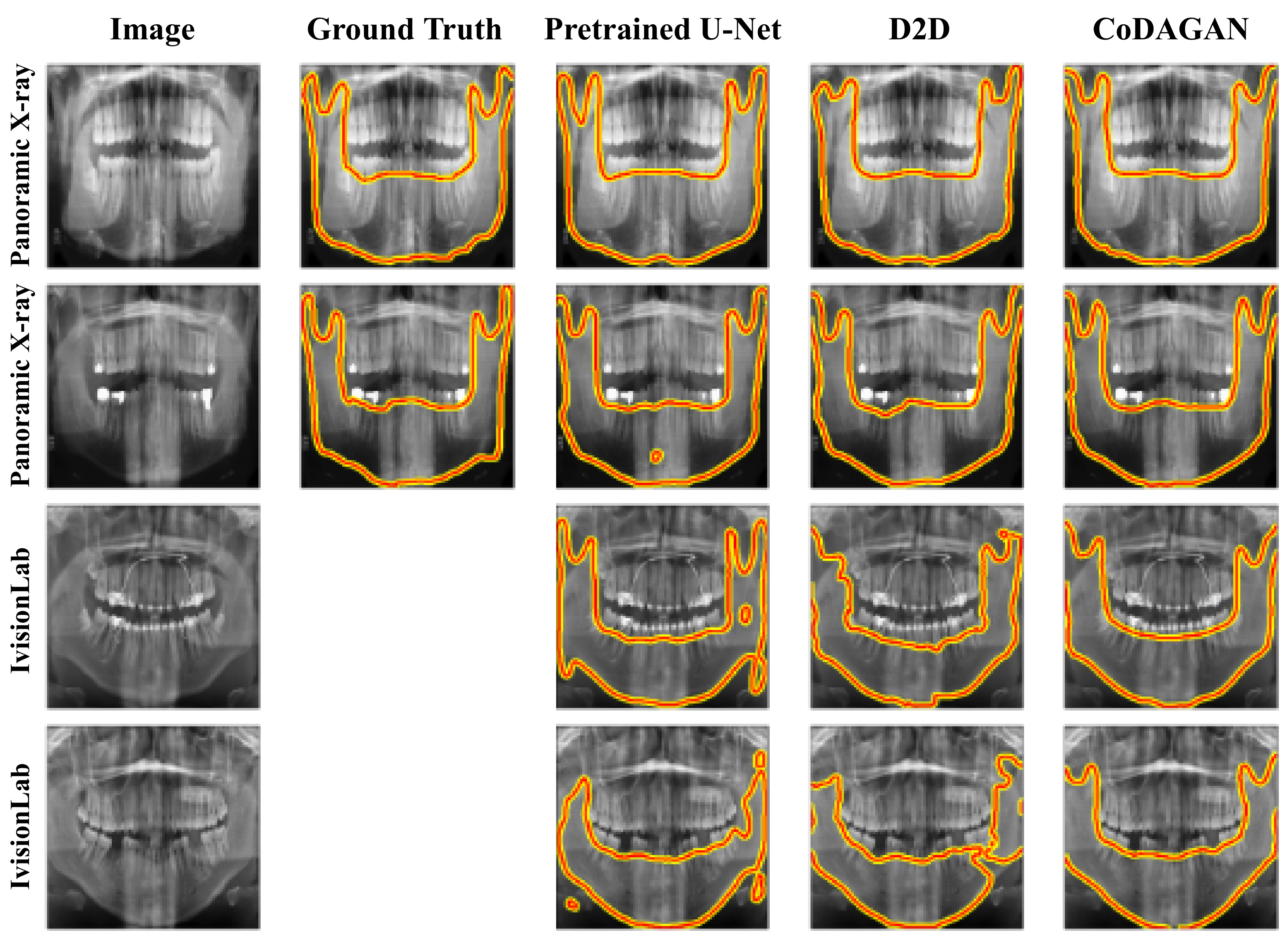}
        \label{fig:segmentation_dxr_mandible}
    }%
    \caption{Qualitative segmentation results in DXR images for two distinct tasks: $\bm{E_{0\%}}$ teeth (a) and $\bm{E_{0\%}}$ mandible (b).}
    \label{fig:segmentation_dxr}
\end{figure*}

\begin{figure*}[!t]
    \centering
    \renewcommand{\currprop}{0.9\columnwidth}
    \subfloat[]{
        \includegraphics[width=\currprop]{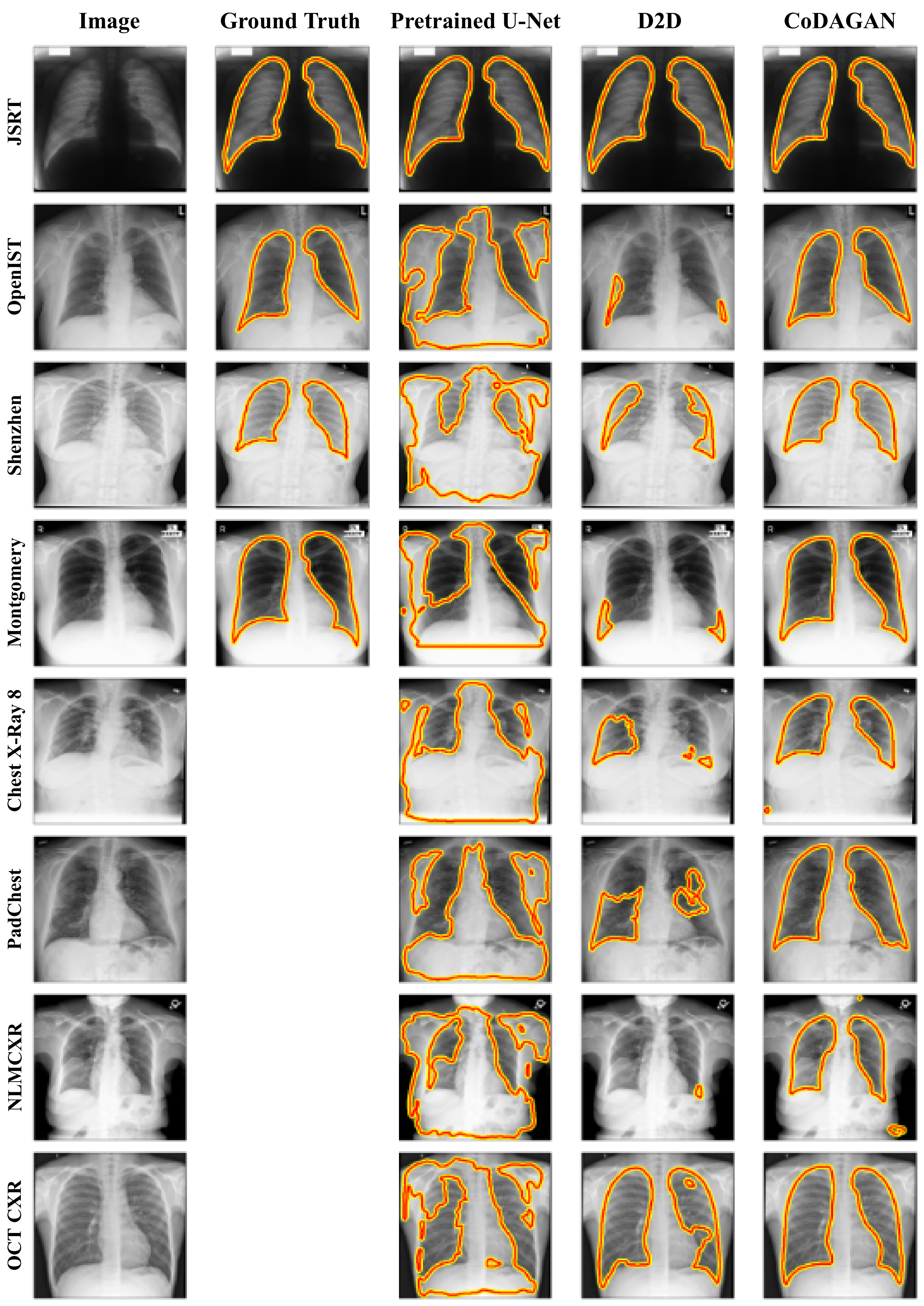}
        \label{fig:segmentation_cxr_lungs}
    }%
    \hfil
    \subfloat[]{
        \includegraphics[width=\currprop]{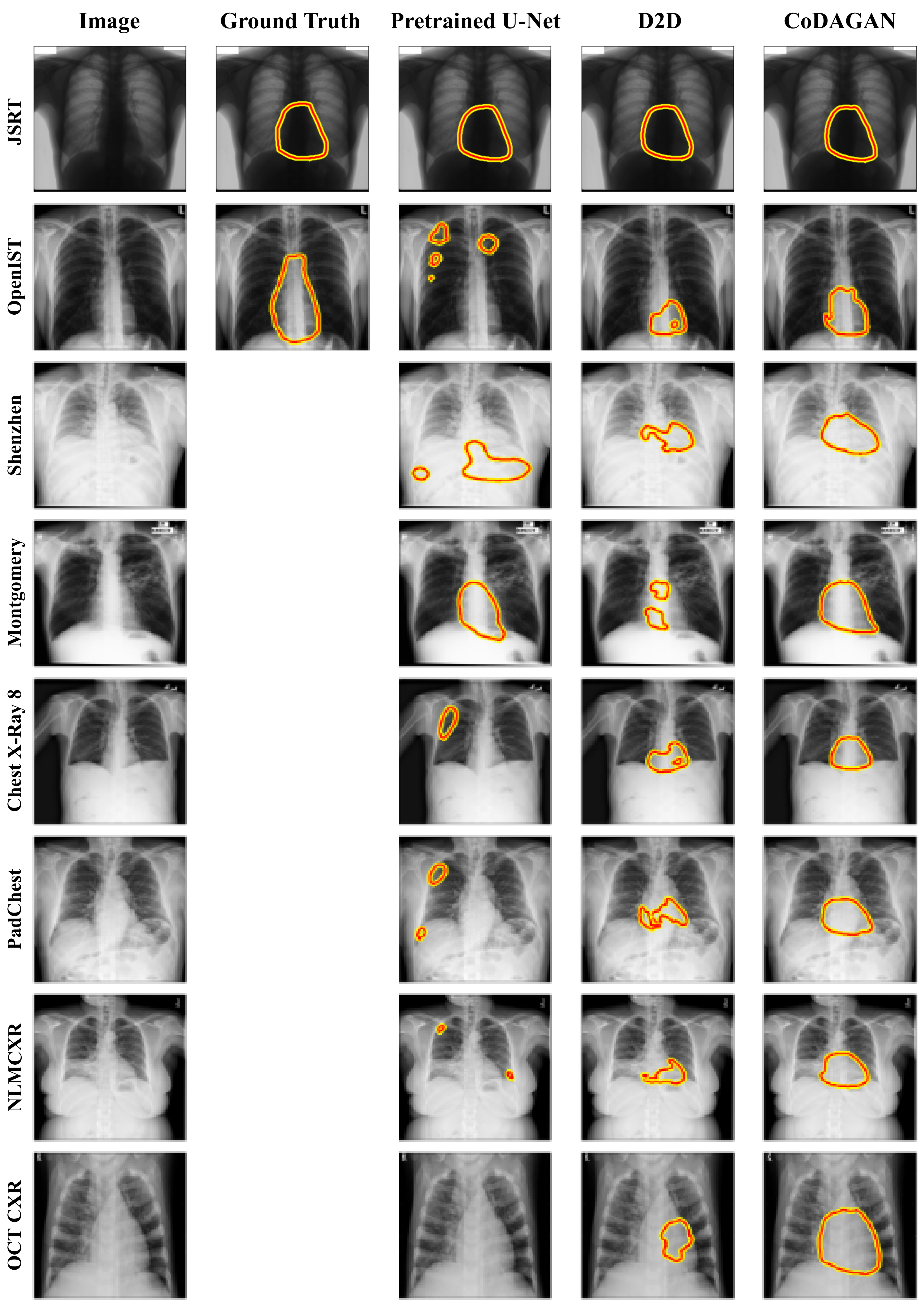}
        \label{fig:segmentation_cxr_heart}
    }%
    \\
    \subfloat[]{
        \includegraphics[width=\currprop]{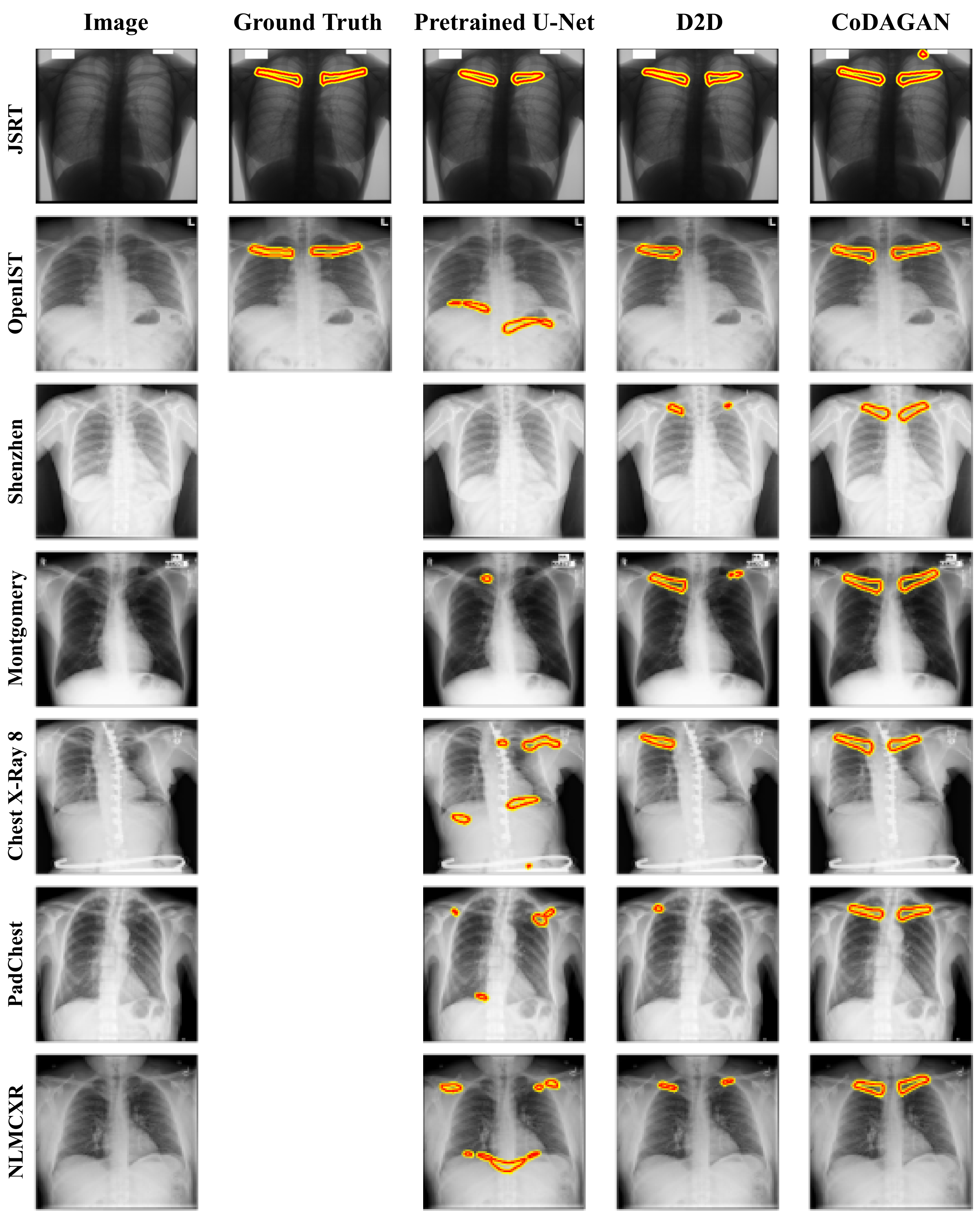}
        \label{fig:segmentation_cxr_clavicles}
    }
    \caption{Qualitative segmentation results in CXR images for three distinct tasks: $\bm{E_{0\%}}$ lungs (a), $\bm{E_{0\%}}$ heart (b) and $\bm{E_{0\%}}$ clavicles (c).}
    \label{fig:segmentation_cxr}
\end{figure*}

Figures~\ref{fig:segmentation_mxr},~\ref{fig:segmentation_dxr} and~\ref{fig:segmentation_cxr} show segmentation qualitative results for two tasks of mammographic image segmentation, two tasks of DXR segmentation and three tasks of CXR segmentation, respectively. Figure~\ref{fig:segmentation_mxr_pectoral} presents predictions for pectoral muscle segmentation $E_{0\%}$, while Figure~\ref{fig:segmentation_mxr_breast} shows breast region segmentation on experiment $E_{0\%}$. Experiment $E_{0\%}$ for lung field segmentation can be seen in Figure~\ref{fig:segmentation_cxr_lungs}, while heart and clavicle segmentation DA experiments ($E_{0\%}$) are shown respectively on Figures~\ref{fig:segmentation_cxr_heart} and~\ref{fig:segmentation_cxr_clavicles}. At last, DXR segmentations from UDA experiments can be seen in Figure~\ref{fig:segmentation_dxr} for both teeth (Figure~\ref{fig:segmentation_dxr_teeth}) and mandible (Figure~\ref{fig:segmentation_dxr_mandible}) segmentation. Columns for all figures present the original sample, the ground truth segmentation for the specific task for this sample when available and predictions from Pretrained U-Nets, D2D and CoDAGANs for visual comparison.

Each row in Figures~\ref{fig:segmentation_mxr_pectoral} and~\ref{fig:segmentation_mxr_breast} highlights one sample from each one of the six MXR datasets used in our experiments. One can see in both figures that D2D underperformed in most cases, failing to predict any pectoral muscle pixel as positive in multiple samples from target datasets. UDA for breast region segmentation also proved to be a hard task for D2D, as in most samples it segmented either only the pectoral muscle or background. While in the pectoral muscle segmentation task most methods were able to successfully ignore the labels in the background of some digitized datasets such as DDSM, MIAS and LAPIMO, these artifacts were shown to be harder to compensate for on breast region segmentation, as all baselines and CoDAGANs wrongly and frequently segmented them as part of the breast. We observed overwhelmingly better results in our qualitative assessment from CoDAGANs, when compared to all other baselines. CoDAGAN superiority proved to be stable both in easier target datasets such as MIAS or BCDR and in more difficult ones as DDSM A and LAPIMO, which contain extremely low contrast and large digitization artifacts, respectively. At last, as the breast boundary contour is fuzzy and extremely hard to segment even for humans in non-FFDM datasets, all methods either underrepresent or overrepresent positive breast pixels in these regions in most samples and datasets.

Figure~\ref{fig:segmentation_dxr_teeth} shows teeth segmentation predictions for both source (IvisionLab) and target (Panoramic X-ray) datasets, while Figure~\ref{fig:segmentation_dxr_mandible} presents DXR mandible segmentations using Panoramic X-ray as source and IvisioLab as target. DXR results show that Pretrained U-Nets and D2D, as expected from a supervised setting, yield mostly predictions in the source dataset for both tasks. However, both methods underperform in the target datasets, missing the segmentation of several teeth and mislabeling mandible regions as background. CoDAGANs achieve much more consistent results in the target datasets, once again evidencing the method's capabilities in UDA. However, CoDAGAN predictions were observed to be less robust for modeling sharp corners in the shapes probably due to the smaller spatial resolution of representation $I$ when compared to the images themselves, which may lead to loss of small detail and slightly smoother shape contours. This issue might be fixed by passing the outputs of the encoder layers in $G_{\mathbb{E}}$ to the supervised model $M$ in order to preserve spatial information, much like a skip connection does.

Figure~\ref{fig:segmentation_cxr_lungs} shows DA results for lung field segmentation in 4 fully labeled datasets (JSRT, OpenIST, Shenzhen and Montgomery) and 4 other target unlabeled datasets (Chest X-Ray 8, PadChest, NLMCXR and OCT CXR). We reiterate that one single CoDAGAN was trained for all datasets and made all predictions contained in the last column of Figure~\ref{fig:segmentation_cxr_lungs}. One should notice that the target datasets in this case are considerably harder than the source ones due to poor image contrast, presence of unforeseen artifacts as pacemakers, rotation and scale differences and a much wider variety of lung sizes, shapes and health conditions. Yet the DA procedure using CoDAGANs for lung segmentation was adequate for the vast majority of images, only presenting errors in distinctly difficult images. As the source dataset (JSRT) has completely distinct visual patterns when compared to the target datasets, both Pretrained U-Nets and D2D are not able to properly compensate for domain shift in these cases, yielding grossly wrong predictions.

Heart and clavicle segmentation (Figures~\ref{fig:segmentation_cxr_heart} and~\ref{fig:segmentation_cxr_clavicles}) are harder tasks than lung segmentation due to heart boundary fuzziness and a high variability of clavicle sizes, shapes and positions. In addition, clavicle segmentation is a highly unbalanced task. Those factors, paired with the fact that the well-behaved samples from the JSRT dataset are the only source of labels to this task contributed to higher segmentation error rates mainly in clavicle segmentation. Results for heart and clavicles are presented for the same 8 datasets as lung segmentation, but only a small subset of OpenIST contains labels for clavicles and heart. Even with all these hampers, CoDAGANs still yielded consistent prediction maps for hearts and clavicles across all target datasets, while baselines are, again, unable to compensate for domain shifts.

\begin{figure*}[!t]
    \centering
    \renewcommand{\currprop}{0.578571429\columnwidth}
    \begin{minipage}[!t]{\columnwidth}
        \centering
        \subfloat[]{
            \includegraphics[width=\currprop]{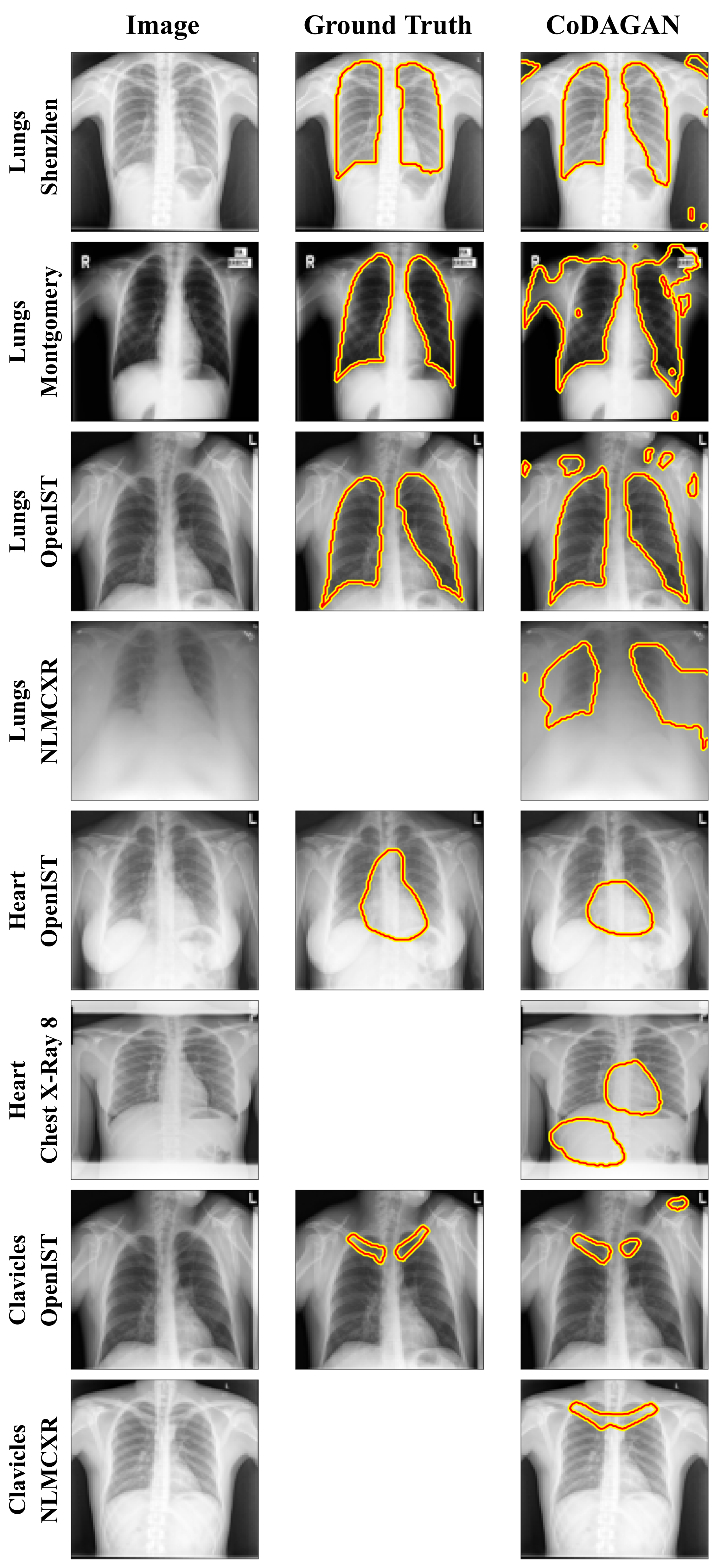}
            \label{fig:segmentation_errors_cxr}
        }%
        \vfill
    \end{minipage}
    \begin{minipage}[!t]{\columnwidth}
        \centering
        \subfloat[]{
            \includegraphics[width=\currprop]{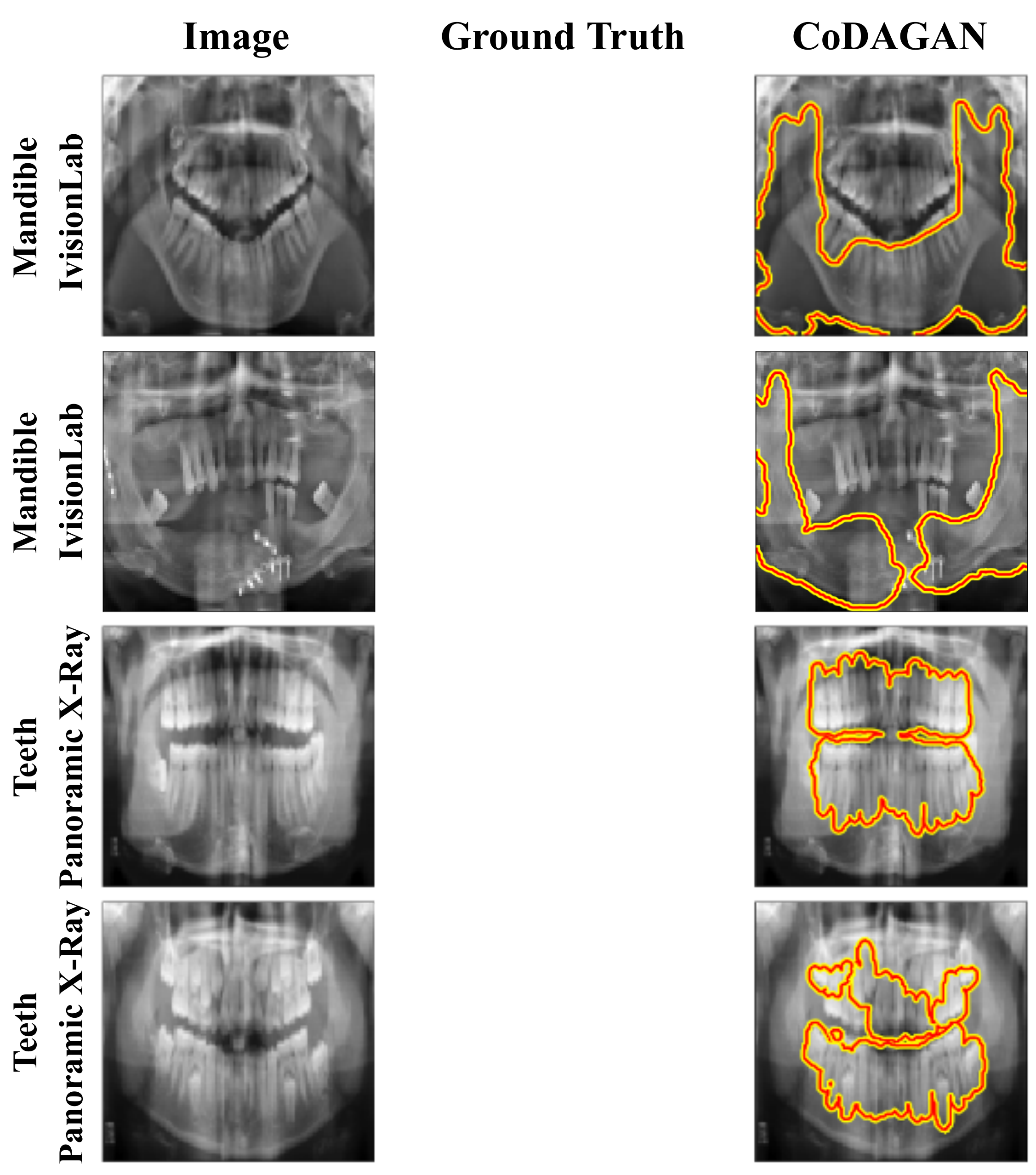}
            \label{fig:segmentation_errors_dxr}
        }
        \\
        \subfloat[]{
            \includegraphics[width=\currprop]{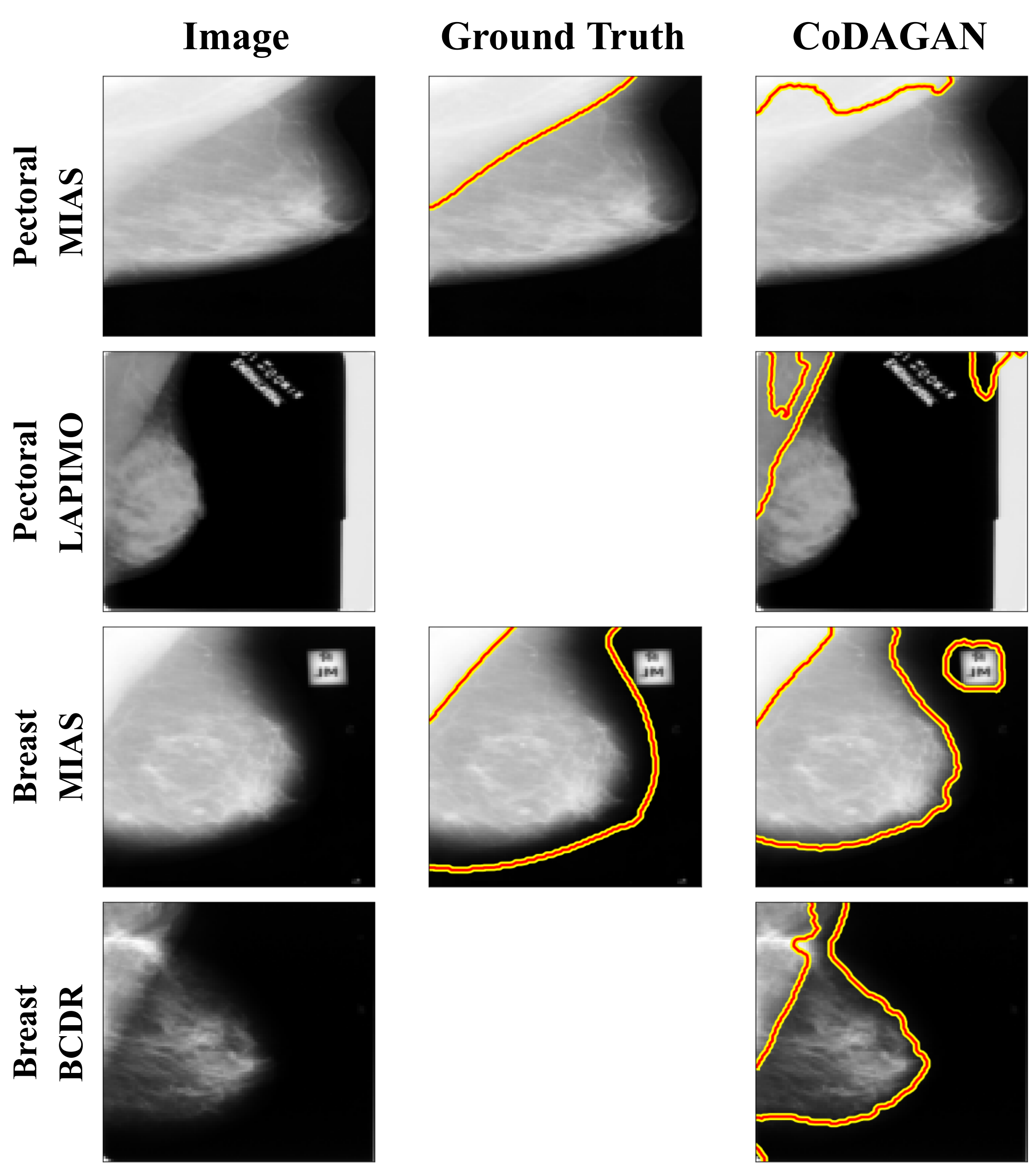}
            \label{fig:segmentation_errors_mxr}
        }
    \end{minipage}
    \caption{Noticeable errors in CoDAGAN UDA results for unlabeled target datasets in three domains: CXRs (a), DXRs (b) and MXRs (c).}
    \label{fig:segmentation_errors}
\end{figure*}

Figure~\ref{fig:segmentation_errors} presents a visual assessment of segmentation errors in CXR (Figure~\ref{fig:segmentation_errors_cxr}), DXR (Figure~\ref{fig:segmentation_errors_dxr}) and MXR (Figure~\ref{fig:segmentation_errors_mxr}) tasks for some samples of target datasets in UDA scenarios. A full assessment of results and both CoDAGAN and baseline errors can be seen in this project's webpage.

One can see that several lung predictions by CoDAGANs yielded small isles of false positives in other bony areas of CXRs (Figure~\ref{fig:segmentation_errors_cxr}) as well as in the background of the images due to wrongly compensated domain shifts. While most of these errors can be corrected by simply filtering for keeping only the larger contiguous areas lung field segmentation and heart segmentation, this would be harder to implement for clavicles due to their smaller relative sizes in CXR exams. Extremely low contrast images as the NLMCXR sample presented in the fourth row of Figure~\ref{fig:segmentation_errors_cxr} presented a challenge for CoDAGANs on all CXR tasks, being the most common source of missed predictions for our method.

We noticed that there were large inter-dataset labeling differences for all CXR tasks. For instance, several OpenIST heart labels contain larger heart delineations than JSRT labels, which led to a larger number of false negatives on OpenIST, as can be seen in the fifth row of Figure~\ref{fig:segmentation_errors_cxr}. Also, clavicle labels on JSRT delineate only pixels within lung borders, while OpenIST labels delineate the whole pair of bones both inside and outside the lung fields. We employed a binary mask between clavicles and lungs for each labeled OpenIST sample in order to fix this discrepancy in labeling characteristics.

Even though both Pretrained U-Nets and D2D yielded worse general results in DXR tasks, CoDAGANs still missed a considerable number of teeth, failed to separate the upper and lower dental arches and wrongfully split mandibles, as shown in Figure~\ref{fig:segmentation_errors_dxr}. At last, MXR prediction errors can be seen in Figure~\ref{fig:segmentation_errors_mxr}, mainly in denser breasts, which hamper the differentiation between pectoral muscle and breast tissue and due to fuzzy breast-boundary borders. Some of the non-FFDM datasets also contain digitization artifacts in the background, which were frequently misclassified as breast pixels. Therefore, there is still a lot of room for improvement in CoDAGAN's domain shift compensation capabilities.



\subsection{Qualitative Analysis of Isomorphic Representations}
\label{sec:results_qualitative_i}

\begin{figure*}[!t]
    \centering
    \renewcommand{\currprop}{0.425\textheight}
    \subfloat[]{
        \includegraphics[height=\currprop]{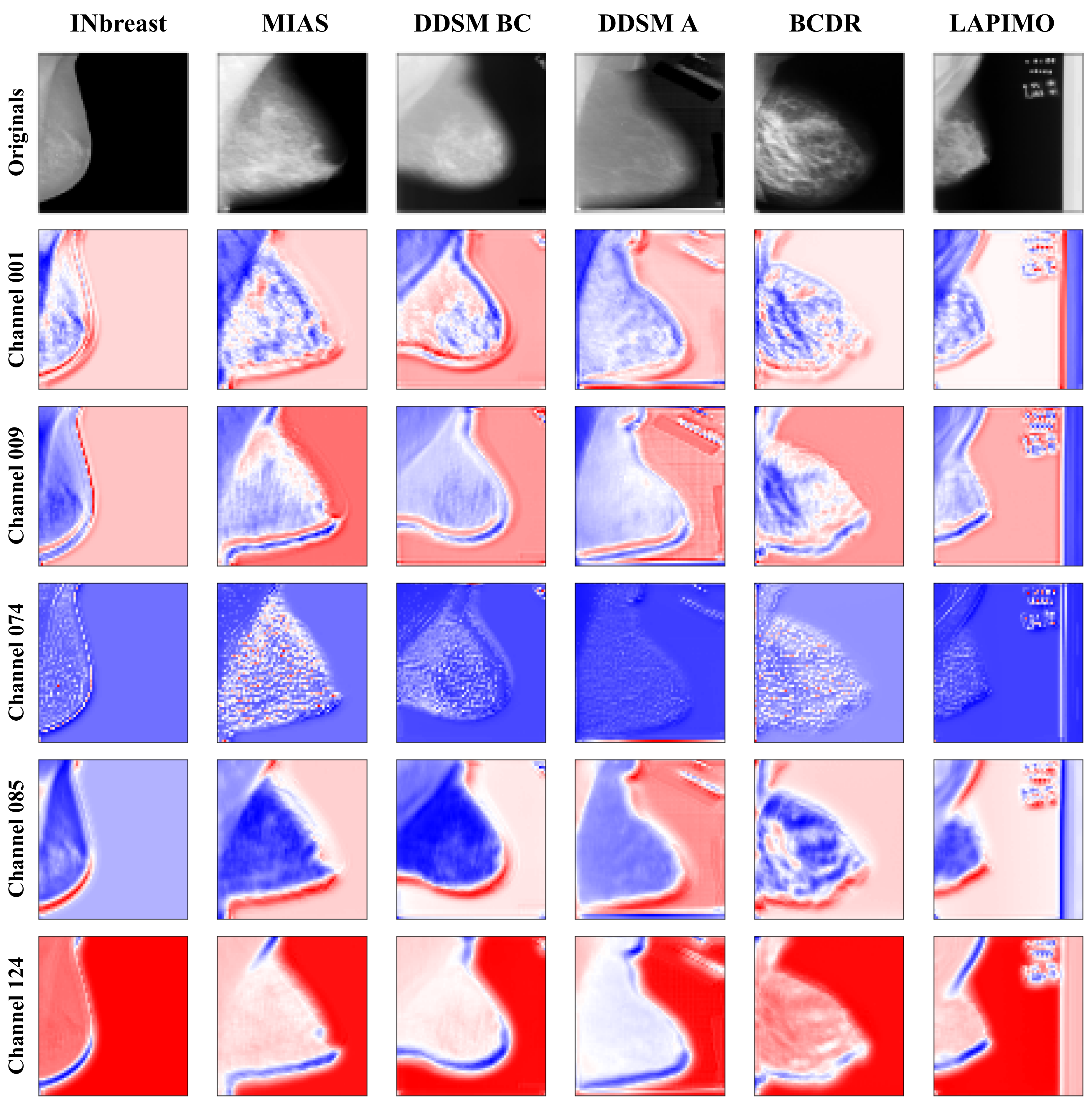}
        \label{fig:iso_mxr}
    }
    \hfil
    \subfloat[]{
        \includegraphics[height=\currprop]{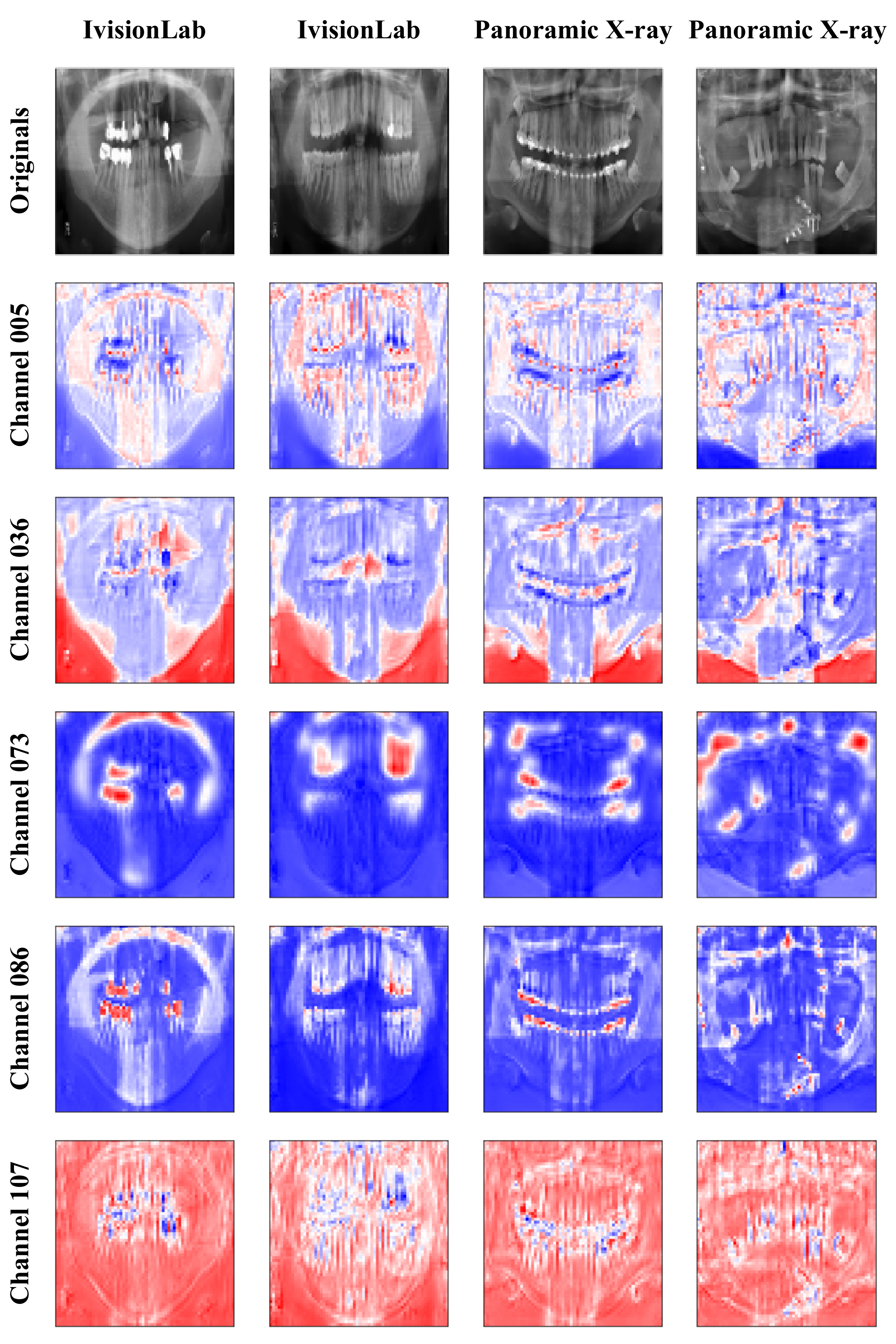}
        \label{fig:iso_dxr}
    }
    \\
    \subfloat[]{
        \includegraphics[height=\currprop]{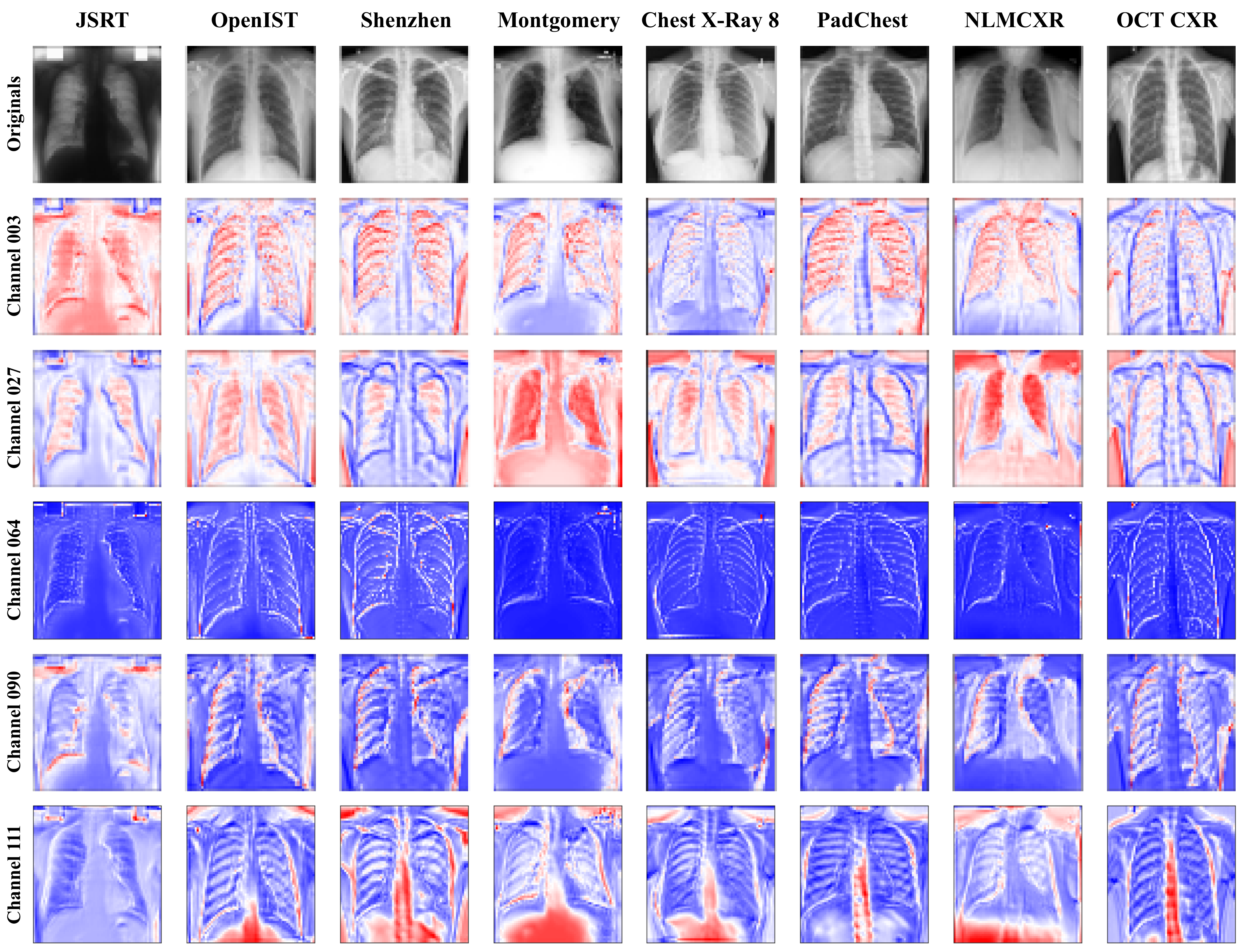}
        \label{fig:iso_cxr}
    }
    \caption{Original images and five different activation channels for samples of (a) MXRs, (b) DXRs and (c) CXRs.}
    \label{fig:iso}
\end{figure*}

Another important qualitative assessment to be performed in CoDAGANs is to visually assess that the same objects in distinct datasets are represented similarly in $I$-space. This is shown in Figure~\ref{fig:iso} for three different activation channels in MXRs (Figure~\ref{fig:iso_mxr}), DXRs (Figure~\ref{fig:iso_dxr}) and CXRs (Figure~\ref{fig:iso_cxr}) from distinct datasets.

In Figure~\ref{fig:iso_mxr}, high density tissue patterns and important object contours in the images from INbreast, MIAS, DDSM BC, DDSM A, BCDR and LAPIMO are encoded similarly by CoDAGANs. Breast boundaries are also visually similar across samples from all mammographic datasets, as CoDAGANs are able to infer that these information is semantically similar despite the differences in the visual patterns of the images. Visual patterns that compose the patient's anatomical structures, such as ribs and lung contours, in Figure~\ref{fig:iso_cxr} are visibly similar in the samples from all eight CXR datasets: JSRT, OpenIST, Shenzhen, Montgomery, Chest X-Ray 8, PadChest, NLMCXR and OCT CXR. The third radiological domain used in our comparisons is composed of two different DXRs datasets: IvisionLab and Panoramic X-Ray (Figure~\ref{fig:iso_dxr}). It is easy to notice the common patterns encoded by CoDAGANs for the same semantic areas of the distinct images such as the teeth edges and mandible contours. One should notice that despite the clear visual distinctions between the original samples from the different datasets in all domains, the isomorphic representations were visually alike across samples from the domains. These results show that CoDAGANs successfully create a joint representation for high semantic-level information which encodes analogous visual patterns across datasets in a similar manner. In other words, different convolutional channels in $I$ activate visual patterns with the same semantic information from the distinct datasets in a similar manner. This feature of encoding a joint distribution between domains by looking only to the marginal distributions of the samples is what allows CoDAGANs to perform UDA, SSDA and FSDA with high accuracy.

\subsection{Low Dimensionality}
\label{sec:results_dimensionality}

\begin{figure*}[!t]
    \centering
    \renewcommand{\currprop}{0.4\textwidth}
    \subfloat[]{
        \begin{tikzpicture}
            \draw (0, 0) node[inner sep=0] {\includegraphics[page=1, clip, trim=0.2cm 0.2cm 0.2cm 0.2cm, width=\currprop]{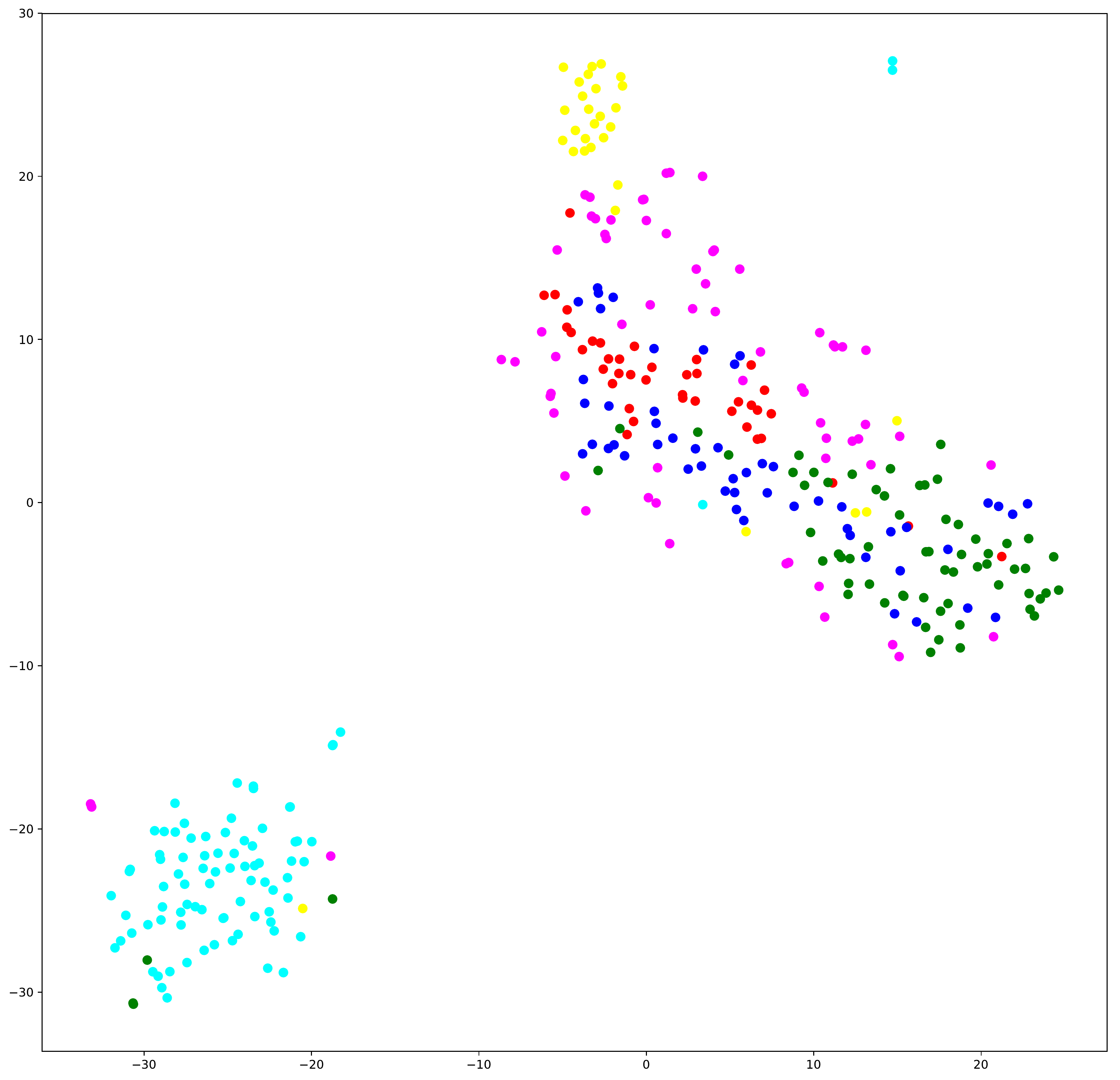}};
            \draw (-2.30, 2.25) node[inner sep=0] {\includegraphics[width=0.1\textwidth]{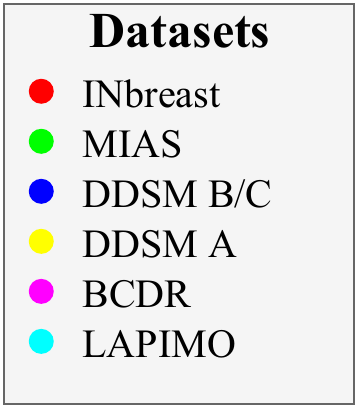}};
        \end{tikzpicture}
        \label{fig:lowres_mxr_1}
    }%
    \hfil
    \subfloat[]{
        \begin{tikzpicture}
            \draw (0, 0) node[inner sep=0] {\includegraphics[page=1, clip, trim=0.2cm 0.2cm 0.2cm 0.2cm, width=\currprop]{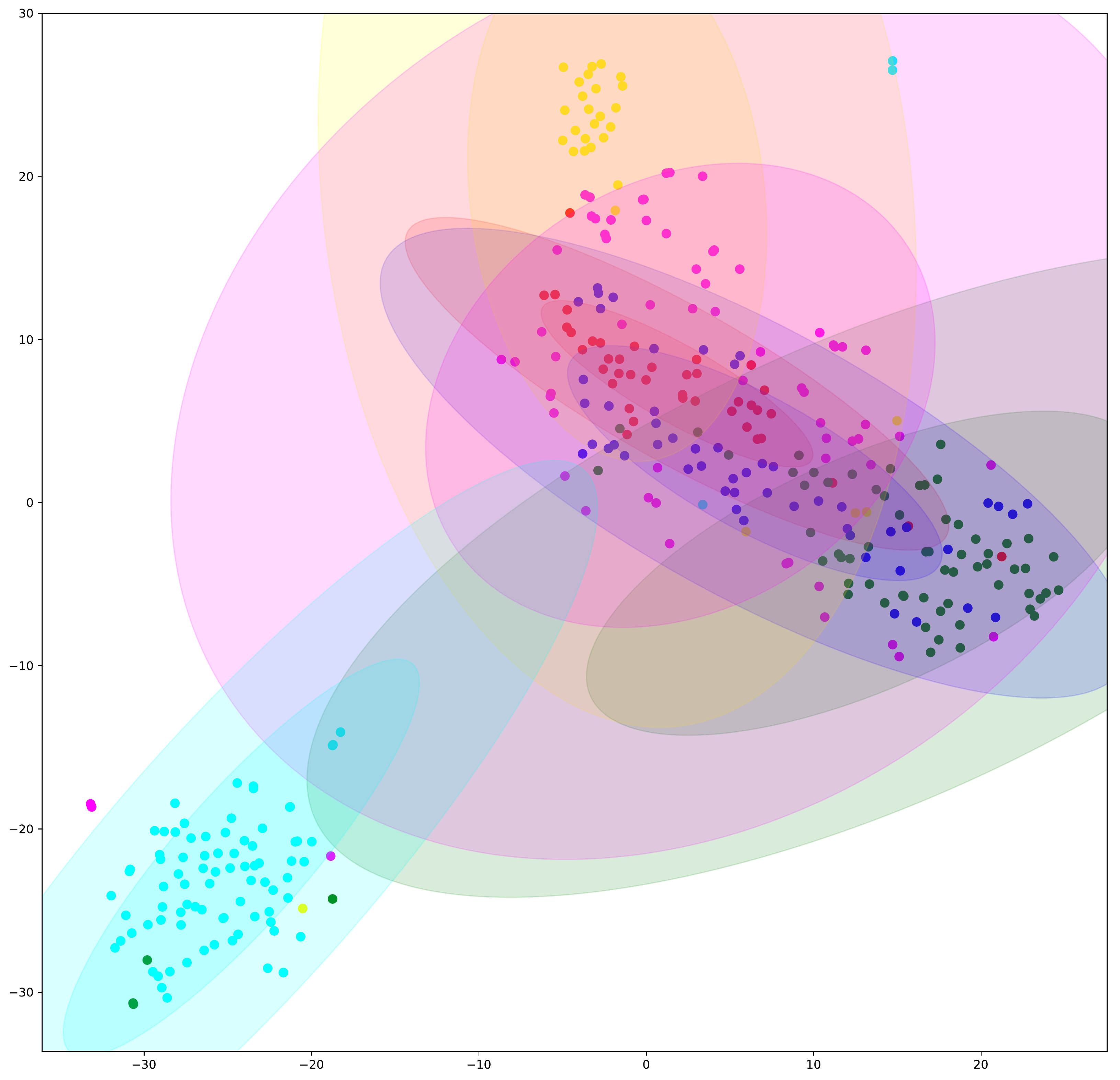}};
            \draw (-2.30, 2.25) node[inner sep=0] {\includegraphics[width=0.1\textwidth]{figs/legends_mxr.pdf}};
        \end{tikzpicture}
        \label{fig:lowres_mxr_2}
    }
    \\
    \subfloat[]{
        \begin{tikzpicture}
            \draw (0, 0) node[inner sep=0] {\includegraphics[page=1, clip, trim=0.2cm 0.2cm 0.2cm 0.2cm, width=\currprop]{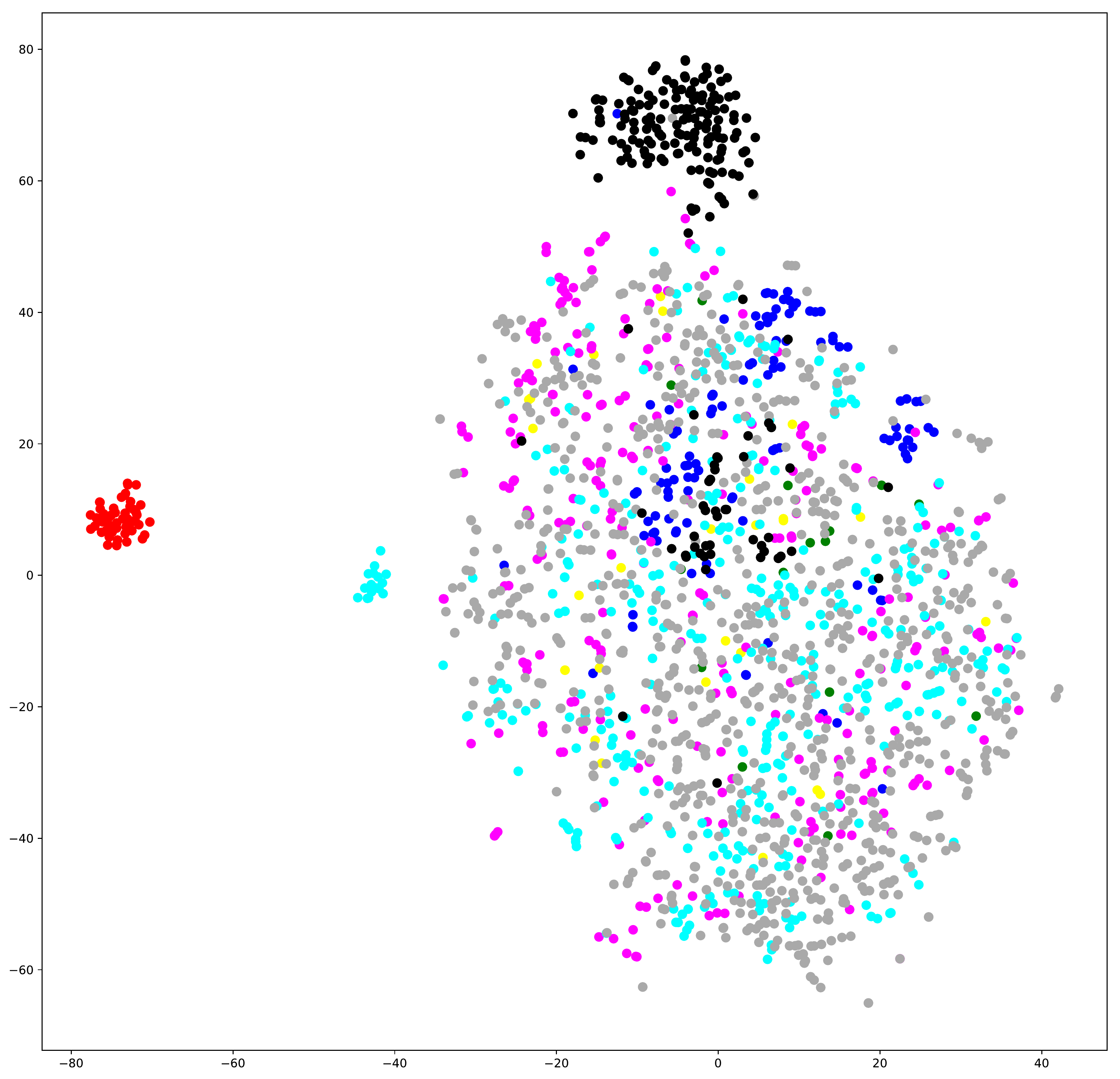}};
            \draw (-2.30, 2.00) node[inner sep=0] {\includegraphics[width=0.1\textwidth]{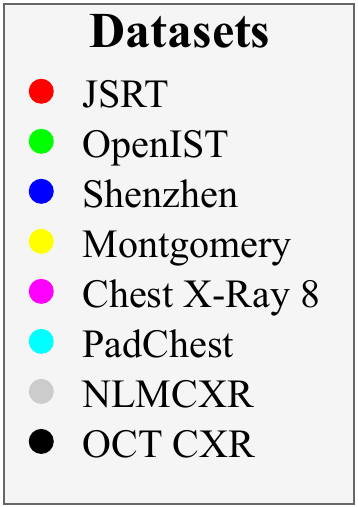}};
        \end{tikzpicture}
        \label{fig:lowres_cxr_1}
    }%
    \hfil
    \subfloat[]{
        \begin{tikzpicture}
            \draw (0, 0) node[inner sep=0] {\includegraphics[page=1, clip, trim=0.2cm 0.2cm 0.2cm 0.2cm, width=\currprop]{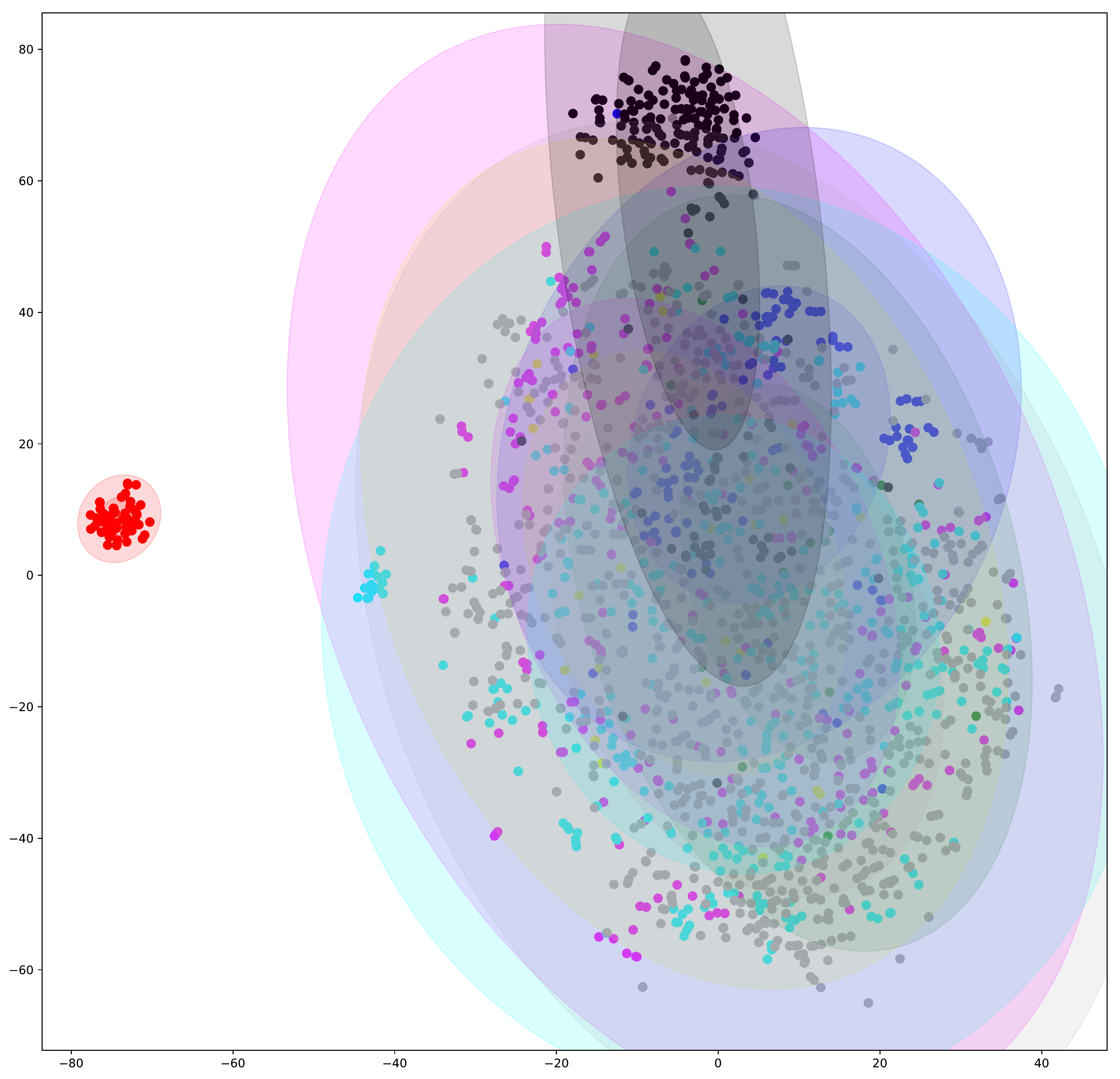}};
            \draw (-2.30, 2.00) node[inner sep=0] {\includegraphics[width=0.1\textwidth]{figs/legends_cxr.pdf}};
        \end{tikzpicture}
        \label{fig:lowres_cxr_2}
    }
    \caption{2D projections of in $I$-space for MXRs in pectoral muscle segmentation (a, b) and CXRs in the lung segmentation task (c, d). The original 2D-projected samples after PCA/t-SNE (a, c) are shown conjointly with gaussian fits over the data (b, d) for both domains.}
    \label{fig:lowres}
\end{figure*}

In order to view the data distributions of samples from the different datasets in the $I$-space of CoDAGAN representations, we reduced the dimensionality of $I$ to a 2D visualization using Principal Component Analysis (PCA) and the t-Distributed Stochastic Neighbor Embedding (t-SNE) algorithm \cite{Maaten:2008}. First, in order to reduce computational requirements, we reduced the original $524,288$ dimensions of $I$ to a $200$-dimensional space using PCA and applied t-SNE on the remaining components. We also fit a gaussian on the data distributions for each dataset using the Gaussian Mixture Model (GMM) from sklearn\footnotemark \footnotetext{\url{https://scikit-learn.org/stable/}}. The resulting 2D visualizations of the mammography and CXR datasets can be seen in Figure~\ref{fig:lowres}.

Figures~\ref{fig:lowres_mxr_1} and~\ref{fig:lowres_mxr_2} show respectively the original 2D representation of the $I$-space from PCA/t-SNE and the GMM fit for the data on the mammographic datasets. Visual analysis of Figure~\ref{fig:lowres_mxr_2} shows the domain shifts between LAPIMO and the other MXR datasets. This is due to the fact that LAPIMO samples have a characteristic digitization artifact on one side of all samples, as can be easily seen in Figures~\ref{fig:segmentation_mxr_pectoral} and~\ref{fig:segmentation_mxr_breast}. 
Not coincidentally, these artifacts in LAPIMO samples hampered CoDAGAN abilities to compensate for domain shift and severely hampered the segmentation quality in all baselines.

A similar pattern can be seen in Figures~\ref{fig:lowres_cxr_1} and~\ref{fig:lowres_cxr_2}, which show respectively the 2D projections of CXR datasets and the GMM fits for these data. JSRT samples have the most standardized data among all CXR datasets, containing only high visual quality samples with fixed posture, high contrasts between anatomical structures (i.e. lungs, ribs, etc) and no major lung shape-distorting illnesses (i.e. pneumonia, tuberculosis. etc). Other datasets -- such as Chest X-Ray 8, Montgomery and Shenzhen -- present more real-world scenarios with a high variety of lung shapes and sizes and smaller control over patient's position during the exam, that is, higher rotation, scale and translations in these images. Thus samples from the JSRT dataset in Figure~\ref{fig:lowres_cxr_2} are clustered in a small region in the 2D projection of $I$-space, while the other datasets contain more spread samples in this projection. This result evidences that the use of distinct sources of data should better enforce satisfactory Domain Generalization for the supervised model $M$ in CoDAGANs.

Another visibly distinct cluster in Figure~\ref{fig:lowres_cxr_2} is formed of samples pertaining to the OCT CXR set. Samples from this dataset were noticeably harder to segment due to their smaller contrast range. OCT CXR patients also performed the exam on a distinct position with their arms pointing upward, contrary to all other CXR data used in our experiments. These visual features reinforce this dataset's distinction from other CXRs in our experiments and explain its homogeneity in the 2D projections of Figure~\ref{fig:lowres_cxr_2}.

Another use for these 2D projections could be to perform inference from datasets that were never trained by the algorithm, effectively achieving Domain Generalization \cite{Zhang:2017} for new samples. This Domain Generalization CoDAGAN could find the natural cluster closer to the new data according to a dissimilarity metric and assign the novel samples to the cluster. This approach could, therefore, personalize the One-Hot-Encoding so that it better captures the particular visual patterns of previously unseen data.
\section{Conclusion}
\label{sec:conclusion}

This paper proposed and validated a method that covers the whole spectrum of UDA, SSDA and FSDA in dense labeling tasks for multiple source and target biomedical datasets. We performed an extensive quantitative and qualitative experimental evaluation on several distinct domains, datasets and tasks, comparing CoDAGANs with both traditional and recent baselines in the DA literature. CoDAGANs were shown to be an effective DA method that could learn a single model that performs satisfactory predictions for several different datasets in a domain, even when the visual patterns of these data were clearly distinct. The proposed method was able to gather both labeled and unlabeled data in its inference process, making it highly adaptable to a wide variety of data scarcity scenarios in SSDA.

We showed that CoDAGANs can be build upon two distinct Unsupervised Image-to-Image Translation methods (UNIT \cite{Liu:2017} and MUNIT \cite{Huang:2018}), evidencing its agnosticism to the underlying image translation architecture. It is also evident in both our background comparisons (Section~\ref{sec:related}) and in our experimental evaluation (Section~\ref{sec:results}) that CoDAGANs are distinct from simpler D2D approaches, which are recurrent in the literature of image translation for DA \cite{Liu:2016,Bousmalis:2017,Hoffman:2018,Oliveira:2018,Murez:2018,Wu:2018,Zou:2018}.

It was observed in Sections~\ref{sec:results_mxr} and~\ref{sec:results_cxr} that CoDAGANs achieve results in fully unsupervised settings that are comparable to supervised DA methods -- such as Fine-tuning to new data. These experiments also showed that both Pretrained DNNs and D2D approaches were ineffective in scarce labeling scenarios. CoDAGANs presented significantly better Jaccard values in most experiments where labeled data was scarce in the large variety of target datasets studied. Fine-tuning and From Scratch training were only able to achieve good objective results when labeled data was abundant -- that is, in scenarios closer do $E_{100\%}$. It is important to reiterate that label scarcity is a major problem in real world biomedical image tasks, mainly for dense labeling tasks.

CoDAGANs were observed to perform satisfactory DA even when the labeled source dataset was considerably simpler than the target unlabeled datasets, as presented in Section~\ref{sec:results}. In experiment $E_{0\%}$ for CXR lung, clavicle and heart segmentations, the JSRT source dataset contains images acquired in a much more controlled environment than all of the target datasets. In addition, experiment $E_{0\%}$ only used INbreast samples for training and was able to perform DA for more real-world scenario datasets, such as DDSM, BCDR and LAPIMO, albeit with some segmentation artifacts in many samples due to poor variability in training data. Another evidence of the capabilities of CoDAGANs is the good performance in DA tasks even for highly imbalanced classes, as the case of clavicle segmentation, wherein the Region of Interest in the images represents only a tiny portion of the pixels. 

\section*{Acknowledgment}

Authors would like to thank NVIDIA for the donation of the GPUs that allowed the execution of all experiments in this paper. We also thank CAPES, CNPq (424700/2018-2), and FAPEMIG (APQ-00449-17) for the financial support provided for this research.

\bibliographystyle{IEEEtran}
\bibliography{access}

\clearpage
\begin{IEEEbiography}[{\includegraphics[width=1in,height=1.25in,clip,keepaspectratio]{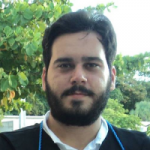}}\vspace{7mm}]{Hugo Oliveira}
obtained his BSc. (2014) and MSc. (2016) degrees in Computer Science by Universidade Federal da Para\'{i}ba (UFPB). Currently a PhD Candidate in Computer Science by Universidade Federal de Minas Gerais (UFMG), a professor at the Technical College (COLTEC/UFMG) and a member of the Pattern Recognition and Earth Observation (PATREO) Laboratory. Research interests include Machine Learning, Deep Learning, Image Processing, Biomedical Images, Domain Adaptation, Data Compression and Information Theory.
\end{IEEEbiography}
\vspace{-90mm}
\begin{IEEEbiography}[{\includegraphics[width=1in,height=1.25in,clip,keepaspectratio]{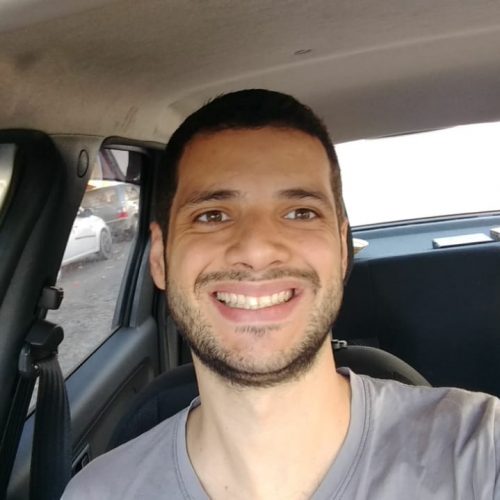}}\vspace{7mm}]{Edemir Ferreira}
obtained his BSc. degree in Computational Mathematics by Universidade Federal de Minas Gerais (2013) and MSc. degree in Computer Science by Universidade Federal de Minas Gerais (2016). He has experience in Geoscience with focus on Remote Sensing problems.
\end{IEEEbiography}
\vspace{-90mm}
\begin{IEEEbiography}[{\includegraphics[width=1in,height=1.25in,clip,keepaspectratio]{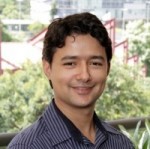}}\vspace{7mm}]{Jefersson dos Santos}
received the Ph.D. degree in computer science from the Universit\'{e} de Cergy-Pontoise, Cergy, France, and from the University of Campinas, Campinas, Brazil, in 2013.
He is currently an Adjunct Professor with the Department of Computer Science, Universidade Federal de Minas Gerais, Belo Horizonte, Brazil. He has been holding the Research Productivity Scholarship from the Brazilian Research Council (CNPq) since 2016. He has published several articles in journals with high impact factor and selective editorial policy. He has also published more than 30 articles in important conferences of remote sensing, image processing, and computer vision. He has experience in coordinating research with Brazilian funding agencies and research and development projects with companies in those topics. He is the Founder and the Coordinator of the Laboratory of Pattern Recognition and Earth Observation, PATREO (http://www.patreo.dcc.ufmg.br), one of the Brazil's pioneer groups focused on the development of computer vision and machine learning for remote sensing applications.
\end{IEEEbiography}

\EOD

\end{document}